\documentclass[10pt,twocolumn,letterpaper,table]{article}

\usepackage[pagenumbers]{cvpr} 

\definecolor{cvprblue}{rgb}{0.21,0.49,0.74}
\usepackage[pagebackref,breaklinks,colorlinks,allcolors=cvprblue]{hyperref}
\usepackage{multirow}
\usepackage{booktabs}
\usepackage{amssymb, amsmath}
\usepackage{siunitx}
\usepackage{bm}
\usepackage{xfp}
\usepackage{comment}
\usepackage{microtype}
\usepackage{xspace}
\usepackage{accents}
\usepackage{nicefrac}
\usepackage{animate}

\newcommand{\diffcolordown}[1]{%
    \ifnum\fpeval{sign(#1)}=1
        \ifdim#1pt>100pt
            \textcolor{blue}{$>$}%
        \else
            \textcolor{blue}{#1}%
        \fi
    \else
        \textcolor{red}{#1}%
    \fi
}
\newcommand{\diffcolorup}[1]{%
    \ifnum\fpeval{sign(#1)}=1
        \ifdim#1pt>100pt
            \textcolor{red}{$<$}%
        \else
            \textcolor{red}{#1}%
        \fi
    \else
        \textcolor{blue}{#1}%
    \fi
}

\newcommand{\fpcamspaceoffsets}{CSO\xspace}
\newcommand{\fpdepthchangeopflow}{$\Delta D$+OF\xspace}
\newcommand{\fpendpoint}{EP\xspace}

\definecolor{darkgreen}{RGB}{30,150,30}
\definecolor{darkblue}{RGB}{0,0,127}
\definecolor{darkyellow}{RGB}{171,133,0}
\definecolor{darkred}{RGB}{180,20,20}
\definecolor{darkmagenta}{RGB}{127,0,127}
\definecolor{darkcyan}{RGB}{0,127,127}
\definecolor{chromeyellow}{rgb}{1.0, 0.65, 0.0}
\definecolor{amber}{rgb}{1.0, 0.75, 0.0}

\def\paperID{7740} 
\def\confName{CVPR}
\def\confYear{2025}

\title{Zero-Shot Monocular Scene Flow Estimation in the Wild}

\author{Yiqing Liang\textsuperscript{1,2} \and
Abhishek Badki\textsuperscript{1}$^*$ \and
Hang Su\textsuperscript{1}$^*$ \and
James Tompkin\textsuperscript{2} \and
Orazio Gallo\textsuperscript{1} \\
\vspace{-7mm}
\and
\textsuperscript{1}NVIDIA Research
\and
\textsuperscript{2}Brown University}
\begin{document}

\twocolumn[{%
\renewcommand\twocolumn[1][]{#1}%
\maketitle

\begin{center}
    \centering
    \captionsetup{type=figure}

\animategraphics[width=.24\textwidth,poster=0006]{10}{figures/teaser/framesDavisDataset_judo_00011/frame_low_}{0000}{0012}
\animategraphics[width=.24\textwidth,poster=0006]{10}{figures/teaser/framesDavisDataset_loading_00031/frame_low_}{0000}{0012}
\animategraphics[width=.24\textwidth,poster=0006]{10}{figures/teaser/framesRgbStackingDataset_gear_v2_3-basket_front_right_rgb_img-0_00016/frame_low_}{0000}{0012}
\animategraphics[width=.24\textwidth,poster=0006]{10}{figures/teaser/framesKITTIDataset_000063_00000/frame_low_}{0000}{0012}
\animategraphics[width=.24\textwidth,poster=0006]{10}{figures/teaser/framesDavisDataset_mbike-trick_00045/frame_low_}{0000}{0012}
\animategraphics[width=.24\textwidth,poster=0006]{10}{figures/teaser/framesDavisDataset_pigs_00046/frame_low_}{0000}{0012}
\animategraphics[width=.24\textwidth,poster=0006]{10}{figures/teaser/framesRgbStackingDataset_rgb2-obs_basket_back_left_pixels_00218/frame_low_}{0000}{0012}
\animategraphics[width=.24\textwidth,poster=0006]{10}{figures/teaser/framesKITTIDataset_000137_00000/frame_low_}{0000}{0012}

    \captionof{figure}{Our model for monocular scene flow estimation predicts accurate pointmaps and 3D offsets for dynamic scenes from two input images (shown below each pane) captured by two cameras $C_1$ and $C_2$ at times $t_1$ an $t_2$, respectively. The animation shows an interpolation of the pointmaps from $C_1$ at $t_1$ to $C_2$ at $t_2$. These examples are from datasets not seen in training, showcasing the strong generalization abilities of our method. [\textbf{Animated figures --- Please click on each in Adobe Reader to play.}]}
    \label{fig:teaser}
\end{center}%
}]

\newcommand\blfootnote[1]{%
  \begingroup
  \renewcommand\thefootnote{}\footnote{#1}%
  \addtocounter{footnote}{-1}%
  \endgroup
}
\blfootnote{
  \noindent$^*$\,indicates equal contribution
}

\vspace{-5mm}

\begin{abstract}
    Large models have shown generalization across datasets for many low-level vision tasks, like depth estimation, but no such general models exist for \emph{scene flow}.
    Even though scene flow has wide potential use, it is not used in practice because current predictive models do not generalize well.
    We identify three key challenges and propose solutions for each.
    First, we create a method that jointly estimates geometry and motion for accurate prediction.
    Second, we alleviate scene flow data scarcity with a data recipe that affords us 1M annotated training samples across diverse synthetic scenes.
    Third, we evaluate different parameterizations for scene flow prediction and adopt a natural and effective parameterization.
    Our resulting model outperforms existing methods as well as baselines built on large-scale models in terms of 3D end-point error, and shows zero-shot generalization to the casually captured videos from DAVIS and the robotic manipulation scenes from RoboTAP.
    Overall, our approach makes scene flow prediction more practical in-the-wild.
    \vspace{-5mm}
\end{abstract}

\noindent\textbf{Release:} \href{https://research.nvidia.com/labs/lpr/zero\_msf/}{https://research.nvidia.com/labs/lpr/zero\_msf//}. We will release the code and weights upon acceptance.

\section{Introduction}
\label{sec:intro}

Scene flow (SF) captures geometric transformations of dynamic scenes by the motion of points in 3D.
As such, its accurate estimation can benefit applications like augmented reality~\cite{Klein2007ParallelTA}, autonomous driving~\cite{6248074}, and robotics~\cite{Xu2024FlowAT}.
Yet, despite its potential, it is not commonly used in practice.
One contributing factor is the limited generalization ability of existing methods.
This may be surprising given the progress of large-scale models~\cite{Bommasani2021OnTO} on other low-level vision tasks.
For instance, the state-of-the-art approaches for tasks like image segmentation~\cite{segmentanything} and depth estimation (single and multi-view)~\cite{dust3r,mast3r,depth_anything_v1,depth_anything_v2} 
lead benchmarks with impressive results.

Why is SF estimation different? 
We identify three critical challenges that hinder the generalization ability of feedforward SF estimation from monocular RGB videos.

First, geometry and motion are entangled, and thus require joint reasoning.
That is because the 2D displacement we observe in image space is the combined effect of depth and motion, so an error in estimating either one will result in incorrect SF predictions.
Even when predicting depth and motion jointly, estimating SF remains ill-posed because of the scale ambiguity, requiring effective learning to resolve it.

Second, good generalization requires training on large and diverse datasets, but labeled SF data are scarce.
Further, some datasets are metric and some are in relative units, which can compromise learning if not properly accounted for.
While this scale issue is well known in single-image depth estimation~\cite{Ranftl2022}, it has not been addressed for SF estimation.
Existing methods do not account for both of these factors, and therefore struggle to generalize to out-of-domain data.

Third, the estimation of SF is sensitive to the choice of parametrization.
We find that, using depth + optical flow or taking the difference between the estimated 3D points before and after motion
both compromise the quality of the results.

Based on these observations, we develop a new model for monocular SF estimation.
We note that recent works in depth estimation like DUSt3R~\cite{dust3r} and MASt3R~\cite{mast3r} show impressive generalization abilities for static scenes, but they do not model motion.
To tackle the entanglement of geometry and motion, rather than using a separate network for motion, we extend them to predict offsets \emph{jointly with geometry}, and train them for both tasks.
Interestingly, we show that this joint training also improves their depth estimates for dynamic scenes, further supporting our observation that geometry and motion are entangled.
To address the second challenge, we use datasets across several domains: indoors and outdoors, and with a diverse set of ground truth annotations.
To help joint geometry and motion prediction quality, we use optical flow to supervise the projection of the SF in image space.
Importantly, to exploit metric and relative datasets, we use a simple and scale-alignment mechanism across datasets.
Finally, aware of the importance of parameterization choice, we study alternatives and propose to represent SF as pointmaps plus 3D motion offsets.

In evaluation against state-of-the-art monocular SF methods~\cite{Hur:2020:SSM,yang2020upgrading} and Depth+Flow baselines, our approach improves both SF and geometry estimation in a zero-shot setting across both real and synthetic monocular SF estimation benchmarks.
Figure~\ref{fig:teaser} uses our SF and pointmap estimates to animate point motion over time.
The quality of these results shows the zero-shot generalization ability of our method, since none of those datasets were used in training.
This strong generalization ability is in contrast to existing SF estimation methods that have reduced performance on data not seen in training.

In summary, we introduce the first scene flow approach that leverages large-scale data for accurate, in-the-wild estimation.
We show that joint training of geometry and motion is central to its success, and that training for motion also helps depth estimation when the scene is dynamic.
Not least, we describe a data and training strategy that is key to our method's ability to learn from large data.

\section{Related Work}
\label{sec:related}

\subsection{Scene Flow Estimation}
\label{sec:rel-sf}
Scene flow (SF) estimation is the task of estimating 3D motion and geometry fields~\cite{sceneflow05}.
Often, works focus on the autonomous driving scenario, and few studies have been conducted on in-the-wild generalization.

Many SF estimation methods, including the earliest ones, assume explicit 3D structure information as input.
It could be a calibrated stereo pair~\cite{zhang2001,sceneflow05,pons2007,huguet2007,basha2010,vogel2011,wedel2011,vogel2013,vogel2014,vogel2015,Menze2015ObjectSF,ren2017,behl2017,ma2019drisf,Schuster2018sceneflowfields,ilg2018,jiang2019sense,mayer2016,saxena2019,lee2019learning,liu2019unsupervised,wang2019unos,Jiao2021EffiScene,Lai2019Bridging}, 
depth images~\cite{hadfield2011,Hornacek2014,Lv18eccv,Qiao2018sfnet,quiroga2014,Thakur2018,teed2021raft3d,jiang2021gma,Zhang2021separableFlow,liu2022camliflow,sui2022craft,mehl2023Mfuse,zhou2023FusionRAFT}
or 3D point clouds~\cite{HPLFlowNet,behl2018pointflownet,NEURIPS2018_f5f8590c,liu2019flownet3d,tishchenko2020self,Mittal2020JGF,wu2020pointpwc,puy20flot,wei2020pv,Kittenplon2020FlowStep3DMU,Baur2021slim,Li2021hcrf,Li2021SPF,dong2022exploiting,ding2022fh,wang2022ThreeDFlow,Jund2022,wang2022pointmotion,battrawy2024rms,Vedder2024zeroflow,zhang2024seflow,pontes2020,Deng2023WACV,li2021neural,li2022rigidflow,Li2023ICCV,lang2023scoop,khatri2024trackflow,zhang2023gmsf,cheng2023multi,yang2023vidar,ouyang2021occlusion,wang2021festa,Cheng2022BiPointFlowNetBL,zhang2024diffsf,liu2024difflow3d,gojcic2021weakly3dsf,Peng2023delflow,Ahuja2024optflow}.
Among point-cloud-based methods,
the community has explored run-time optimization techniques for generalization~\cite{pontes2020,li2021neural,Li2023ICCV,lang2023scoop,Deng2023WACV,Ahuja2024optflow}.
Despite improving accuracy, these techniques incur high computation costs.

Although we also leverage pointmaps in scene flow estimation, our approach is feedforward and therefore much more efficient.

\subsection{Monocular Scene Flow Estimation}
We tackle the more general problem where no 3D information is provided: monocular scene flow (MSF) estimation.
MSF requires methods to estimate both SF and
3D structure from a sequence of monocular images, which is ill-posed by nature. 
Thus, many algorithms~\cite{monosf,yang2020upgrading,Ling2023opticalexpansion,geonet,dfnet,EPC++,Chen2019geometric,Ranjan2019CC} decompose the task into subproblems. 
The first MSF work, Mono-SF~\cite{monosf}, assumes that piecewise planar surface elements and rigid bodies are enough to approximate autonomous driving scenes. 
Optical-Expansion-based methods~\cite{yang2020upgrading,Ling2023opticalexpansion} upgrade 2D optical flow to 3D scene flow.
Multi-task approaches~\cite{geonet,dfnet,EPC++,Chen2019geometric,Ranjan2019CC} jointly predict depth, optical flow, and camera motion.

These methods produce scene flow that is limited in accuracy due to the strong assumptions made, 
and are surpassed by Self-Mono-SF~\cite{Hur:2020:SSM},
the first MSF method to predict 3D motion vectors directly alongside depth. 
After Self-Mono-SF~\cite{Hur:2020:SSM}'s success, later works devote to improving it. 
Multi-Mono-SF~\cite{Hur:2021:SSM} extends it into multi-frame setting; 
RAFT-MSF~\cite{raftmsf} uses recurrent structure to enhance performance;
EMR-MSF~\cite{emrmsf} applies ego-motion rigidity prior to further boost accuracy. 

Despite the progress made, the methods largely focus on autonomous driving scenarios and only show very limited generalization capabilities to scenes out of that domain~\cite{Hur:2021:SSM}.
Our method shows superior in-domain performance and also zero-shot generalization to in-the-wild scenes.

\subsection{Zero-shot Geometry Estimation and Beyond}
Deep neural networks pre-trained on large-scale datasets are reshaping AI with 
their zero-/few-shot adaptability~\cite{Bommasani2021OnTO,touvron2023llamaopenefficientfoundation}. 
Their success in language and vision has now extended to 3D applications, 
with models like DUSt3R~\cite{dust3r} leveraging large-scale pretraining of cross-view completion~\cite{croco_v2} to enable a wide range of 3D tasks.

Given a pair of uncalibrated stereo images,
DUSt3R represents the 3D scene by pointmaps for both images in left camera's coordinate system with implicit correspondence searching. 
Later, MASt3R~\cite{mast3r} ties DUSt3R to metric space and improves on image matching. These large-scale geometry prediction models have empowered a variety of 3D applications,
including novel view synthesis~\cite{fan2024instantsplat,smart2024splatt3r,chen2024densepointcloudsmatter,yu2024lmgaussianboostsparseview3d},
surface reconstruction~\cite{raj2024spurfiessparsesurfacereconstruction},
incremental reconstruction~\cite{wang2024spann3r} and temporal generation~\cite{liu2024reconxreconstructscenesparse}.

These models are limited to static scenes and cannot reason about motion explicitly. 
MonST3R~\cite{zhang2024monst3r}---a concurrent work---enhances model robustness when dynamic objects are present.
Instead, our method captures the motion field and geometry jointly 
in 3D with one feedforward pass via a single model that is trained end-to-end.
\section{Method}

First, we describe the problem with its inputs and outputs, and our approach to joint feedforward scene flow and geometry prediction~(\cref{sec:overview}). 
Next, we describe our training recipe for in-the-wild generalization (\cref{sec:data-recipe}), including our scale-adaptive optimization (\cref{sec:scale-adaptive_optimization}).
Finally, we explore the impact of different SF parameterizations (\cref{sec:sf-rep}).

\begin{figure}[t]
    \centering
    \includegraphics[width=.4\textwidth]{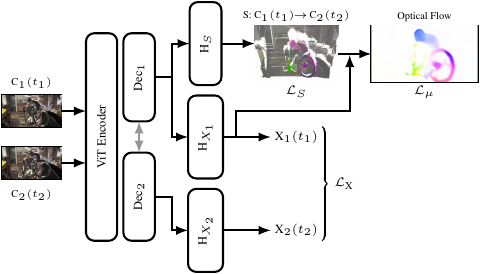}
    \caption{\textbf{Overview.} 
    Our method jointly predicts pointmaps ($X_1$, $X_2$) and scene flow $S$ with an information-sharing ViT 
    backbone followed by three prediction heads ($\text{H}_{X_1}$, $\text{H}_{X_2}$, $\text{H}_S$).  
    }
    \label{fig:overview}
\end{figure}

\subsection{Joint, Generalizable Scene Flow Prediction}
\label{sec:overview}

\paragraph{Problem overview.}

Given two RGB images $I_1, I_2 \in \mathbb{R}^{H \times W \times 3}$ captured by a single, potentially moving camera over time $t$, 
we wish to estimate a 3D mapping that subsumes both 3D camera motion and 3D object motion from the coordinate frame of the first image $C_1$ into the coordinate frame of the second image $C_2$: $(C_1, t_1) \rightarrow (C_2, t_2)$.
We do not have access to camera poses.
This mapping could be parameterized in various different ways, such as start and end points, or motion offset vectors~\cite{Hur:2020:SSM}.

Our method outputs three maps, which combined yield scene flow.
The first two, $\hat{X_1}$ and $\hat{X_2}$ $\in \mathbb{R}^{H \times W \times 3}$, are pointmaps.
They are both represented in the space of camera $C_1$, but at times $t_1$ and $t_2$, respectively.
Recall that pointmaps are a collection of 3D points $\textbf{x}$~\cite{dust3r}.
The third is a scene flow/offset map $\hat{S} \in \mathbb{R}^{H \times W \times 3}$ expressing the mapping $C_1,t_1\rightarrow C_2,t_2$.

\paragraph{Pipeline.} Accurate scene flow estimation is challenging because it requires understanding the geometry of a scene \emph{and} its motion.
We cannot reason about one without considering the other, because after projection to 2D, the two become entangled---and thus scene flow estimation is ill-posed.
Moreover, requiring our approach to generalize to unseen domains further complicates the task.
We design our pipeline with these observations in mind:
First, we use a unified backbone for joint depth and motion prediction.
This enables geometry and motion information sharing.
Second, we adopt CroCoV2~\cite{croco_v2} as our backbone.
This enables direct use of pre-trained weights from state-of-the-art geometry prediction models like DUSt3R~\cite{dust3r} and MASt3R~\cite{mast3r} to bootstrap our model's geometry estimation, which helps improve in-the-wild generalization.

As shown in \cref{fig:overview}, our architecture is a ViT~\cite{dosovitskiy2020vit} model with dual weight-sharing encoder branches ($\text{Enc}_v$), and cross-attended decoders ($\text{Dec}_v$) for each input frame ($v \in \{1, 2\}$). 
Each decoder comprises $B$ blocks ($G_i^v$).
These are followed by specialized DPT-based~\cite{Ranftl2021DPT} heads ($\text{H}_{X_1}$, $\text{H}_{X_2}$, $\text{H}_{S}$) for pointmap and motion offset prediction:
\begin{equation}
    \begin{aligned}
    \hat{X_1} &= \text{H}_{X_1}(G_0^1,\ldots,G_B^1),
    \\
    \hat{X_2} &= \text{H}_{X_2}(G_0^2,\ldots,G_B^2), \text{and}
    \\
    \hat{S} &= \text{H}_{S}\hspace{0.55em}(G_0^1,\ldots,G_B^1).
    \end{aligned}
\end{equation}
Our joint prediction strategy requires a shared internal network representation.
As a result, the 3D prior for geometry prediction can help better handle motion, and simultaneously the motion estimation head H$_S$ can help improve geometry prediction through its learned temporal correspondences.

\subsection{A Recipe to Create Large-scale Scene Flow Data}
\label{sec:data-recipe}

\begin{table*}[t]
    \caption{\textbf{Training datasets in our recipe.} \textit{OF}: Optical Flow, \textit{SF}: Scene Flow. $^*$ For the PointOdyssey dataset, we ignore scenes with smoke.
    }
    \label{tab:data}
    \centering
    \begin{tabular}{llcccrrrr}
    \toprule
    \multirow{2}{*}{Dataset} & \multicolumn{2}{c}{Scene Type} & \multicolumn{2}{c}{Annotation} & \multicolumn{4}{c}{Statistics}\\
    \cmidrule(lr){2-3}  \cmidrule(lr){4-5} \cmidrule(lr){6-9} 
    & Content & Metric & OF & SF & \# Cam & \# Frames &\# Scenes & \textit{W}$\times$\textit{H} \\ 
    \midrule
    SHIFT~\cite{shift2022} & Driving  & \checkmark & \text{\sffamily x} & \text{\sffamily x} & 6 & 150k & 3000 & 1280$\times$800 \\ 
    Dynamic Replica~\cite{karaev2023dynamicstereo} & Indoor & \checkmark & \checkmark & \text{\sffamily x} & 2 & 72k & 484 & 1280$\times$720 \\ 
    Virtual KITTI 2~\cite{cabon2020vkitti2} & Driving  & \checkmark & \checkmark & \checkmark & 2 & 35k & 40 & 1242$\times$375 \\ 
    MOVi-F~\cite{greff2021kubric} & Kubric-style & \text{\sffamily x} & \checkmark & \text{\sffamily x} & 1 & 137k & 5736 & 512$\times$512 \\ 
    PointOdyssey$^*$~\cite{zheng2023point} & Indoor \& Kubric-style &  \checkmark & \text{\sffamily x} & \text{\sffamily x} & 1 & 100k & 39 & 960$\times$540 \\ 
    Spring~\cite{Mehl2023_Spring} & Sintel-style animation & \text{\sffamily x} & \checkmark & \checkmark & 2 & 4k & 30 & 1920$\times$1080 \\ 
    \bottomrule
    \end{tabular}
\end{table*}

If we are to tackle an ill-posed problem with a data-driven approach, we need large and diverse data with high-quality scene flow annotations. 
However, scene flow is more difficult to measure than other low-level properties like depth and optical flow, as it depends upon both geometry and motion.
The resulting scarcity of scene flow datasets has been a major hinderance to the progress of learning-based scene flow prediction methods.
Most prior monocular scene flow works~\cite{Hur:2020:SSM,Hur:2021:SSM,raftmsf,emrmsf} train on a specific dataset (\eg, KITTI raw~\cite{Geiger2013IJRR}), 
but each is limited in size and diversity. Some synthetic datasets exist, but these are narrow in their domains. Overall, generalization is restricted by scarce data.

To improve generalization by covering as diverse scene content as possible, we augment four additional dynamic scene datasets for scene flow estimation:
SHIFT~\cite{shift2022}, DynamicReplica~\cite{karaev2023dynamicstereo}, MOVi-F~\cite{greff2021kubric} and PointOdyssey~\cite{zheng2023point}.

With the multiple cameras in the data, the full data recipe~(\cref{tab:data}) contains over 1M samples across indoor and outdoor settings, and with real and synthetic dynamic scenarios.
This helps to provide accurate and generalizable scene flow prediction.

\paragraph{The data.} Each sample consists of data and corresponding annotations for two consecutive timestamps in a video.
Specifically, a sample includes two RGB maps, $I_1$ and $I_2$, two depthmaps, $D_1$ and $D_2$, and the corresponding validity maps, $M_{D_1}$ and $M_{D_2}$, which indicate whether a depth value is available for each pixel in the depthmap.
It also includes forward optical flow and validity map, $\mu$ and $M_{\mu}$, as well as forward scene flow and validity map, $S$ and $M_S$.
Finally, each sample provides intrinsics $K$, relative transformation from $C_1$ to $C_2$, $[R|t]$, and a metric scale indicator $\mathbf{1}_{\text{m}}$ that is `1' if the depth information is metric.

\paragraph{Augmenting datasets for scene flow prediction.}

Not all datasets provide annotations for scene flow $S$ or optical flow $\mu$. 
When $\mu$ is not provided, we use RAFT~\cite{Teedraft2020} to generate psuedo ground truth, and assume that all pixels are valid.
When $S$ is not given, we compute pseudo ground truth from the depthmaps, camera poses, and optical flow.
To achieve that, we first unproject each pixel $\textbf{p}_1$ in $I_1$ and find the corresponding 3D point
\begin{equation}
    \textbf{x}_1 = \pi^{-1}(D_1[\textbf{p}_1], K),
\end{equation}
where $\pi^{-1}(\cdot,\cdot)$ is the inverse of the camera projection operation.
Next, we find the corresponding pixel $\textbf{p}_2$ to $\textbf{p}_1$ via pixel coordinates $\textbf{p}_1$ and optical flow $\mu$ and unproject it to produce a second 3D point $\textbf{x}_2 \in X_2$ in $C_2$:
\begin{align}
        \textbf{p}_2 &= \textbf{p}_1 + \mu[\textbf{p}_1]
        \\
        \textbf{x}_2 &= \pi^{-1}(D_2[\textbf{p}_2], K).
\end{align}
Finally, we can compute the scene flow pseudo ground truth $\textbf{x}_2 - \textbf{x}_1$.
Note that this uplifting of optical flow and depth to scene flow requires both $\textbf{p}_1, \textbf{p}_2$ to have valid depth and for $\textbf{p}_1$ to have valid optical flow. 

Scene flow is undefined when a 3D point becomes occluded or disoccluded. Even when visibility is not an issue, ground truth data can be in error. 
To cope with this, we check depth and optical flow validity, as well as occlusion/disocclusion.
We follow Meister~\etal~\cite{Meister:2018:UUL} to obtain an occlusion indicator $\textbf{1}_{\text{cyc}}(\textbf{p}_1, \textbf{p}_2)$ through a forward-backward consistency check. 
Scene flow validity is: 
\begin{equation}
    M_S[\textbf{p}_1] = M_\mu[\textbf{p}_1] \land M_{D_1}[\textbf{p}_1] \land M_{D_2}[\textbf{p}_2] \land \textbf{1}_{\text{cyc}}(\textbf{p}_1, \textbf{p}_2).
\end{equation}
We compute ground truth pointmaps by unprojecting depth maps to camera coordinates. 
We transform $D_2$ into $C_1$ via the relative camera pose:
\begin{equation}
    \begin{aligned}
        X_1 &= \pi^{-1}(D_1, K)
        \\
        X_2 &= [R|t]^{-1} \pi^{-1}(D_2, K).
    \end{aligned}
\end{equation}

\subsection{Scale-adaptive Optimization}
\label{sec:scale-adaptive_optimization}

\begin{figure}[t]
    \centering
    \includegraphics[width=.45\textwidth]{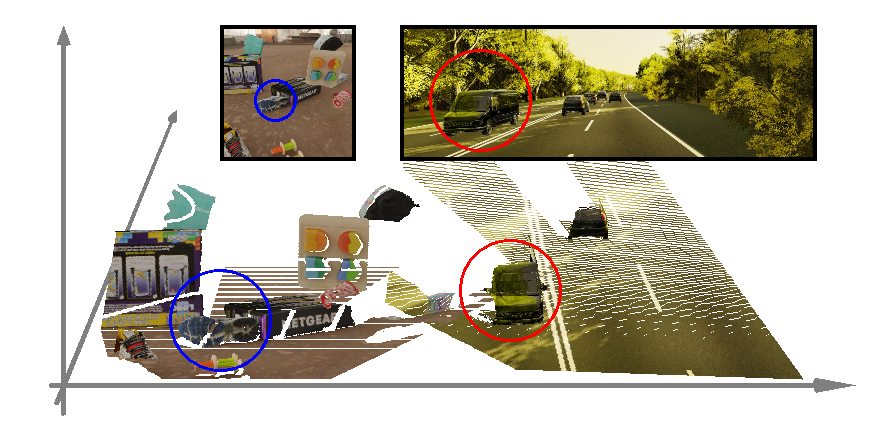}
    \caption{\textbf{Different datasets have different scales.} Here we show a frame from MOVi-F~\cite{greff2021kubric}, which is in relative scale, and one from Virtual KITTI~\cite{cabon2020vkitti2}, which is in metric units. We need to account for this while training for both geometry and motion.}
    \label{fig:data-scale}
\end{figure}

While a large collection of datasets is critical for achieving zero-shot generalization, combining data from diverse datasets, potentially not even designed for our task, is not without cost.
One major problem is scale-inconsistency: some data are metric, others are relative~(\cref{fig:data-scale}). 
MiDaS~\cite{Ranftl2022} uses diverse dataset for relative depth estimation and introduced a scale-invariant loss to deal with this problem. 
We propose a similar solution for scene flow that does not require us to give up metric estimation. 
Instead, we show how to leverage both metric and non-metric data \textit{for metric SF} when properly trained.

Our method's SF and pointmap outputs are both metric. 
When $\mathbf{1}_{\text{m}}=1$, there is nothing to adapt.
When $\mathbf{1}_{\text{m}}=0$, 
we normalize the ground truth, predicted geometry, and scene flow before computing the loss.
Specifically, we compute scale factors for the predicted pointmaps and the ground truth pointmaps separately, 
as the mean distance from $C_1$ origin to all valid points:

\begin{align}
    \hat{z} &= \frac{\sum_{v \in \{1,2\}} \sum_{i} M_{D_v}[i] \cdot \| \hat{X_v}[i] \|_2}{\sum_{v \in \{1,2\}} \sum_{i} M_{D_v}[i]}
    \\
    z &= \frac{\sum_{v \in \{1,2\}} \sum_{i} M_{D_v}[i] \cdot \| X_v[i] \|_2}{\sum_{v \in \{1,2\}} \sum_{i} M_{D_v}[i]},
\end{align}
where $\hat{\cdot}$ indicates predictions.
As scene flow and geometry must share a scale, we can use $z$/$\hat{z}$ to normalize both:
\begin{align}
    \hat{X}^{*}_v = \hat{X_v} / \hat{z}, \quad
    \hat{S}^* = \hat{S} / \hat{z} \\
    X_v^* = X_v / z, \quad
    S^* = S / z,
\end{align}
where $^*$ indicates a normalized value. 
For non-metric data samples, we use normalized pointmaps and scene flow in the loss.
Thus, we have a pointmap loss $\mathcal{L}_X$ and a motion offset map loss $\mathcal{L}_S$:
\begin{align}
    \resizebox{0.9\linewidth}{!}{$
        \mathcal{L}_X = \sum_{v \in \{1,2\}} \sum_{i} M_{D_v}[i] \ \Big(\mathbf{1}_{\text{m}} \|X_v[i] - \hat{X{_v}}[i]\|_1 
    $} \nonumber \\ 
    \resizebox{0.5\linewidth}{!}{$
        + (1-\mathbf{1}_{\text{m}}) \|X_v^{*}[i] - \hat{X_v^{*}}[i]\|_1 \Big) 
    $} \\
    \resizebox{0.6\linewidth}{!}{$
        \mathcal{L}_S = \sum_{i} \ M_S[i] \ \Big(\mathbf{1}_{\text{m}} \|S[i] - \hat{S}[i]\|_1 
    $} \nonumber \\
    \resizebox{0.49\linewidth}{!}{$
    + (1-\mathbf{1}_{\text{m}}) \|S^*[i] - \hat{S^*}[i]\|_1 \Big). 
    $}%
\end{align}
Note that we only compute the loss for pixels with valid ground truth. 

\paragraph{Optical flow supervision.}

While our method predicts geometry and motion with a shared encoder-decoder, $X$ and $S$ are still predicted independently, with no interaction in $\mathcal{L}_X$ nor $\mathcal{L}_S$.
We use optical flow to relate the two in order to improve consistency.
The projection of scene flow to optical flow $\hat{\mu}$ relies on $X$, $S$, and $K$, and so a loss over $\hat{\mu}$ encourages the network to produce $X$ and $S$ that are consistent and agree with ground truth intrinsics $K$ during training.

We compute optical flow as the 2D image-space offset between pointmap projections before and after scene flow:
\begin{align}
    \hat{\mu} &= \pi(\hat{X_1}+\hat{S}, K) - \pi(\hat{X_1}, K) \\
    \mathcal{L}_{\mu} &= \sum_{i}  M_S[i] \ \|\mu[i] - \hat{\mu}[i]\|_1
\end{align}
Again, we only compute the loss for valid pixels.
As optical flow is a projected measure, it is independent of the scaling issue and so no normalization is required.

\paragraph{Final loss.} 
The final optimization target is a weighted mixture of 3D and 2D supervision~(\cref{fig:overview}):

\begin{equation}
    \mathcal{L} = \mathcal{L}_X + \mathcal{L}_S + 0.1 \mathcal{L}_{\mu}
\end{equation}

\subsection{Scene Flow Parameterization Choice}
\label{sec:sf-rep}

\begin{figure}[t]
    \centering
    \includegraphics[width=.4\textwidth]{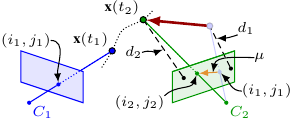}
    \caption{\textbf{SF parameterizations.} Given cameras $C_1$ and $C_2$ capturing a 3D point at two times, $t_1$ and $t_2$, \fpendpoint expresses the displacement as the coordinates of $\textbf{x}(t_2)$ in $C_2$, \fpdepthchangeopflow as $d_2-d_1$ and optical flow $\mu$, and  {\color{darkred}\fpcamspaceoffsets (ours)} as the 3D offset between the two points, in {\color{darkred} red} in the diagram. Note: The faded blue point to the right has not been \emph{transformed} from $C_1$ into $C_2$; it is visualized here with exactly the same numeric coordinates as it has in $C_1$.}
    \label{fig:flow-param}
\end{figure}

In the literature, scene flow has been parameterized in multiple ways, including as offsets as we have so far discussed~\cite{Hur:2020:SSM}, but also as end points directly, or as motion-in-depth + optical flow~\cite{yang2020upgrading,Badki2021ttc}, see Fig.~\ref{fig:flow-param}.
We explore the three parameterizations for a generalizable monocular scene flow model.
To isolate their contribution to the quality of SF estimation, we define three models. 
Each variant shares the same network architecture~(\cref{sec:overview}) and only differs in how we interpret the $3$-dim output of the scene flow head, H$_S$.

\paragraph{Camera-space 3D offsets (\fpcamspaceoffsets).}
This is the parameterization we have discussed thus far,
where predicted scene flow $\hat{S}$ parameterizes both camera motion and object motion as a 3D vector field of offsets. 

\paragraph{Depth change and optical flow (\fpdepthchangeopflow).} 
Typical ground truth scene flow annotations are provided as maps of (disparity/depth in $C_1$ at $t_1$, disparity/depth in $C_2$ at $t_2$, optical flow from $C_1,t_1$ to $C_2,t_2$).
For this parameterization, we assume that optical flow $\hat{\mu}$ is predicted by the first two dimensions of the scene flow head output, and $\Delta{D}$ by the third dimension. 
To penalize $\mathcal{L}_\mu$, we no longer need to project scene flow as optical flow is predicted directly. But, to penalize $\mathcal{L}_S$, we need to compose $\hat{S}$ from $\hat{\mu}$, $\Delta{D}$, $\hat{X_1}$, $K$.
First, we add the depth change to the z value of the pointmap $\hat{X_{1}}_z$ to obtain the depth after motion in camera space $C_2$.
Then, for each pixel $\textbf{p}$ in $I_1$, $\hat{\mu}$ tells us which pixel in $I_2$ it corresponds to.
Last, we can compute the 3D position with depth and ray direction, and SF is computed by subtraction:
\begin{equation}
    \begin{split}
        D_2 &= \hat{X_{1}}_z + \Delta{D} \\
        \hat{S}[\textbf{p}] &= \pi^{-1}(D_2, K)[\textbf{p} + \hat{\mu}[\textbf{p}]] - \hat{X_1}[\textbf{p}].
    \end{split}
\end{equation}

\paragraph{End point (\fpendpoint).}
As the pretrained weights from models~\cite{dust3r,mast3r} help the depth estimation side of the problem, 
and since these models all predict pointmaps instead of offset maps,
we can directly use head H$_S$ to predict for $I_1$ the corresponding 3D point positions in camera space $C_2$ at $t_2$. 
Then, from H$_S$'s prediction, $\hat{S}$, and from H$_{X_1}$'s prediction, $\hat{X_1}$, we can easily compute scene flow as: $\hat{S} - \hat{X_1}$.

\section{Experiments}
Fig.~\ref{fig:teaser} shows the zero-shot performance of our approach across different scenarios, indoor and outdoor scenes, and over a broad range of scene depths.
The animations show interpolation results between the cameras and timestamps of the two input images.
They show a different perspective from the input cameras, where both the depth and scene flow are consistent with each other---and correct.
We also do extensive quantitative evaluation and ablation studies across several datasets, normalization strategies and training schemes, test zero-shot generalization abilities, as well as the impact of different scene flow parametrizations, all of which we describe in detail in the following sections.
However, the key takeaway is much simpler: 
\emph{The design choices we make in terms of architecture, training recipe, and output parameterization, result in a method that can estimate accurate scene flow in the wild for the first time. }

\setlength{\heavyrulewidth}{1.5pt}    
\definecolor{lightgray}{gray}{0.9}

\subsection{Experimental Setting}
\label{exp:setting}

\paragraph{Datasets.}
We evaluate our method's scene flow and geometry estimation capabilities on datasets spanning synthetic and real-world scenarios with 3D flow annotations: 
KITTI Scene Flow~\cite{Menze2015ObjectSF}, 
Spring~\cite{Mehl2023_Spring}, and
VKITTI2~\cite{cabon2020vkitti2}. 
Notably, Spring and VKITTI2 currently lack standardized monocular scene flow evaluation protocols.
Our evaluation protocol uses only monocular (left camera) sequences (see supplemental material).

\paragraph{Implementation details.}

We initialize ViT Encoder, Dec$_1$, Dec$_2$, H$_{X_1}$ and H$_{X_2}$'s parameters 
with MASt3R's~\cite{mast3r} pretrained weights. 
If the \textbf{EP} parameterization is used, H$_S$ is also initialized as H$_{X_1}$.

\paragraph{Comparisons.} We compare with methods from two groups.
For the first group, \underline{SF}, we compare with top-performing scene flow estimation methods on the KITTI scene-flow benchmark~\cite{Menze2015ObjectSF} that have code available. 
For Self-Mono-SF~\cite{Hur:2020:SSM}, we use the checkpoint trained on the KITTI Split~\cite{Godard_2017}. 
We also include OpticalExpansion~\cite{yang2020upgrading}, 
an algorithm to augment depth estimation models with scene flow prediction function in a generalizable fashion.
For OpticalExpansion, we use MASt3R \cite{mast3r} as the depth model, and use the checkpoint trained on \textit{Driving}~\cite{mayer2016} 
to prevent leakage of the KITTI test set during training.

For the second group, \underline{$D$+OF} (note, different from \fpdepthchangeopflow which is a scene parameterization), we build baselines by combining depth and optical flow estimation.
We first compute the depthmaps for both frames, and then use optical flow and camera intrinsics to estimate scene flow.
For depth estimation, we select DUSt3R~\cite{dust3r}, MASt3R~\cite{mast3r}, and DepthAnythingV2~\cite{depth_anything_v2} (metric outdoor version).
For optical flow estimation, we use RAFT~\cite{Teedraft2020}.

\paragraph{Evaluation metrics.} We test models on both Monocular Scene Flow Estimation and Depth Estimation. 
For scene flow evaluation, we follow tradition \cite{HPLFlowNet,liu2019flownet3d}: 
End Point Error (EPE), Accuracy Strict (AccS), Accuracy Relax (AccR), and Outliers (Out). 
For depth estimation, we report both relative depth metrics AbsR-r and $\delta_1$-r,  
 and metric depth metrics AbsR-m and $\delta_1$-m.

\subsection{In/Out-Of-Domain Comparison}
\label{exp:major}

\begin{figure*}[t]
    \centering
    \setlength{\tabcolsep}{1pt}
    \def\w{0.2}
    \def\c{0.1}
    \definecolor{nvidiagreen}{RGB}{118, 185, 0}
    \definecolor{warmred}{RGB}{255, 99, 71}
    \definecolor{brightyellow}{RGB}{255, 215, 0}
    \definecolor{lightgray}{RGB}{211, 211, 211}

    \raisebox{0pt}{\(\color{lightgray}\blacksquare\)} Both methods correct \quad
    \raisebox{0pt}{\(\color{nvidiagreen}\blacksquare\)} \textit{Ours} is correct, Peer is not \quad
    \raisebox{0pt}{\(\color{brightyellow}\blacksquare\)} Peer is correct, \textit{Ours} is not \quad
    \raisebox{0pt}{\(\color{warmred}\blacksquare\)} Both failed

    \vspace{5mm}
    
    \begin{tabular}{cc@{\hspace{1mm}}cccc}
        & Input Images & & vs.~Self-Mono-SF~\cite{Hur:2020:SSM} & vs.~OpticalExpansion~\cite{yang2020upgrading} & vs.~MASt3R~\cite{mast3r}\\
        \multirow{2}{*}{\parbox{4mm}{\vspace{-3mm}\rotatebox[origin=c]{90}{KITTI~\cite{Menze2015ObjectSF}}}} & \includegraphics[width=\w\textwidth]{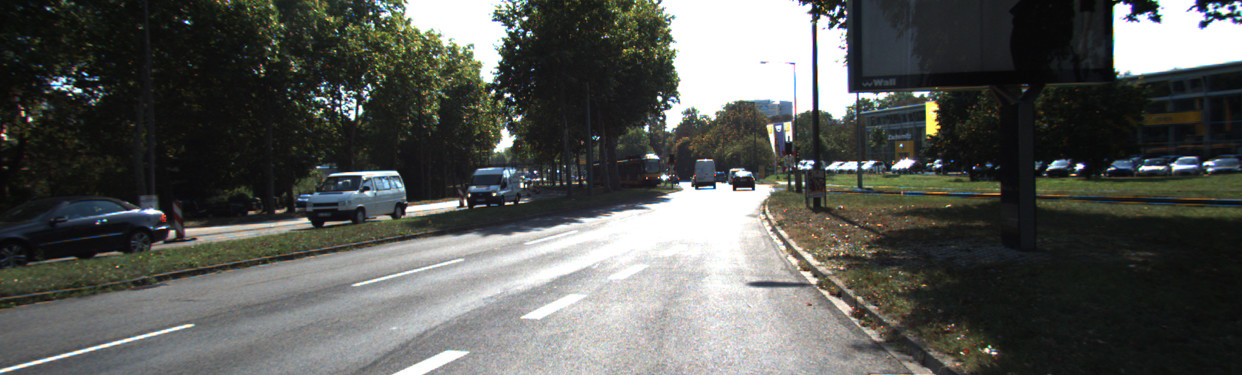} &
        \raisebox{0.1\height}{\rotatebox{90}{$AccR$}} &
        \includegraphics[width=\w\textwidth]{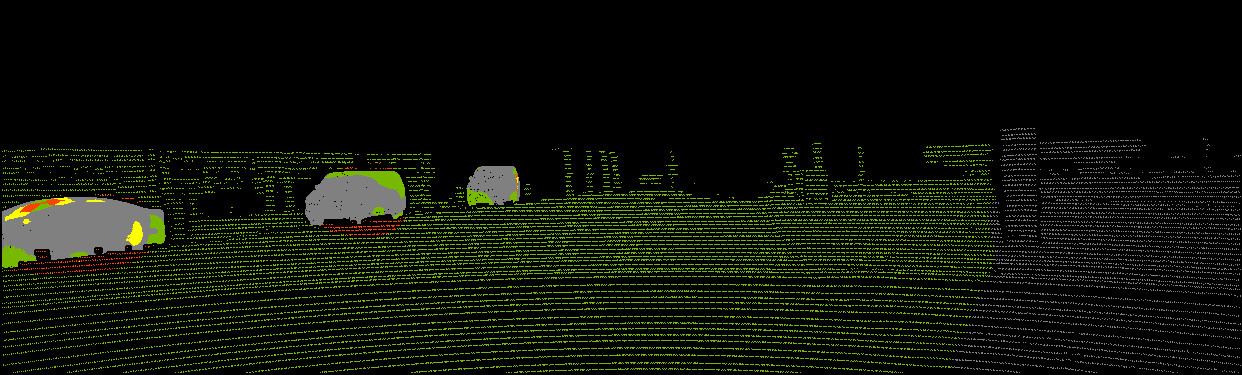} &
        \includegraphics[width=\w\textwidth]{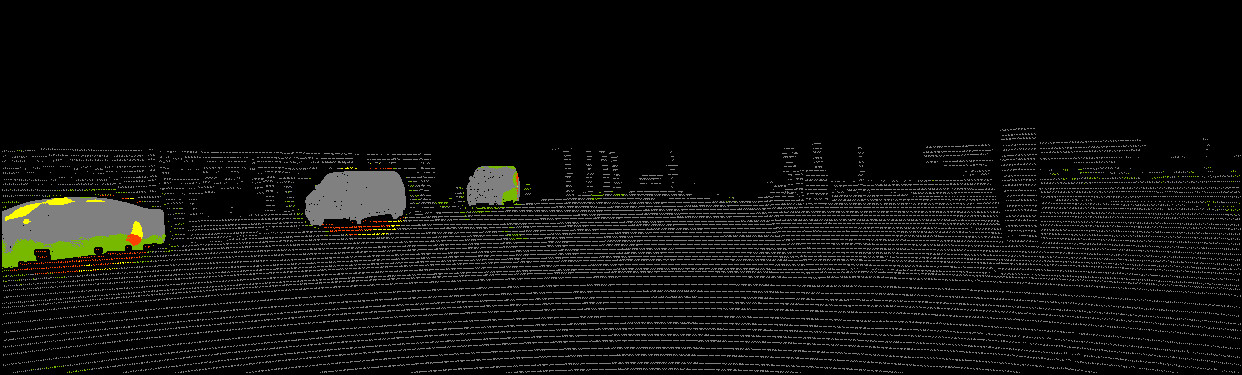} &
        \includegraphics[width=\w\textwidth]{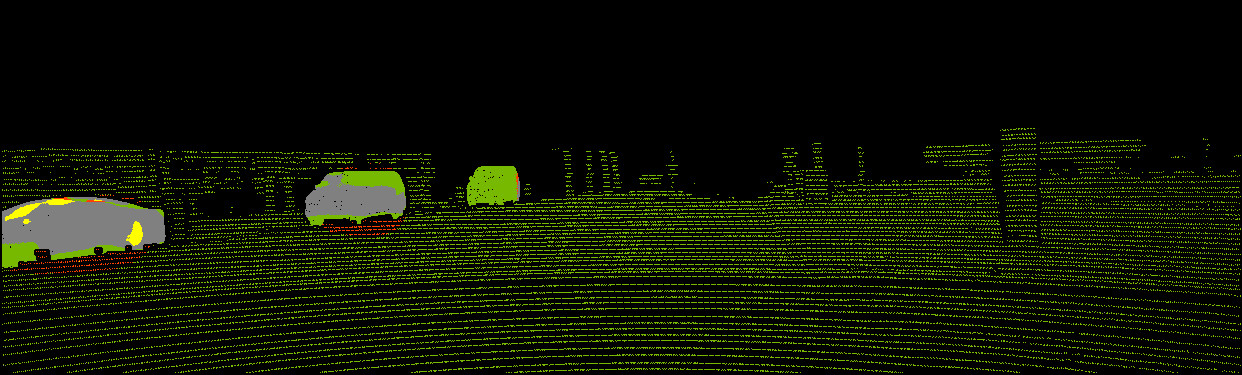}  \\
        & \includegraphics[width=\w\textwidth]{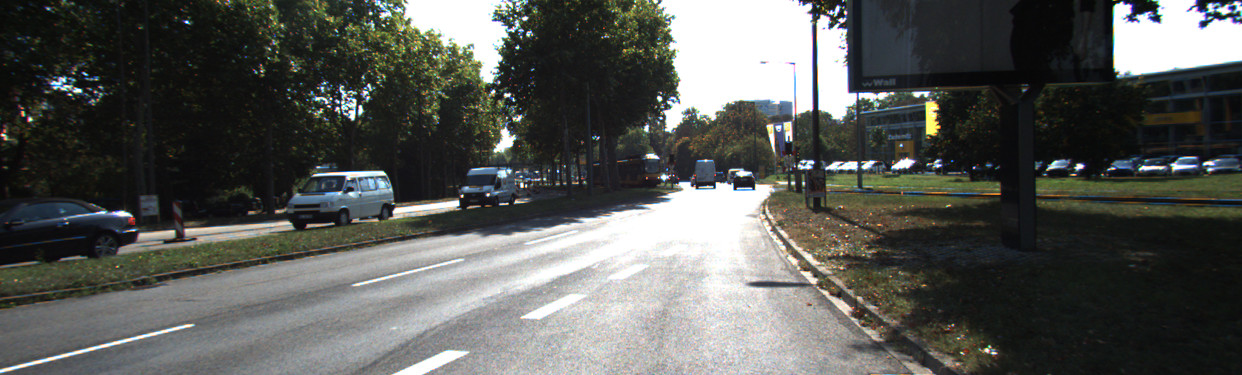} &
        \raisebox{.5\height}{\rotatebox{90}{$\delta_1$-r}} &
        \includegraphics[width=\w\textwidth]{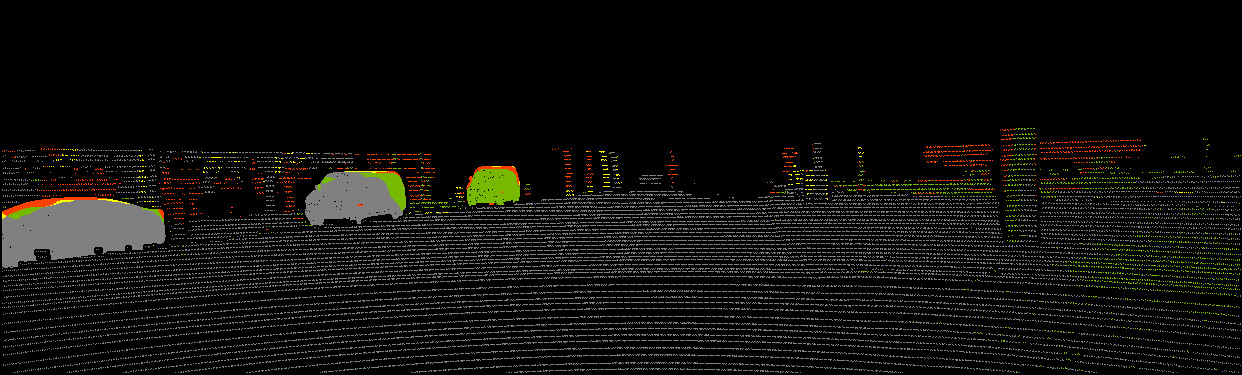} &
        \includegraphics[width=\w\textwidth]{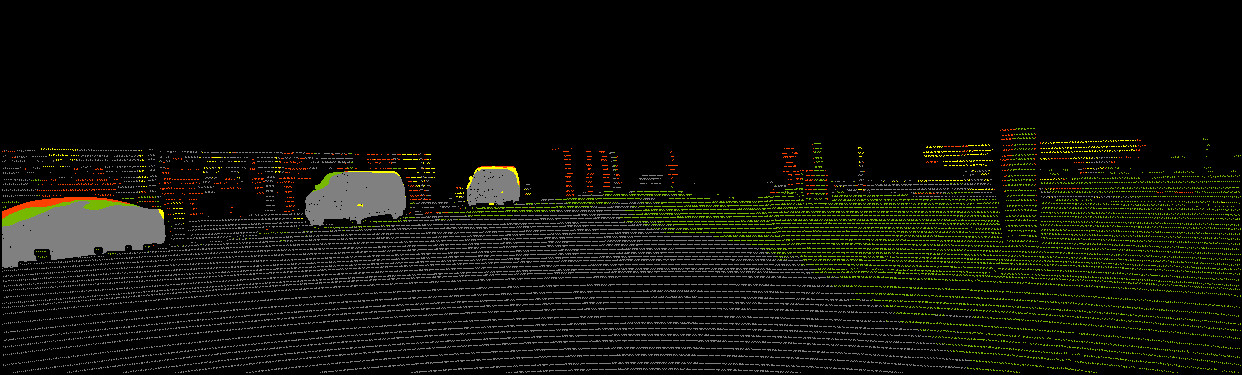} &
        \includegraphics[width=\w\textwidth]{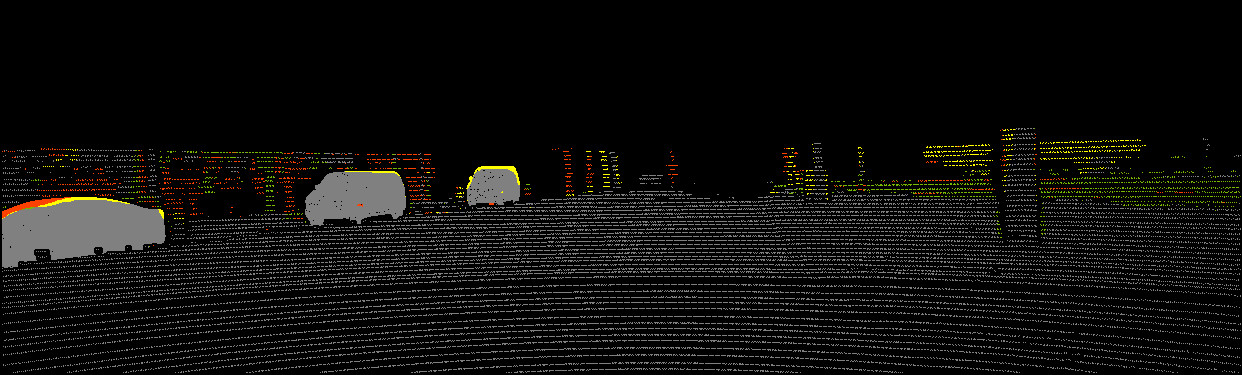}  \\

        \multirow{2}{*}{\parbox{4mm}{\vspace{-6mm}\rotatebox[origin=c]{90}{VKITTI2~\cite{cabon2020vkitti2}}}} & \includegraphics[width=\w\textwidth]{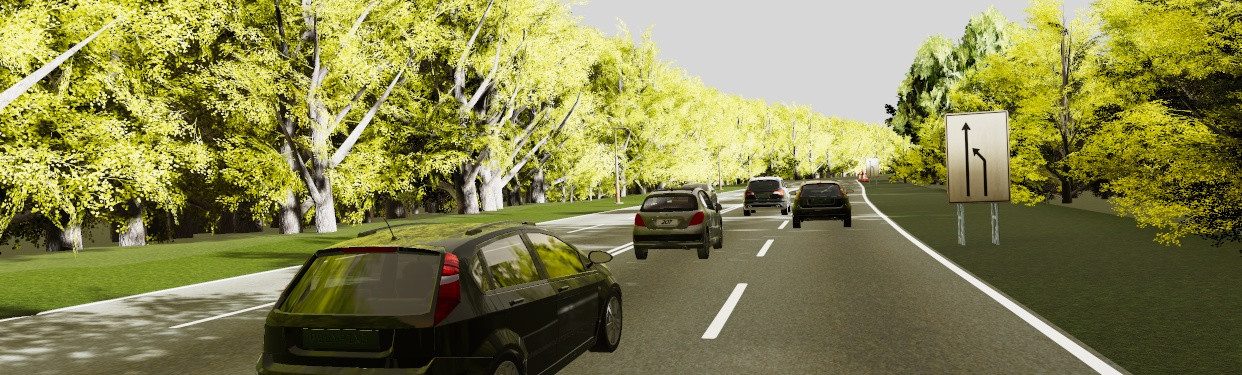} &
        \raisebox{0.1\height}{\rotatebox{90}{$AccR$}} &
        \includegraphics[width=\w\textwidth]{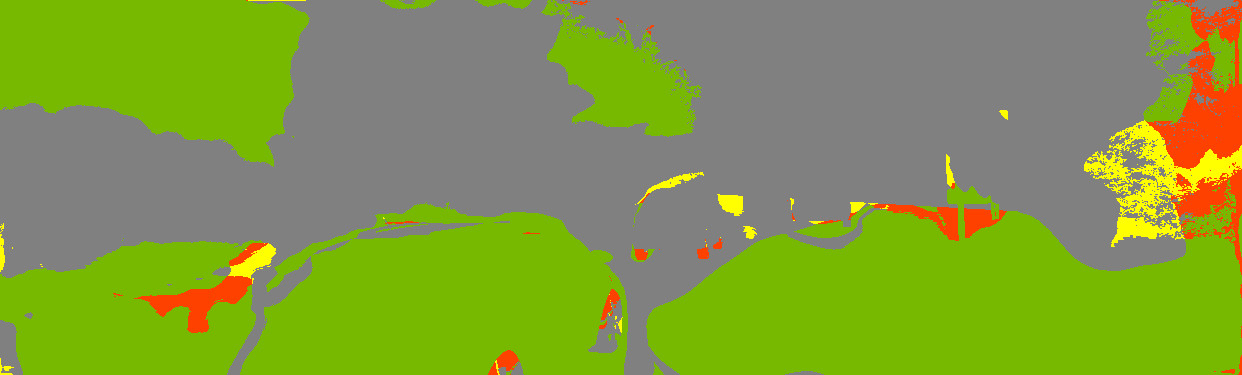} &
        \includegraphics[width=\w\textwidth]{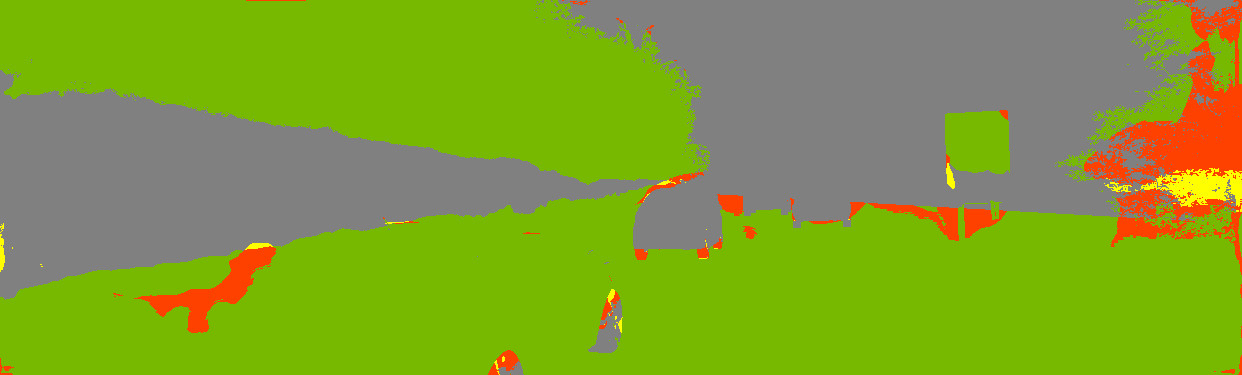} &
        \includegraphics[width=\w\textwidth]{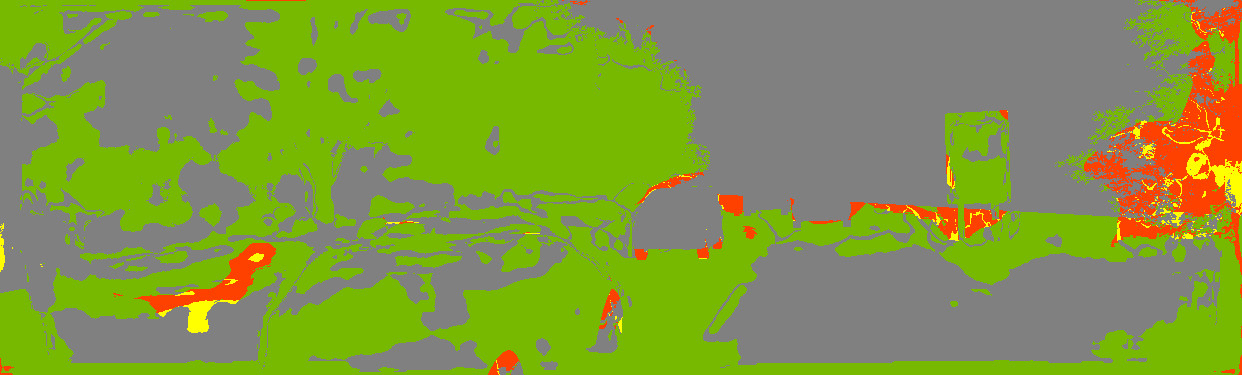}  \\
        & \includegraphics[width=\w\textwidth]{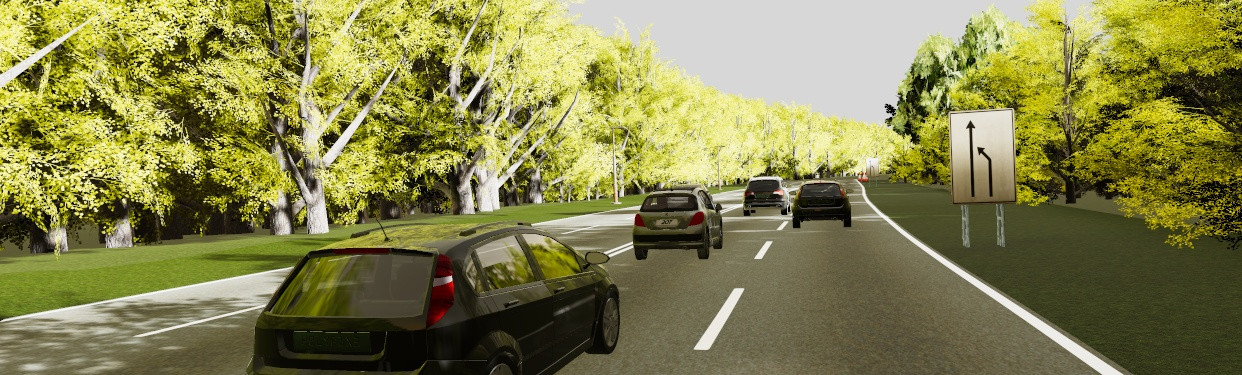} &
        \raisebox{.5\height}{\rotatebox{90}{$\delta_1$-r}} &
        \includegraphics[width=\w\textwidth]{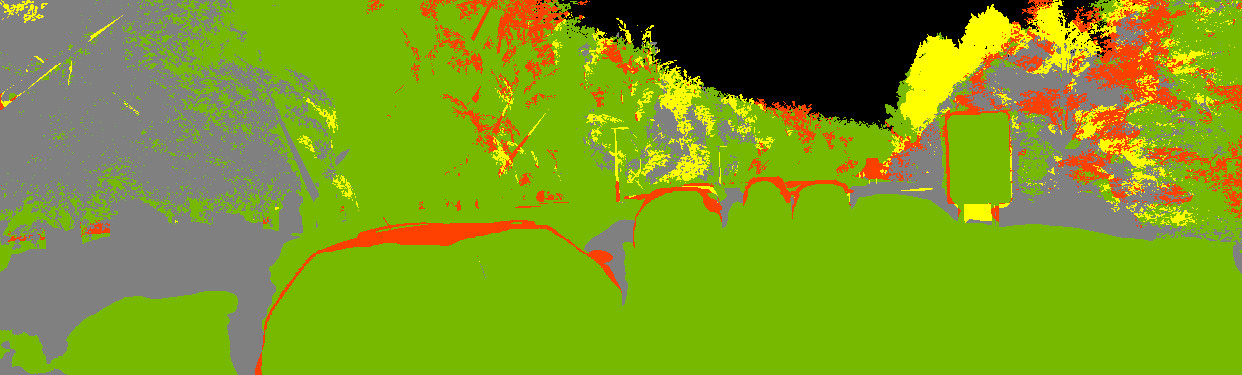} &
        \includegraphics[width=\w\textwidth]{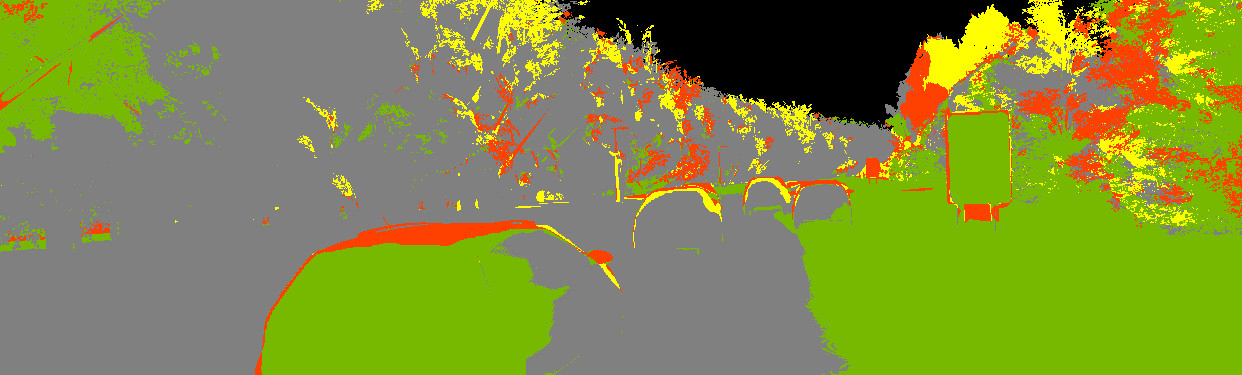} &
        \includegraphics[width=\w\textwidth]{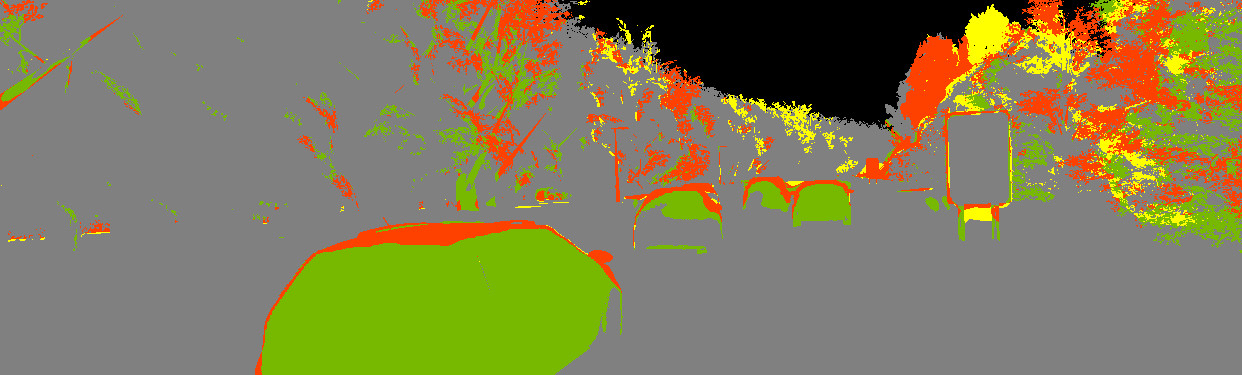}  \\

        \multirow{2}{*}{\parbox{4mm}{\vspace{-3mm}\rotatebox[origin=c]{90}{Spring~\cite{Mehl2023_Spring}}}} & \includegraphics[width=\w\textwidth,trim={0 100pt 0 0},clip]{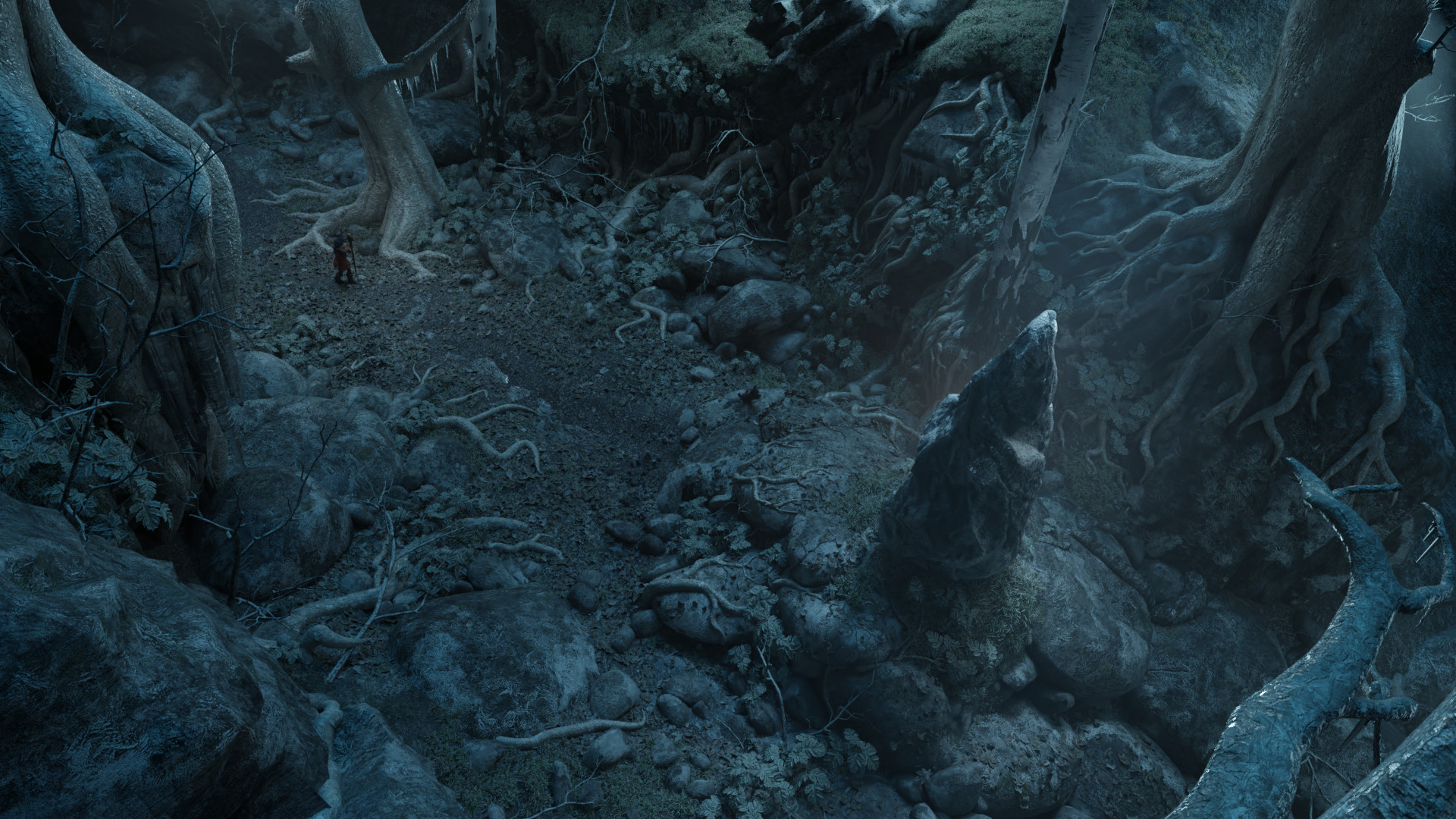} &
        \raisebox{0.1\height}{\rotatebox{90}{$AccR$}} &
        \includegraphics[width=\w\textwidth,trim={0 100pt 0 0},clip]{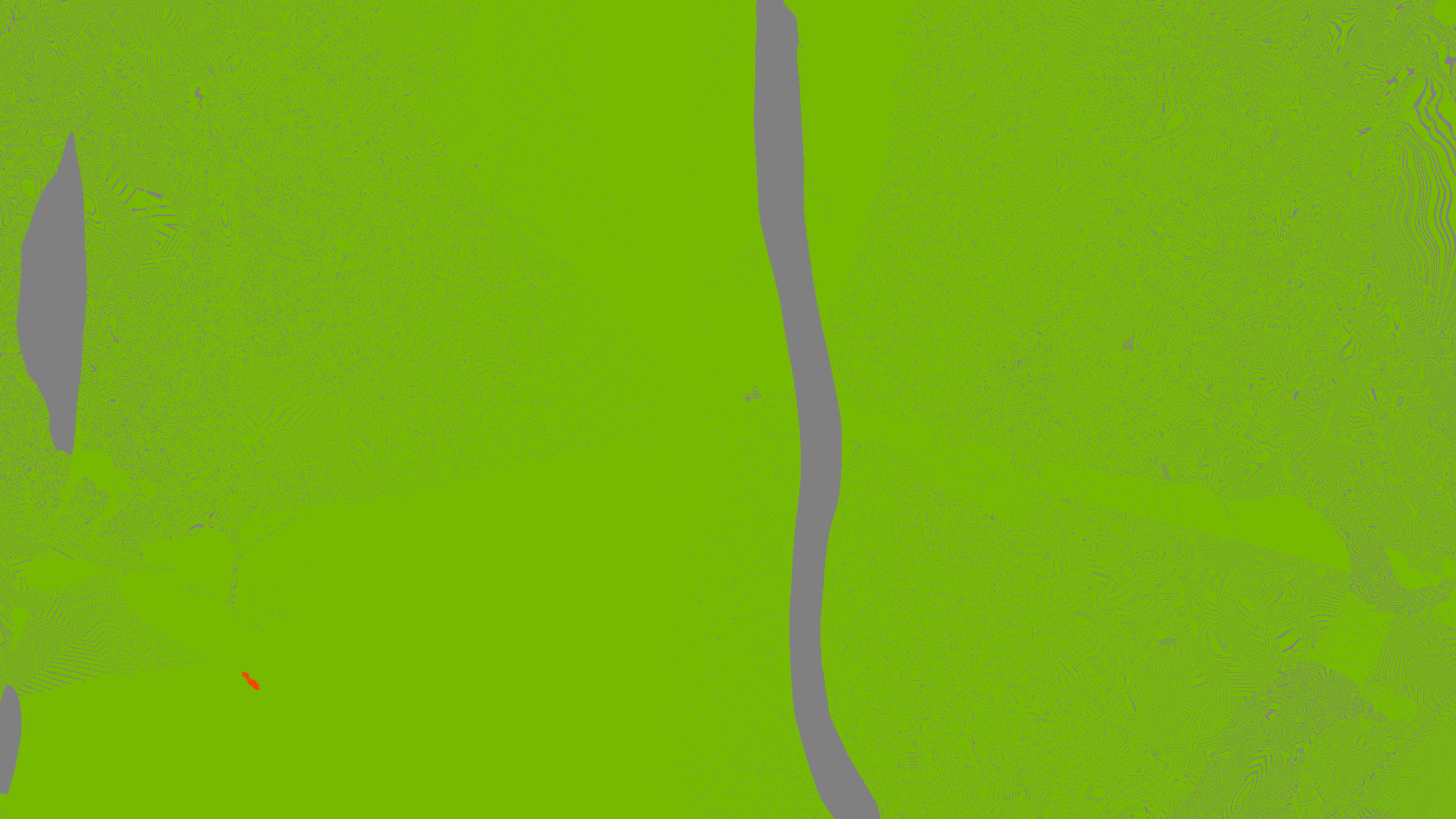} &
        \includegraphics[width=\w\textwidth,trim={0 100pt 0 0},clip]{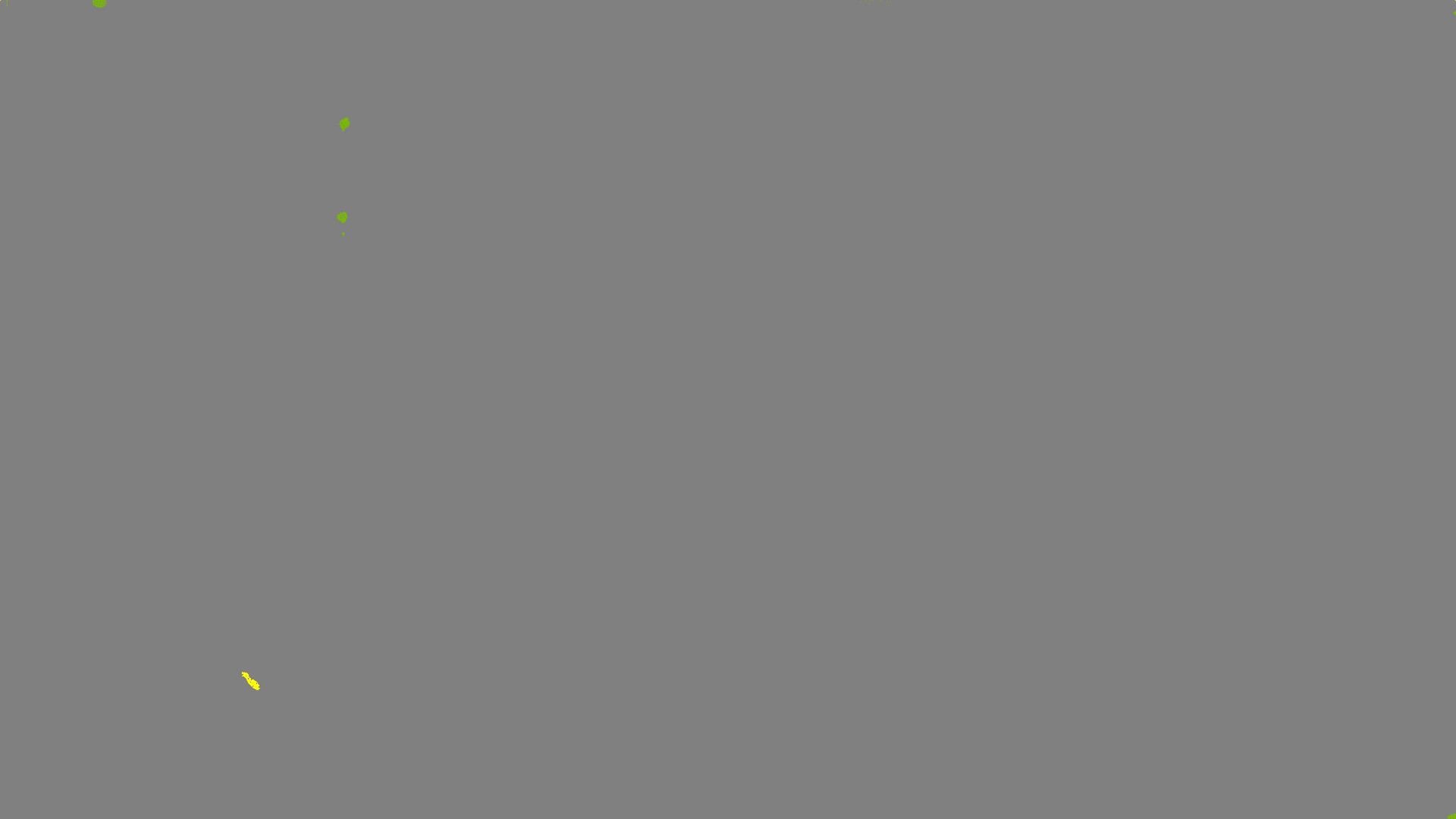} &
        \includegraphics[width=\w\textwidth,trim={0 100pt 0 0},clip]{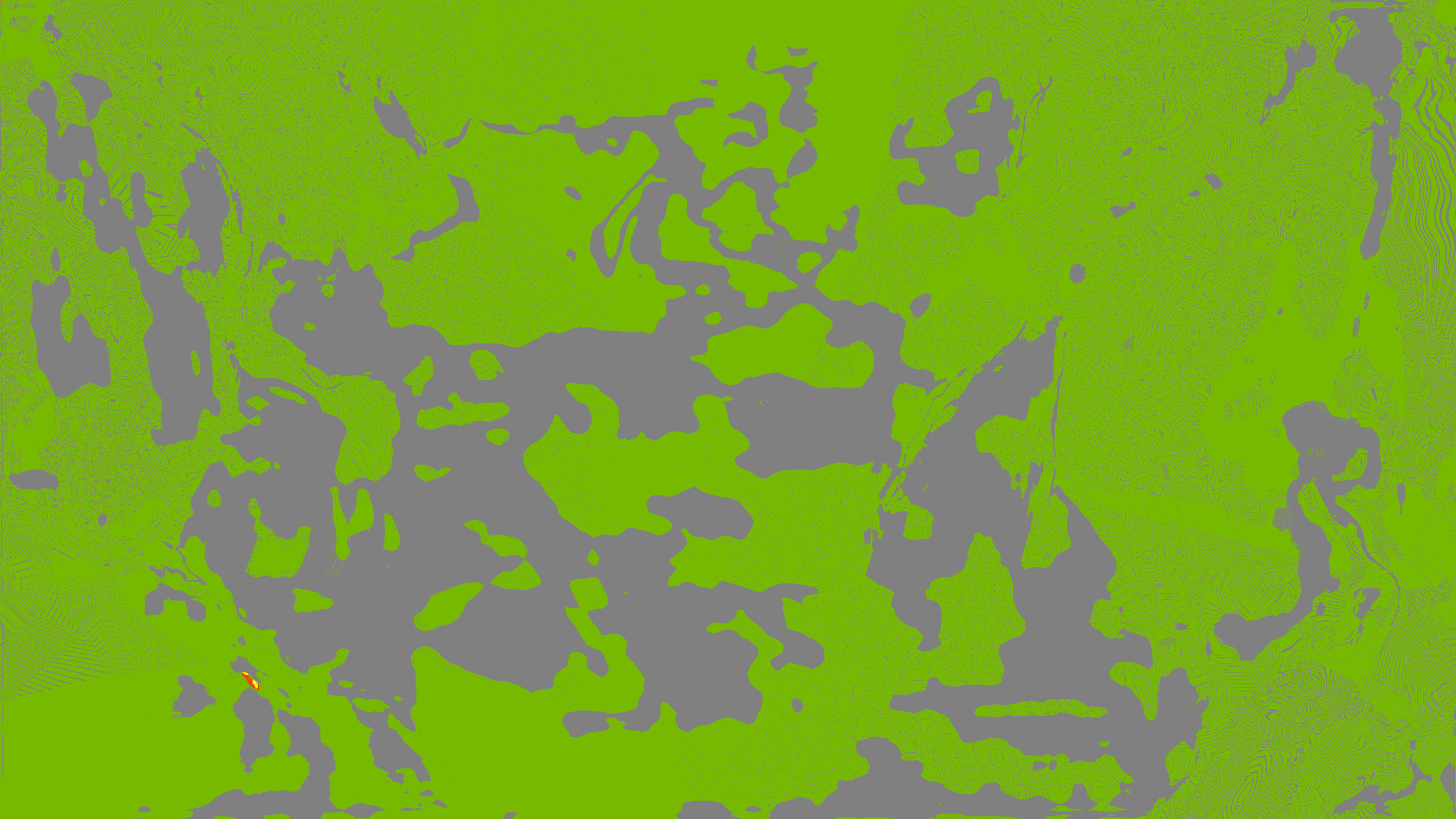}  \\
        & \includegraphics[width=\w\textwidth,trim={0 100pt 0 0},clip]{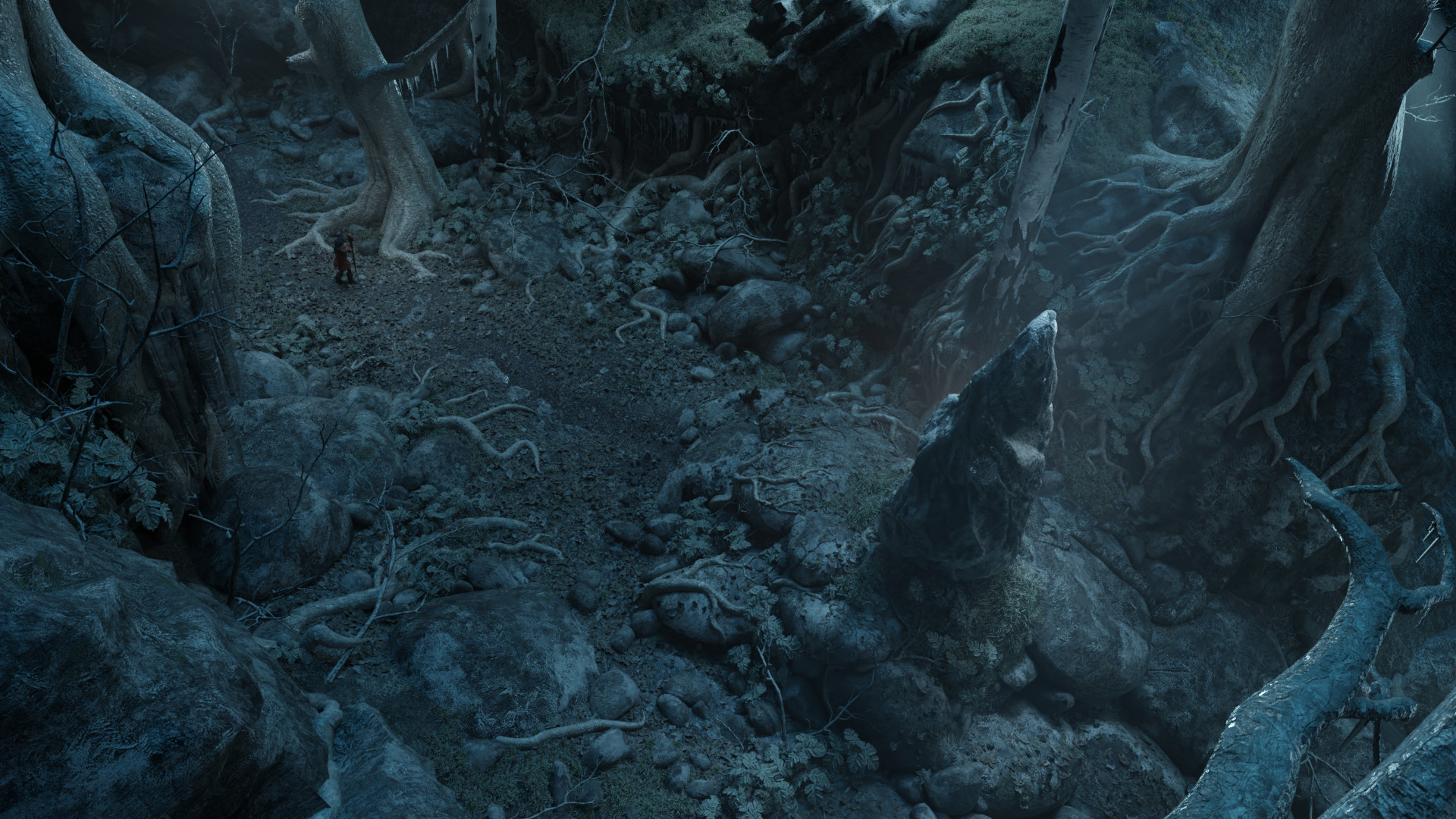} &
        \raisebox{.5\height}{\rotatebox{90}{$\delta_1$-r}} &
        \includegraphics[width=\w\textwidth,trim={0 100pt 0 0},clip]{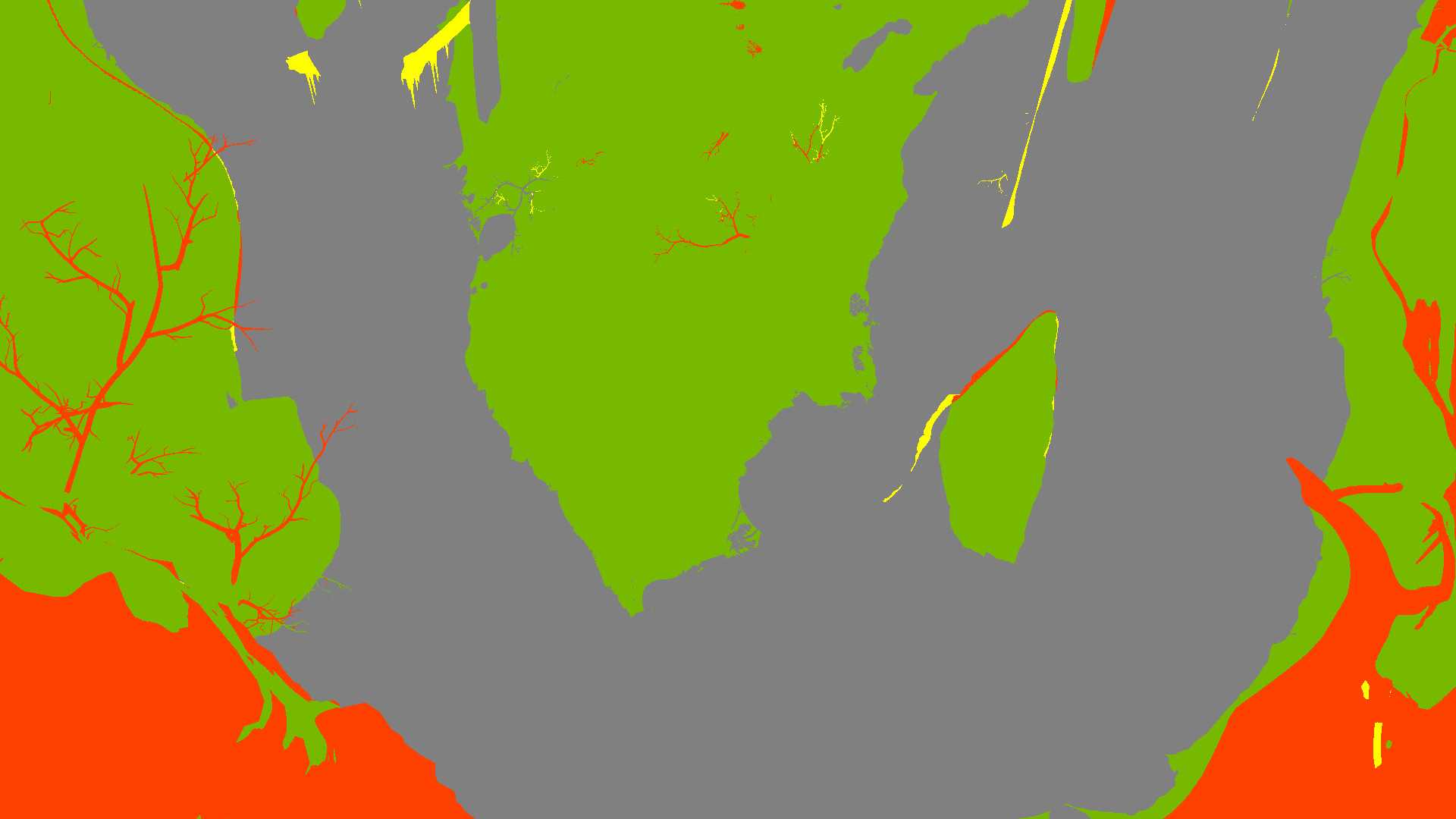} &
        \includegraphics[width=\w\textwidth,trim={0 100pt 0 0},clip]{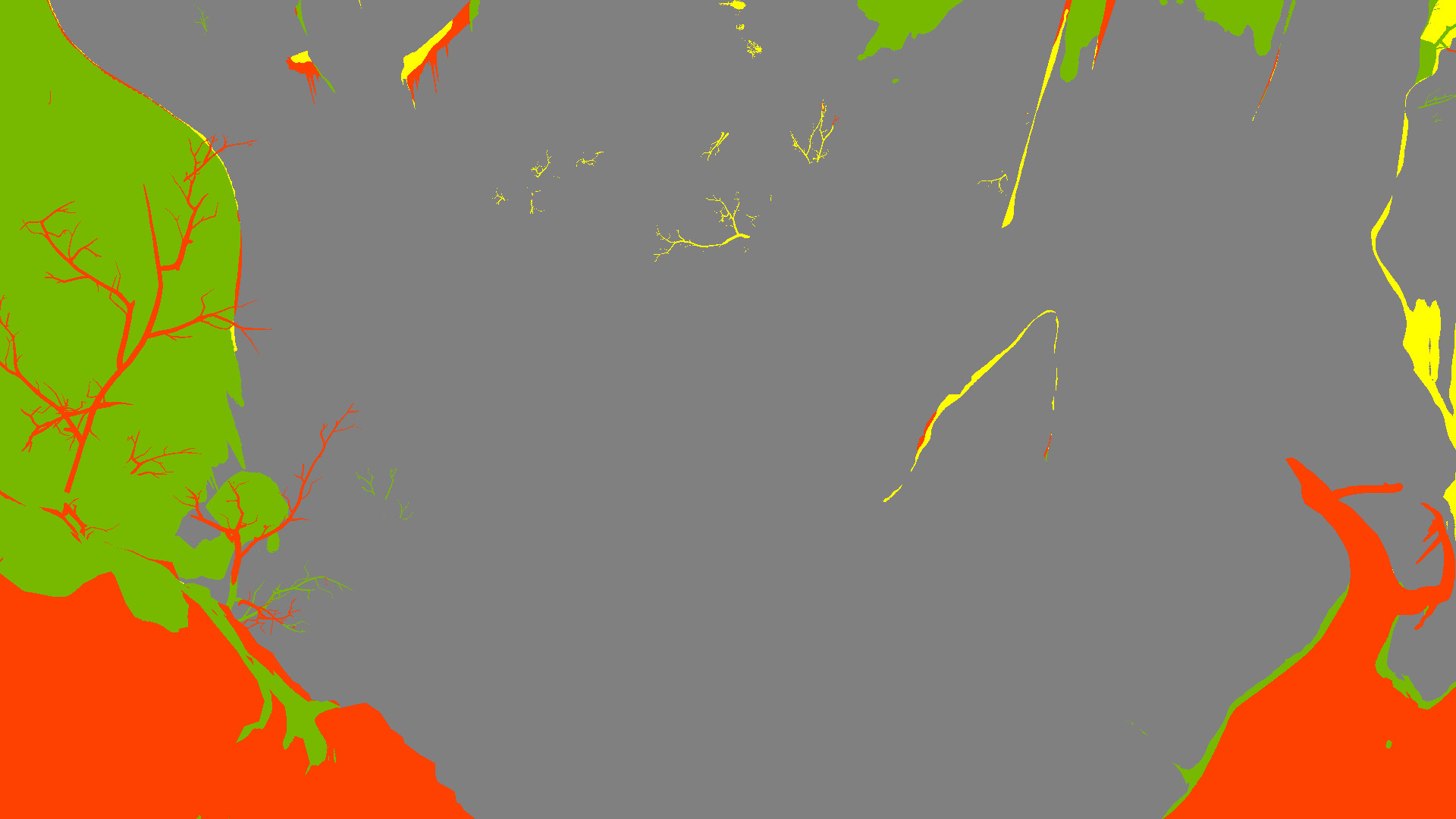} &
        \includegraphics[width=\w\textwidth,trim={0 100pt 0 0},clip]{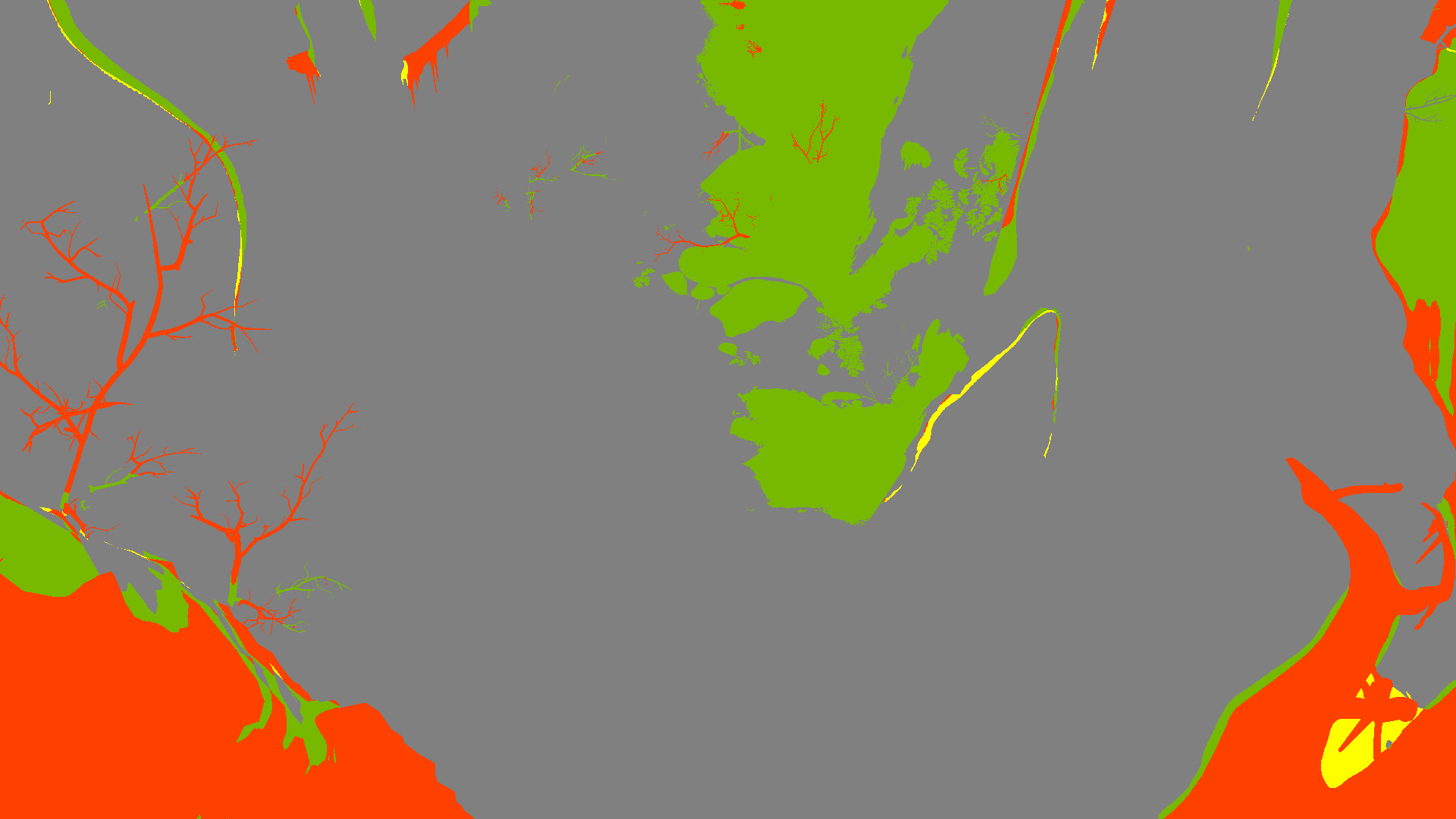}  \\
    
    \end{tabular}
    \caption{
        \textbf{Qualitative Results: Ours vs.~Peers.}
        We compare our method wth the \underline{SF} peers and one representative of \underline{\fpdepthchangeopflow} peers.
        We show qualitative results from all datasets.
        For each scene we show a pair of input images, and for each peer we show accuracy maps against \textit{Ours} for both scene flow~($AccR$) and depth~($\delta_1$).  
        Color legend is shown above.
    }
    \label{fig:peer_compare}
\end{figure*}

We compare under both in-domain and out-of-domain settings on KITTI, VKITTI2 and Spring.
The numerical results are summarized in \cref{tab:in-domain-compare} and visual results are in \cref{fig:peer_compare}. 
Our method achieves superior performance across multiple datasets and metrics
 quantitatively and qualitatively, demonstrating strong generalization capability.

\paragraph{Comparing with monocular scene flow algorithms.}
Our method achieves competitive results with previous monocular scene flow algorithms, rows marked with `$\checkmark$' in \cref{tab:in-domain-compare}, and surpasses them significantly for generalization capabilities. 
\textit{Ours} outperforms Self-Mono-SF in flow quality even for in-domain evaluations, 
achieving EPE of $0.452$ vs.~$0.454$ on KITTI, and $0.190$ vs.~$0.294$ on VKITTI2. 
When evaluating on each dataset, we also train a version of our method that does not use that dataset, listed as \textit{Ours}-\textit{exclude} in the table, which allows us to evaluate our zero-shot generalization ability.
This model significantly outperforms OpticalExpansion in terms of both scene flow and depth estimation on all datasets. 
Self-Mono-SF, trained exclusively on KITTI, fails to generalize to the out-of-domain dataset Spring, 
with AccS degrading from $0.694$ on VKITTI2 to $0.251$ on Spring and AbsR-r degrading from $0.244$ on VKITTI2 to $0.501$ on Spring. 
OpticalExpansion, also trained on autonomous-driving dataset, 
achieves better results on Spring due to its reliance on an auxiliary robust depth estimation component; 
however, its EPE on Spring ($0.029$) still falls short of \textit{Ours}-\textit{exclude}'s EPE on the same out-of-domain dataset ($0.015$).

\paragraph{Comparing with $D$+OF algorithms.}
Estimating monocular scene flow  by simply combining depth and optical flow algorithms leads to poor results. 
This can be appreciated by comparing the performance of our algorithm (\textit{Ours} in Tab.~\ref{tab:in-domain-compare}) with $D$+OF algorithms, which are marked in that table with an `$\text{\sffamily x}$'.
\textit{Ours} consistently outperforms $D$+OF algorithms in scene flow quality.
Even when our algorithm is used in a zero-shot manner, that is, the dataset used for testing was not seen in training, it performs significantly better than competing $D$+OF algorithms actually trained on that dataset.
For instance, our model tested---but not trained---on KITTI (\textit{Ours}-\textit{exclude}) achieves an EPE of $0.641$, compared with $3.404$ by DUSt3R and $3.708$ by MASt3R when they are finetuned on our data recipe following their original training protocols.
This pattern holds across datasets, demonstrating the advantages of our joint approach over separate geometry and motion estimation.

\begin{table}
    \centering
    \resizebox{\columnwidth}{!}{%
    \begin{tabular}{cc@{\hspace{1mm}}l@{\hspace{1mm}}c@{\hspace{1mm}}c@{\hspace{1mm}}c@{\hspace{1mm}}c c@{\hspace{1mm}}c@{\hspace{1mm}}c@{\hspace{1mm}}c}
        \toprule
        \multicolumn{11}{c}{KITTI~\cite{Menze2015ObjectSF} (Real)} \\
        \midrule
        \multirow{2}{*}{} & \multirow{2}{*}{SF} & \multirow{2}{*}{Method} & \multicolumn{4}{c}{Scene Flow Estimation} & \multicolumn{4}{c}{Depth Estimation}\\ 
        \cmidrule(lr){4-7}  \cmidrule(lr){8-11} 
        & & & EPE$\downarrow$ & AccS$\uparrow$ & AccR$\uparrow$ & Out$\downarrow$ & AbsR-r$\downarrow$ & $\delta_1$-r$\uparrow$ & AbsR-m$\downarrow$ & $\delta_1$-m$\uparrow$\\
        \midrule
        \multirow{4}{*}{\parbox{0mm}{\vspace{-0mm}\rotatebox[origin=c]{90}{In}}} & \checkmark & \textit{Ours} &  \textbf{0.452} & \textbf{0.398} & \textbf{0.443} & 0.873 & 0.111 & 0.927 & 0.236 & 0.345 \\
        & \checkmark & Self-Mono-SF\cite{Hur:2020:SSM} & 0.454 & 0.345 & 0.435 & \textbf{0.853} & \textbf{0.100} & 0.905 & \textbf{0.105} & \textbf{0.879} \\
        [5pt] 
        & \text{\sffamily x}  & DepthAnythingV2\cite{depth_anything_v2} & 1.835 & 0.326 & 0.361 & 0.931 & 0.104 & 0.896 & 0.134 & 0.860\\
        & \text{\sffamily x} & DUSt3R$^*$\cite{dust3r} & 3.404 & 0.268 & 0.269 & 0.997 & 0.129 & 0.901 & 0.983 & 0.000 \\
        & \text{\sffamily x} & MASt3R$^*$\cite{mast3r} & 3.708 & 0.264 & 0.267 & 0.999 & 0.108 & \textbf{0.929} & 0.245 & 0.288 \\
        \midrule
        \multirow{5}{*}{\parbox{0mm}{\vspace{-0mm}\rotatebox[origin=c]{90}{Out}}} & \checkmark 
        & \textit{Ours}-\textit{exclude} & \textbf{0.641} & \textbf{0.392} & \textbf{0.431} & \textbf{0.899} & 0.116 & \textbf{0.922} & \textbf{0.256} & \textbf{0.237} \\
        & \checkmark & OpticalExpansion\cite{yang2020upgrading} & 2.537 & 0.348 & 0.363 & 0.968 & 0.132 & 0.871 & 0.529 & 0.172 \\
        [5pt] 
        & \text{\sffamily x} & DUSt3R\cite{dust3r} & 2.088 & 0.273 & 0.278 & 0.986 & 0.144 & 0.832 & 0.968 & 0.000 \\
        & \text{\sffamily x} & MASt3R\cite{mast3r} & 3.320 & 0.277 & 0.280 & 0.996 & \textbf{0.088} & 0.921 & 0.582 & 0.075\\
        & \text{\sffamily x} & MonST3R\cite{zhang2024monst3r} & 3.594 & 0.269 & 0.272 & 0.999 & 0.099 & 0.897 & 0.951 & 0.000 \\
        \toprule
        \multicolumn{11}{c}{VKITTI2 \cite{cabon2020vkitti2} (Synthetic)} \\
        \midrule
        \multirow{2}{*}{} & \multirow{2}{*}{SF} & \multirow{2}{*}{Method} & \multicolumn{4}{c}{Scene Flow Estimation} & \multicolumn{4}{c}{Depth Estimation}\\ 
        \cmidrule(lr){4-7}  \cmidrule(lr){8-11} 
        & & & EPE$\downarrow$ & AccS$\uparrow$ & AccR$\uparrow$ & Out$\downarrow$ & AbsR-r$\downarrow$ & $\delta_1$-r$\uparrow$ & AbsR-m$\downarrow$ & $\delta_1$-m$\uparrow$\\
        \midrule
        \multirow{4}{*}{\parbox{0mm}{\vspace{-0mm}\rotatebox[origin=c]{90}{In}}} & \checkmark & \textit{Ours} & \textbf{0.190} & \textbf{0.780} & \textbf{0.876} & \textbf{0.484} & \textbf{0.137} & \textbf{0.859} & 0.236 & 0.634 \\
        & \checkmark & Self-Mono-SF\cite{Hur:2020:SSM} & 0.294 & 0.694 & 0.751 & 0.589 & 0.244 & 0.568 & 0.224 & 0.589 \\ 
        [5pt] 
        & \text{\sffamily x} & DUSt3R$^*$\cite{dust3r} & 8.822 & 0.630 & 0.645 & 0.910 & 0.176 & 0.783 & 0.981 & 0.000 \\
        & \text{\sffamily x} & MASt3R$^*$\cite{mast3r} & 5.914 & 0.635 & 0.681 & 0.827 & 0.138 & 0.858 & \textbf{0.223} & \textbf{0.672} \\
        \midrule
        \multirow{5}{*}{\parbox{0mm}{\vspace{-0mm}\rotatebox[origin=c]{90}{Out}}} & \checkmark 
        & \textit{Ours}-\textit{exclude} & \textbf{0.409} & \textbf{0.727} & \textbf{0.772} & \textbf{0.696} & \textbf{0.144} & \textbf{0.836} & \textbf{0.168} & \textbf{0.778}  \\
        & \checkmark & OpticalExpansion\cite{yang2020upgrading}  & 1.809 & 0.654 & 0.662 & 0.946 & 0.194 & 0.727 & 0.203 & 0.746 \\
        [5pt] 
        & \text{\sffamily x} & DUSt3R\cite{dust3r} & 2.279 & 0.633 & 0.673 & 0.811 & 0.246 & 0.586 & 0.970 & 0.000 \\
        & \text{\sffamily x} & MASt3R\cite{mast3r} & 2.946 & 0.666 & 0.713 & 0.797 & 0.166 & 0.827 & 0.676 & 0.050 \\
        & \text{\sffamily x} & MonST3R\cite{zhang2024monst3r} & 2.206 & 0.629 & 0.659 & 0.853 & 0.235 & 0.606 & 0.949 & 0.000 \\
        \toprule
        \multicolumn{11}{c}{Spring \cite{Mehl2023_Spring} (Synthetic)} \\
        \midrule
        \multirow{2}{*}{} & \multirow{2}{*}{SF} & \multirow{2}{*}{Method} & \multicolumn{4}{c}{Scene Flow Estimation} & \multicolumn{4}{c}{Depth Estimation}\\ 
        \cmidrule(lr){4-7}  \cmidrule(lr){8-11} 
        & & & EPE$\downarrow$ & AccS$\uparrow$ & AccR$\uparrow$ & Out$\downarrow$ & AbsR-r$\downarrow$ & $\delta_1$-r$\uparrow$ & AbsR-m$\downarrow$ & $\delta_1$-m$\uparrow$\\
        \midrule
        \multirow{3}{*}{\parbox{0mm}{\vspace{-0mm}\rotatebox[origin=c]{90}{In}}} & 
        \checkmark & \textit{Ours} & \textbf{0.013} & \textbf{0.989} & \textbf{0.999} & \textbf{0.813} & \textbf{0.280} & \textbf{0.597} & \textbf{0.679} & \textbf{0.119} \\
        [5pt] 
        & \text{\sffamily x} & DUSt3R$^*$\cite{dust3r} & 0.803 & 0.242 & 0.402 & 0.867 & 0.324 & 0.545 & 0.965 & 0.000 \\
        & \text{\sffamily x} & MASt3R$^*$\cite{mast3r} & 0.985 & 0.208 & 0.267 & 0.867 & 0.288 & 0.550 & 0.689 & 0.064 \\
        \midrule
        \multirow{6}{*}{\parbox{0mm}{\vspace{-0mm}\rotatebox[origin=c]{90}{Out}}} & \checkmark & \textit{Ours}-\textit{exclude} & \textbf{0.014} & \textbf{0.992} & \textbf{1.000} & \textbf{0.787} & \textbf{0.294} & \textbf{0.599} & \textbf{0.682} & 0.130 \\
        & \checkmark & Self-Mono-SF\cite{Hur:2020:SSM} & 1.005 & 0.251 & 0.328 & 0.880 & 0.501 & 0.353 & 0.784 & 0.044 \\
        & \checkmark & OpticalExpansion\cite{yang2020upgrading}  & 0.029 & 0.873 & 0.971 & 0.805 & 0.430 & 0.468 & 0.804 & 0.004 \\
        [5pt] 
        & \text{\sffamily x} & DepthAnythingV2\cite{depth_anything_v2} & 0.208 & 0.496 & 0.658 & 0.820 & 0.379 & 0.519 & 0.758 & \textbf{0.181} \\
        & \text{\sffamily x} & DUSt3R\cite{dust3r}  & 0.275 & 0.431 & 0.629 & 0.831 & 0.476 & 0.447 & 0.939 & 0.000 \\
        & \text{\sffamily x} & MASt3R\cite{mast3r} & 0.610 & 0.315 & 0.446 & 0.855 & 0.343 & 0.546 & 0.811 & 0.009 \\
        \bottomrule
    \end{tabular}
    }
    \caption{
        \textbf{In/Out-of-Domain Quantitative Results: Ours vs.~Peers.}
        We exclude DepthAnythingV2 for VKITTI2 and MonST3R for the Spring dataset as they used the test set for training.
        $^*$: Finetuned following the original training protocol on our data recipe.
        }
    \label{tab:in-domain-compare}
\end{table}

\subsection{Scale-Adaptive Optimization}
Scale-adaptive optimization is key for our large-scale, diverse data recipe, 
as we show in \cref{tab:ablation} on KITTI dataset.     

\paragraph{Scale-handling strategies.}
The choice significantly impacts both scene flow and geometry estimation.
As mentioned in Sec.~\ref{sec:scale-adaptive_optimization}, to train across datasets that provide both metric and relative depth, we need to properly handle scale.
Here we analyze the impact of four strategies.
First, we train our model by simply rescaling our prediction to match the ground truth before computing the loss (\emph{Align} in \cref{tab:ablation}).
Then, we follow DUSt3R's strategy, which normalizes the ground truth and our prediction to be between $0$ and $1$ (\emph{Always} in \cref{tab:ablation}).
We also follow MASt3R's strategy, which assumes the ground truth to be metric and thus does not perform any normalization (\emph{Never} in \cref{tab:ablation}).
Finally, we combine the latter two strategies and normalize the ground truth and our prediction when the ground truth is relative depth and do nothing when it is metric (\emph{Xor} in \cref{tab:ablation}).
The Xor-based variants achieve the best scene flow estimation (EPE: $0.452$, $0.501$), significantly outperforming \emph{Align} (EPE: $1.596$), \emph{Always} (EPE: $0.899$), and \emph{Never} (EPE: $0.758$). 
Therefore, we use it as our training strategy.

\paragraph{Optical flow supervision.} $\mathcal{L}_\mu$ improves both scene flow and geometry.
Our best model trained without the flow loss $\mathcal{L}_\mu$ (`$\text{\sffamily x}$' in the table) shows a drop of $0.049$ in EPE for scene flow.
Perhaps surprisingly, using $\mathcal{L}_\mu$ also improves depth estimation with an improvement of $0.034$ in terms of  $\delta_1$-m.

\subsection{Scene Flow Parameterization}
We show that zero-shot monocular scene flow estimation is sensitive to parameterization (\cref{sec:sf-rep}) with an ablation study on the KITTI test set~(\cref{tab:ablation}). 
\fpcamspaceoffsets achieves superior scene flow accuracy (AccS: $0.398$) 
compared to both \fpendpoint (AccS: $0.281$) and \fpdepthchangeopflow (AccS: $0.311$) 
while maintaining comparable depth estimation quality ($\delta_1-r$: $0.927$ vs. $0.927$). 
Notably, \fpdepthchangeopflow's performance significantly degrades in scene flow estimation when camera poses are unknown, 
as evidenced by its poor EPE ($11.905$) on KITTI. 
This validates our choice of \fpcamspaceoffsets parameterization for robust scene flow estimation under real-world conditions.

\begin{table}
    \centering
    \resizebox{\columnwidth}{!}{
    \begin{tabular}{c@{\hspace{1mm}}l@{\hspace{1mm}}c c@{\hspace{2mm}}c@{\hspace{2mm}}c@{\hspace{1mm}}c c@{\hspace{1mm}}c@{\hspace{2mm}}c@{\hspace{1mm}}c}
    \toprule
    \multirow{2}{0.8cm}{Flow Param.} & \multicolumn{2}{c}{Loss Design} & \multicolumn{4}{c}{Scene Flow Estimation} & \multicolumn{4}{c}{Depth Estimation}\\ 
    \cmidrule(lr){2-3} \cmidrule(lr){4-7} \cmidrule(lr){8-11}
     & Norm & $\mathcal{L}_\mu$ & EPE$\downarrow$ & AccS$\uparrow$ & AccR$\uparrow$ & Out$\downarrow$ & AbsR-r$\downarrow$ & $\delta_1$-r$\uparrow$ & AbsR-m$\downarrow$ & $\delta_1$-m$\uparrow$\\
    \midrule
    \multirow{5}{*}{\fpcamspaceoffsets}  
    & Align & \checkmark  & 1.596 & 0.291 & 0.295 & 0.993 & 0.241 & 0.620 & 0.994 & 0.000 \\
    & Always & \checkmark & 0.899 & 0.362 & 0.384 & 0.931 & 0.120 & 0.915 & 0.352 & 0.129 \\
    & Never & \checkmark  & 0.758 & 0.323 & 0.359 & 0.941 & \textbf{0.106} & \textbf{0.930} & 0.247 & 0.281 \\
    & Xor & \text{\sffamily x} & \underline{0.501} & \underline{0.367} & \underline{0.410} & \underline{0.903} & 0.111 & 0.927 & \underline{0.240} & \underline{0.319}  \\
    \rowcolor{gray!20} & Xor & \checkmark &  \textbf{0.452} & \textbf{0.398} & \textbf{0.443} & \textbf{0.873} & \underline{0.111} & \underline{0.927} & \textbf{0.236} & \textbf{0.345}  \\
    \toprule
    \fpendpoint   & \multirow{3}{*}{Xor} & \multirow{3}{*}{\checkmark}  & \underline{0.618} & 0.281 & \underline{0.362} & \underline{0.915} & \textbf{0.109} & 0.926 & \textbf{0.228} & \textbf{0.391} \\
    \fpdepthchangeopflow  &  &  & 11.905 & \underline{0.311} & 0.323 & 0.993 & 0.109 & \underline{0.927} & \underline{0.231} & \underline{0.373} \\
    \rowcolor{gray!20} \fpcamspaceoffsets &  &  &  \textbf{0.452} & \textbf{0.398} & \textbf{0.443} & \textbf{0.873} & \underline{0.111} & \textbf{0.927} & 0.236 & 0.345  \\
    \bottomrule
    \end{tabular}
    }
    \caption{\textbf{Ablate over Scale-Adaptive Optimization and Scene Flow Parameterization on KITTI.}
    (a) Flow Parameterization (\underline{\fpcamspaceoffsets}: Camera-space 3D offsets. 
    \underline{\fpdepthchangeopflow}: Depth change and optical flow. 
    \underline{\fpendpoint}: End Point).\\
    (b) Loss Design $\rightarrow$ \textit{Norm}: how to handle Dataset Scales with Normalization. $\mathcal{L}_\mu$: whether to use Projected 2D Loss. \\
    }
    \label{tab:ablation}
\end{table}

\subsection{Additional Qualitative Results}

\begin{figure*}[t]
    \setlength{\tabcolsep}{5pt}
    \renewcommand{\arraystretch}{0.6}  
    \vspace{-8mm}
    \resizebox{\textwidth}{!}{
        \tiny
    \begin{tabular}{rr}        
        Input &  Groundtruth\\
        \includegraphics[width=.3\columnwidth,trim={0 -10pt 0 30pt},clip]{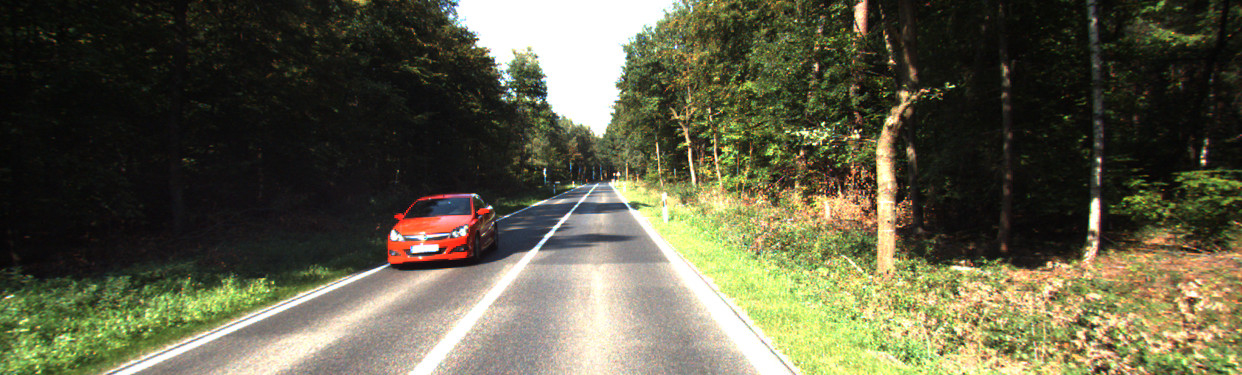}
        &
        \includegraphics[width=.3\columnwidth,trim={0 0 0 10pt},clip]{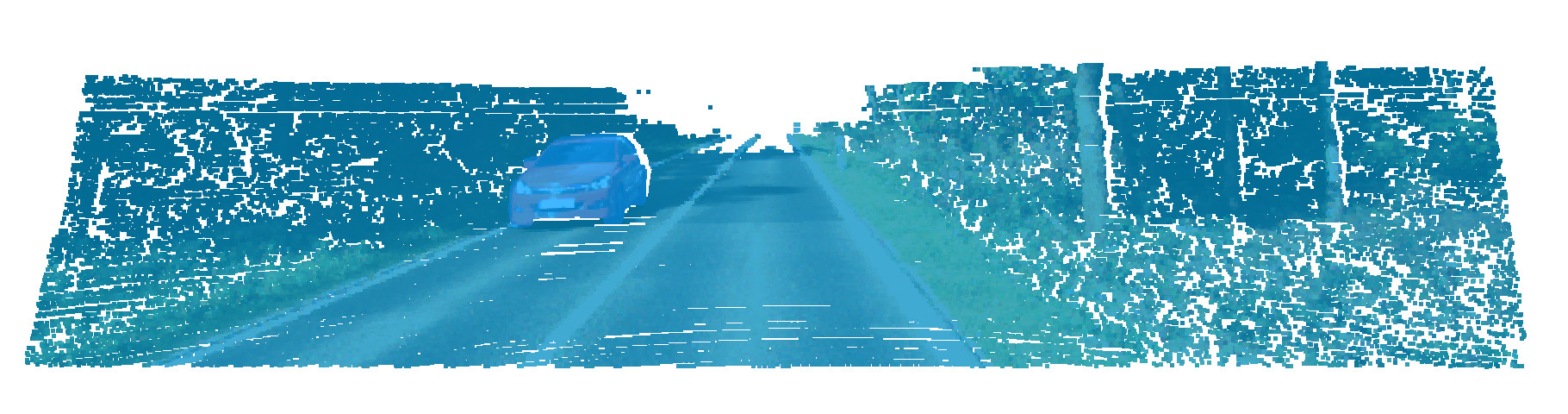}
       \\
       Self-Mono-SF~\cite{Hur:2020:SSM} & Ours
       \\
       \includegraphics[width=.3\columnwidth,trim={0 0 0 10pt},clip]{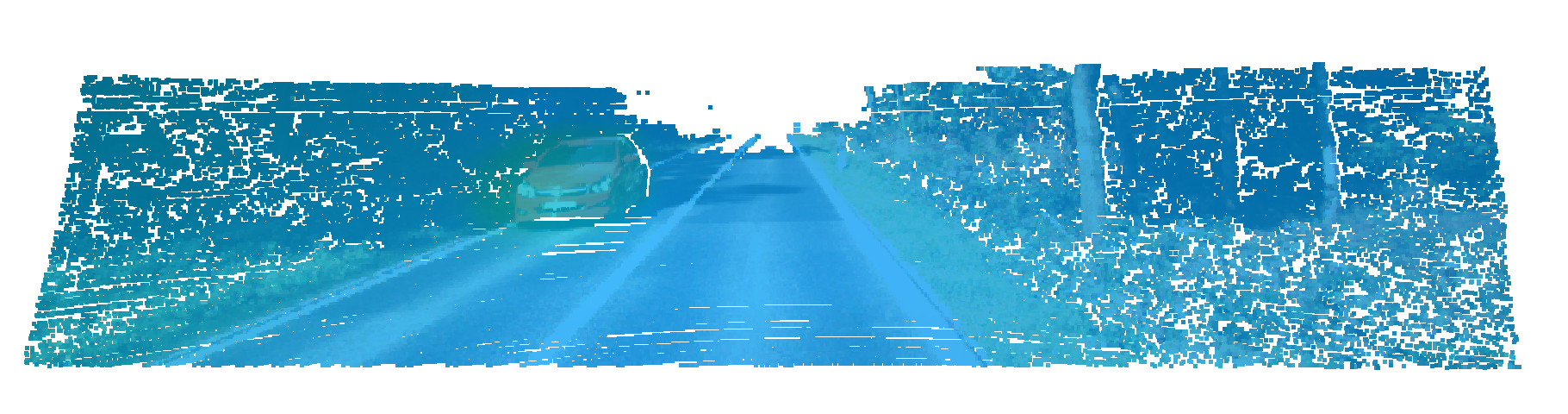}
        &
        \includegraphics[width=.3\columnwidth,trim={0 0 0 10pt},clip]{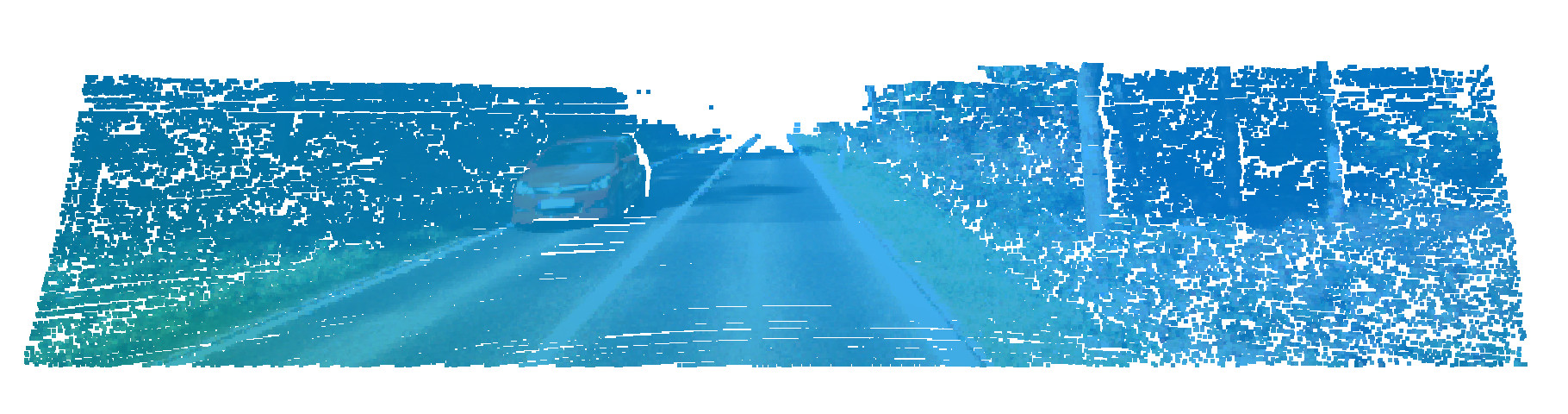}
        \\
        OpticalExpansion~\cite{yang2020upgrading} & MASt3R~\cite{mast3r}
       \\
       \includegraphics[width=.3\columnwidth,trim={0 0 0 10pt},clip]{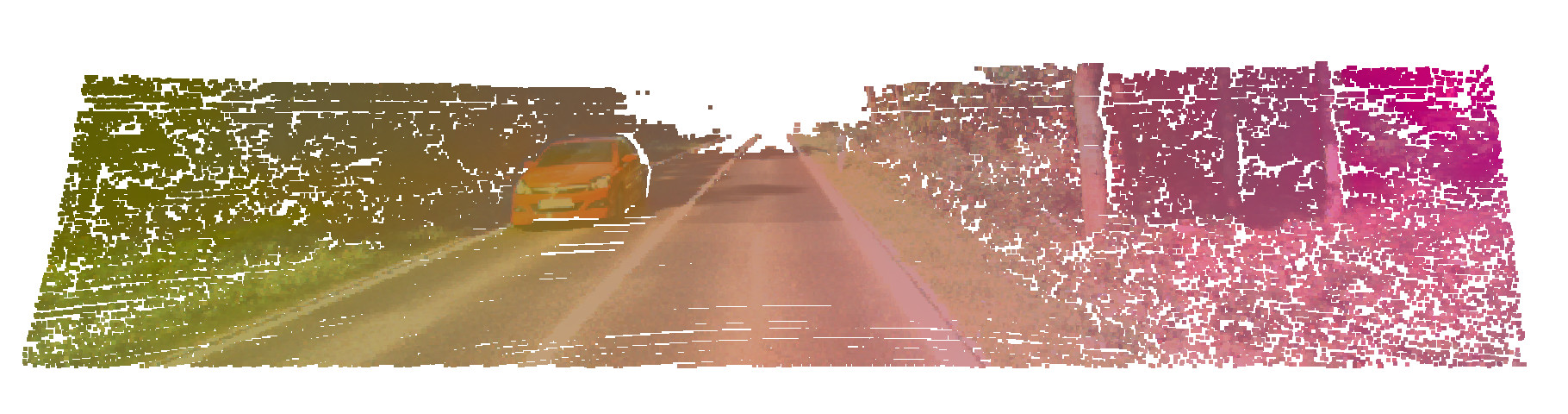}
        &
        \includegraphics[width=.3\columnwidth,trim={0 0 0 10pt},clip]{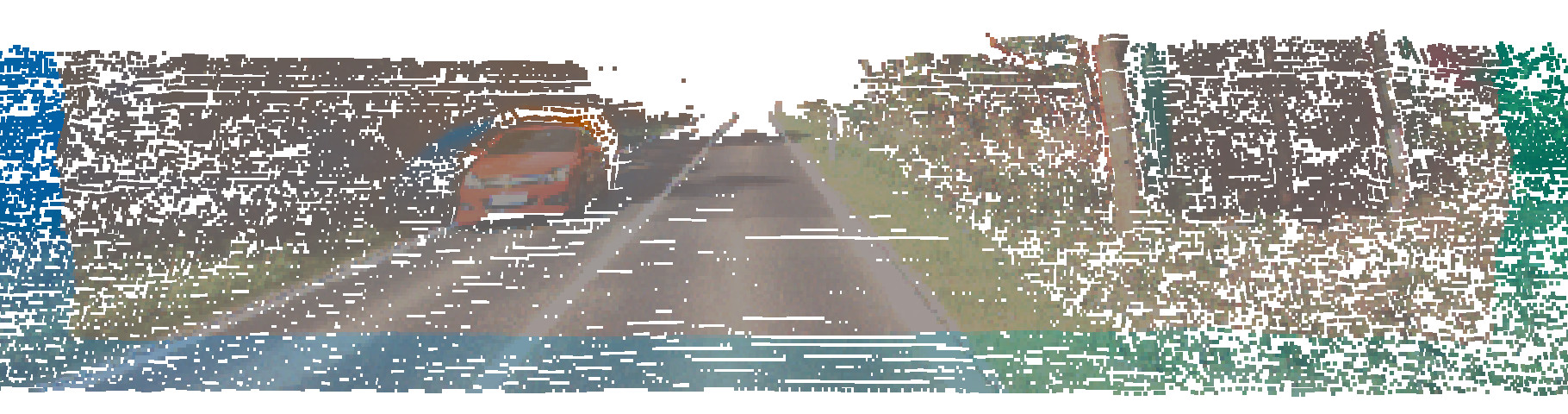}
        \\
        \midrule
        Input &  Groundtruth\\
        \includegraphics[width=.3\columnwidth,trim={0 -10pt 0 30pt},clip]{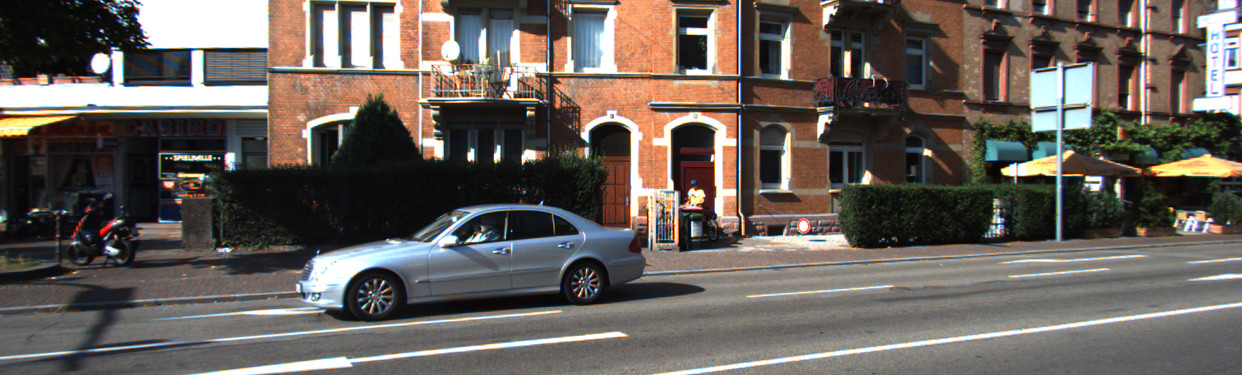}
        &
        \includegraphics[width=.3\columnwidth,trim={0 0 0 10pt},clip]{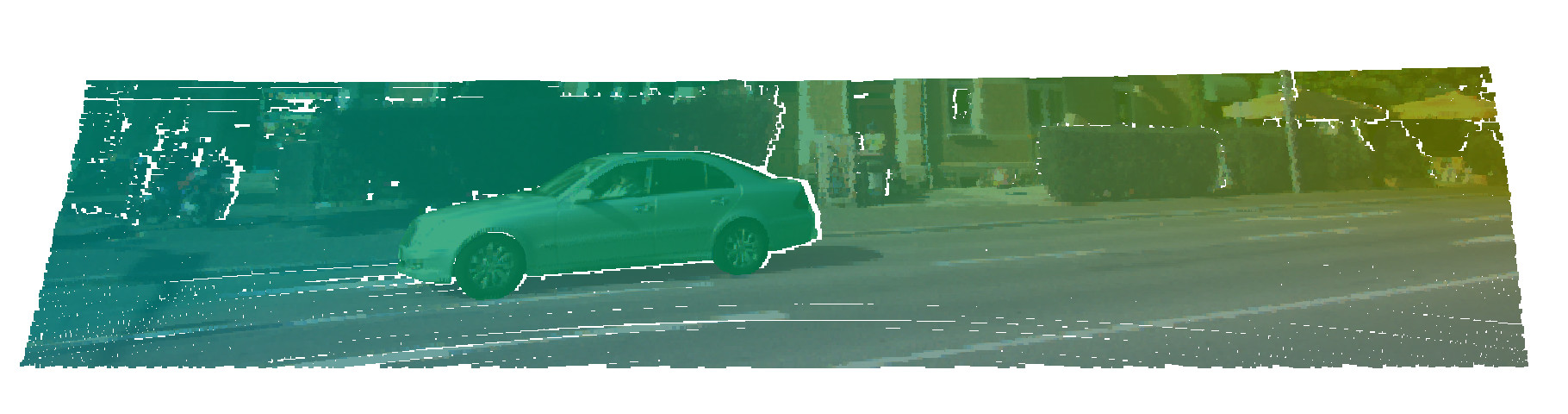}
       \\
       Self-Mono-SF~\cite{Hur:2020:SSM} & Ours
       \\
       \includegraphics[width=.3\columnwidth,trim={0 0 0 10pt},clip]{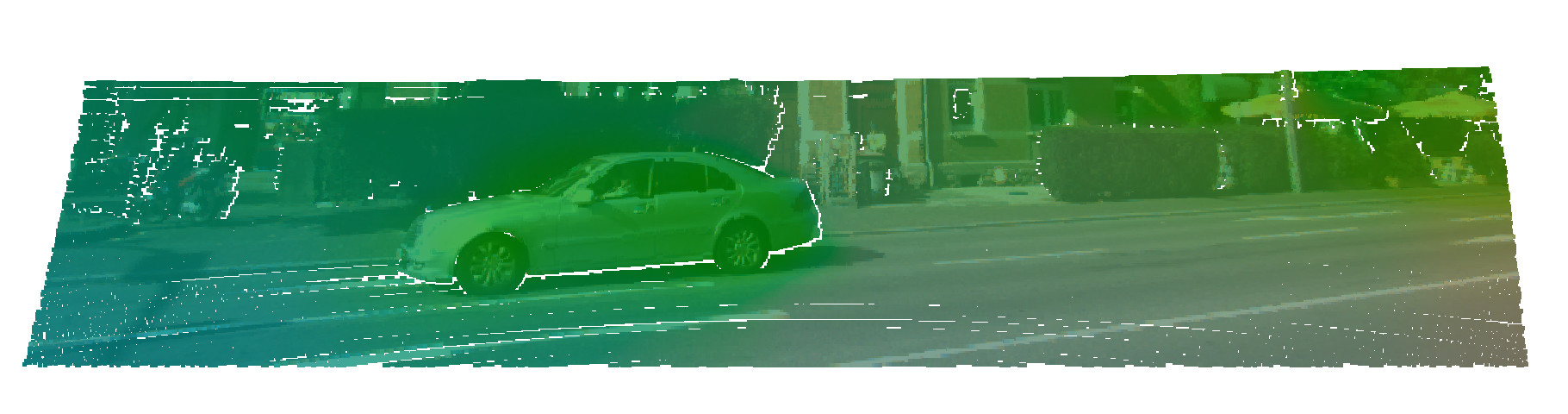}
        &
        \includegraphics[width=.3\columnwidth,trim={0 0 0 10pt},clip]{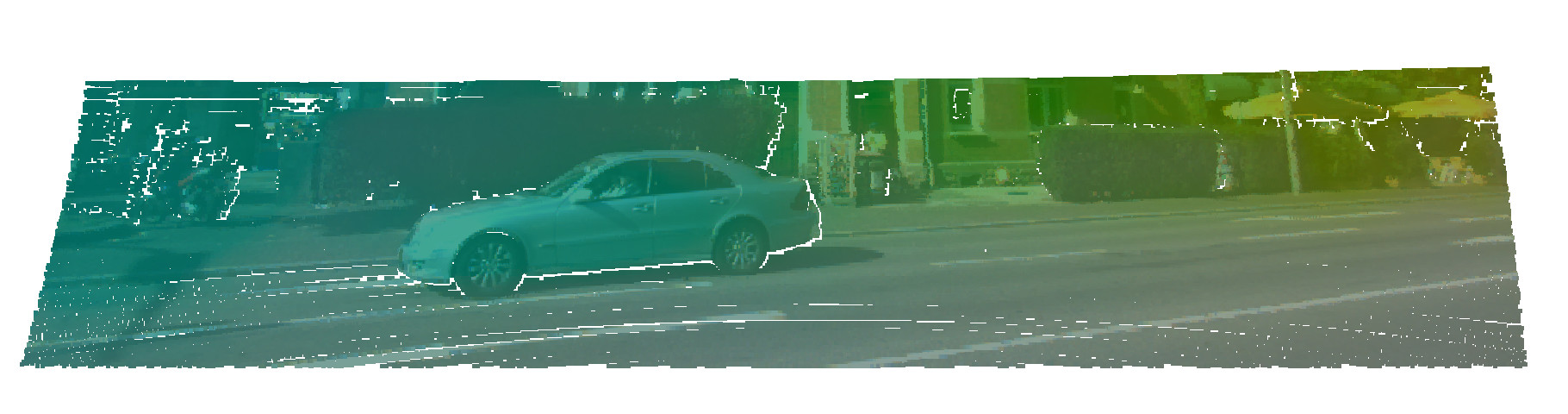}
        \\
        OpticalExpansion~\cite{yang2020upgrading} & MASt3R~\cite{mast3r}
       \\
       \includegraphics[width=.3\columnwidth,trim={0 0 0 10pt},clip]{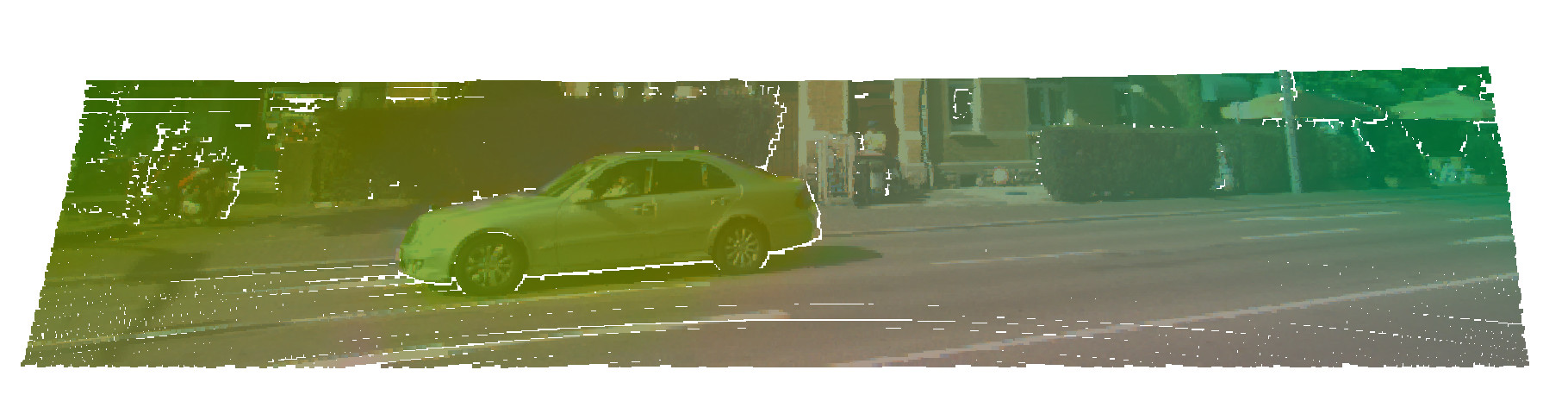}
        &
        \includegraphics[width=.3\columnwidth,trim={0 0 0 10pt},clip]{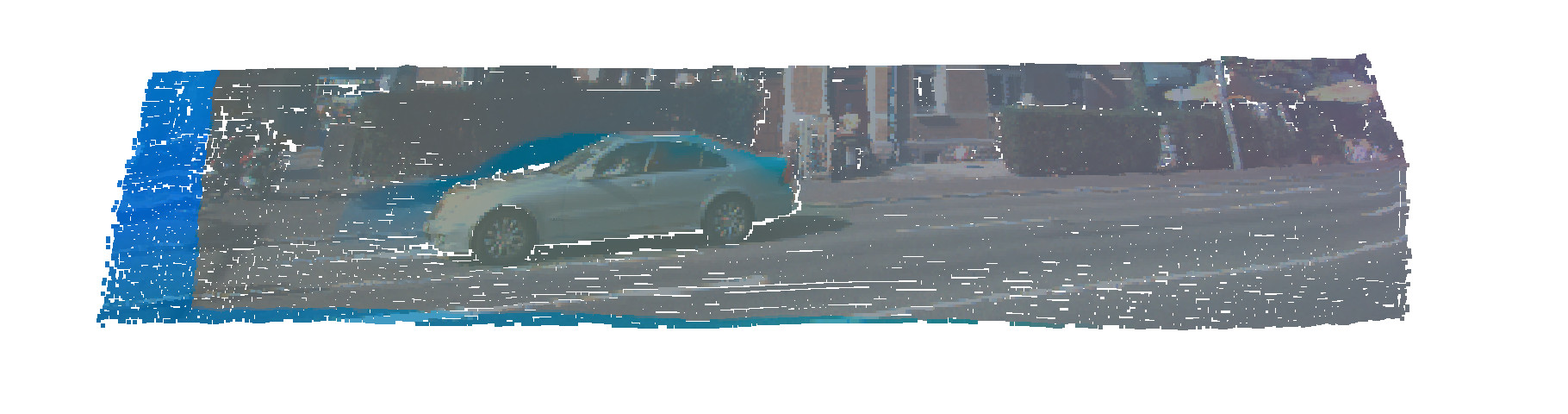}
        \\
        \midrule
        Input &  Groundtruth\\
        \includegraphics[width=.3\columnwidth,trim={0 -10pt 0 30pt},clip]{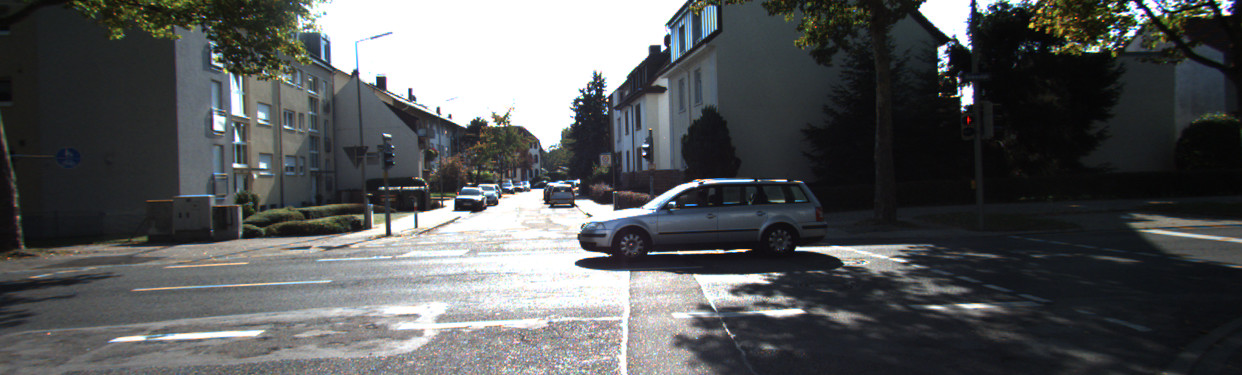}
        &
        \includegraphics[width=.3\columnwidth,trim={0 0 0 10pt},clip]{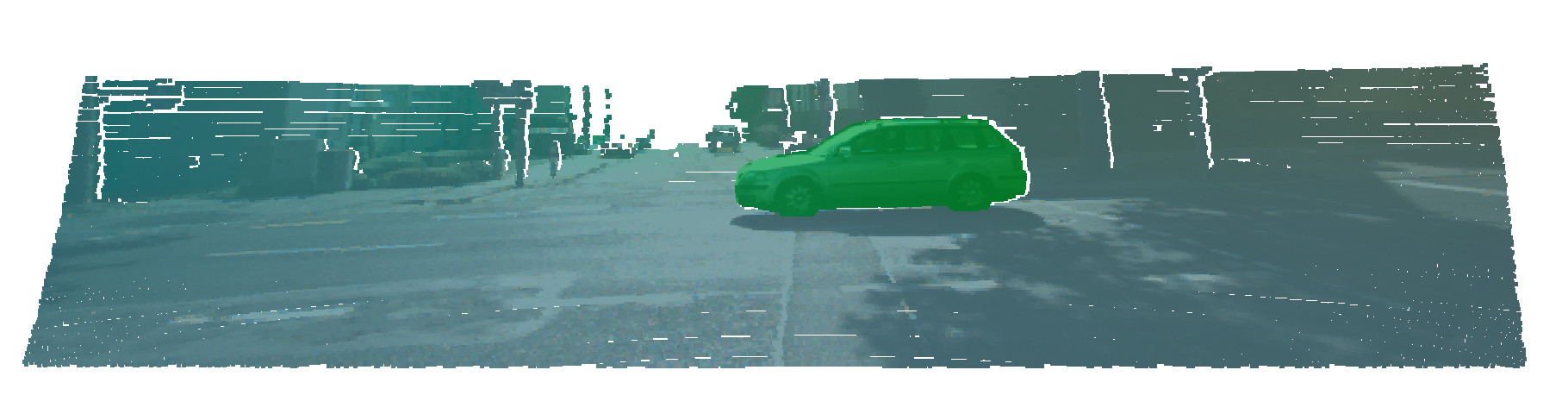}
       \\
       Self-Mono-SF~\cite{Hur:2020:SSM} & Ours
       \\
       \includegraphics[width=.3\columnwidth,trim={0 0 0 10pt},clip]{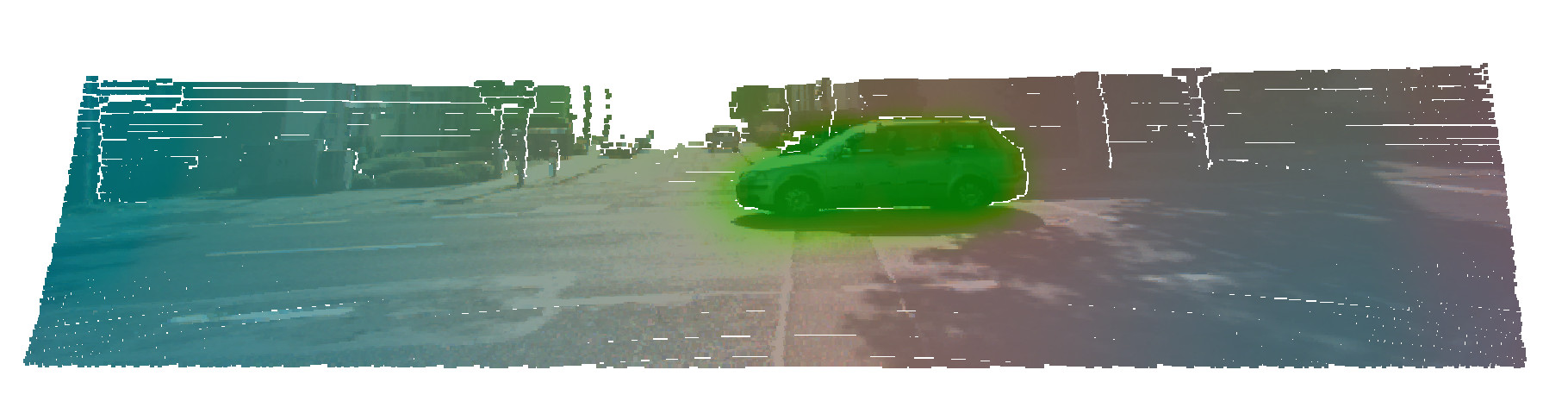}
        &
        \includegraphics[width=.3\columnwidth,trim={0 0 0 10pt},clip]{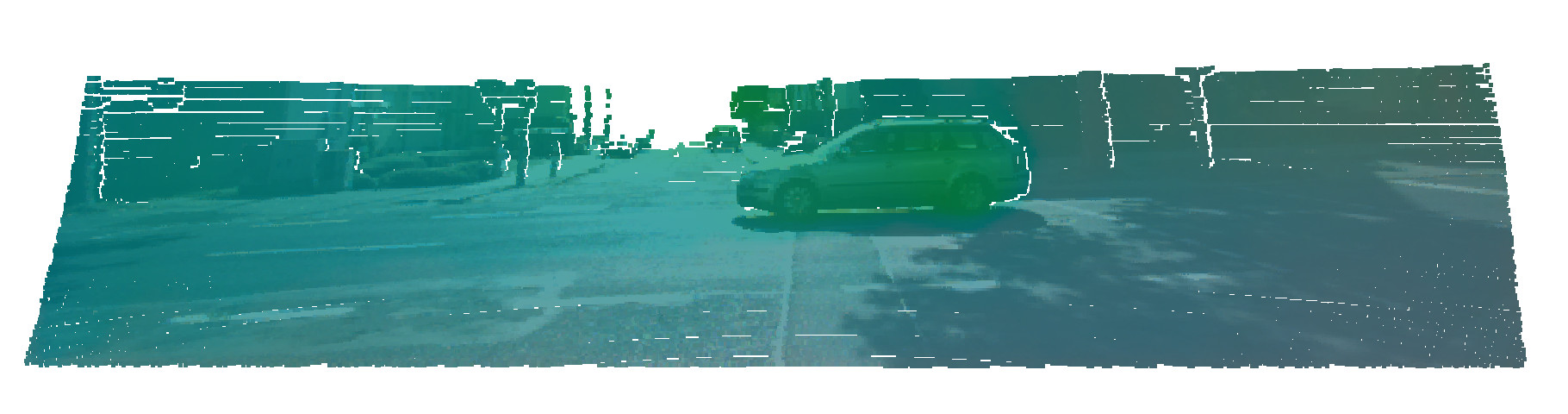}
        \\
        OpticalExpansion~\cite{yang2020upgrading} & MASt3R~\cite{mast3r}
       \\
       \includegraphics[width=.3\columnwidth,trim={0 0 0 10pt},clip]{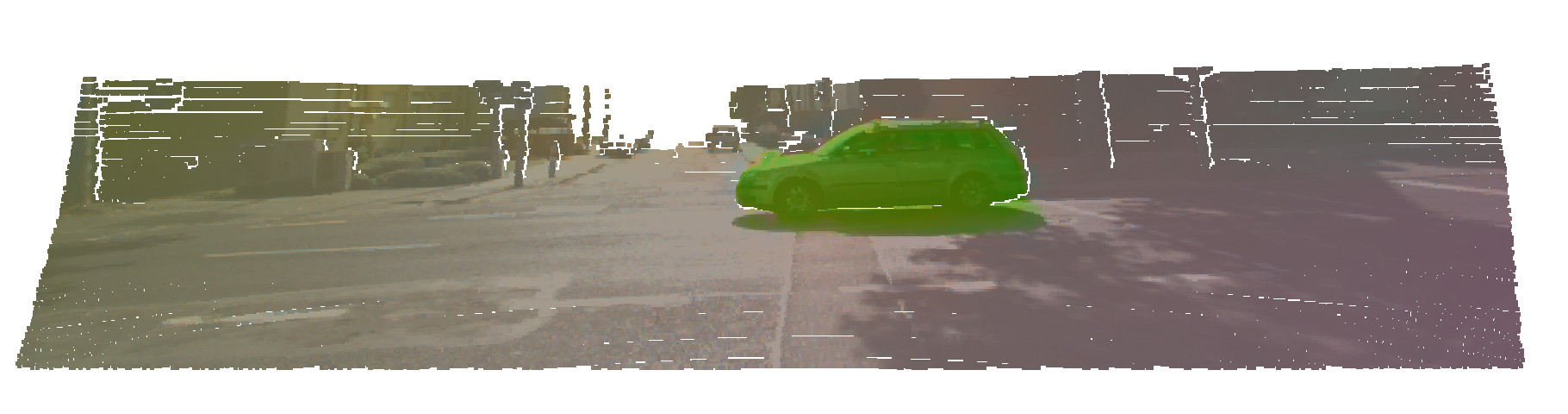}
        &
        \includegraphics[width=.3\columnwidth,trim={0 0 0 10pt},clip]{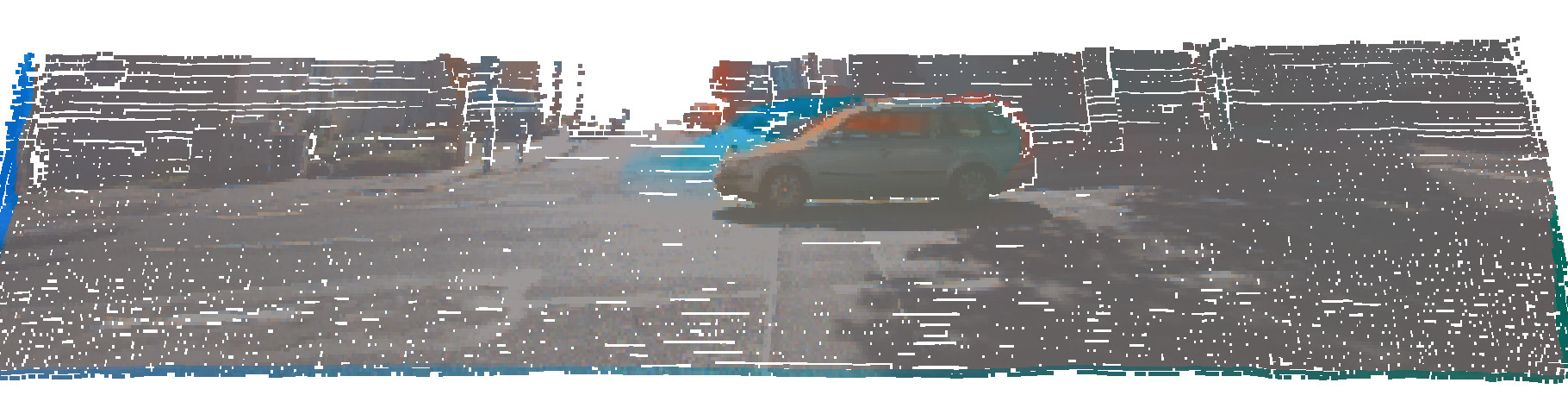}
        \\
    \end{tabular}
    }
    \vspace{-2mm}
    \caption{\textbf{Additional Qualitative Results on KITTI~\cite{Menze2015ObjectSF}}, 
    with scene flow represented by CIE-LAB~\cite{Hur:2021:SSM} colormap overlaid on corresponding 3D structure estimated for first input image.
    Ours method shows the highest similarity to the groundtruth.}
    \label{tab:quali_kitti}
\end{figure*}

\begin{figure*}[t]
    \setlength{\tabcolsep}{5pt}
    \renewcommand{\arraystretch}{0.6}  
    \vspace{-8mm}
    \resizebox{\textwidth}{!}{
        \tiny
    \begin{tabular}{rr}        
        Input & \\
        \includegraphics[width=.3\columnwidth,trim={0 0 0 0},clip]{figures/finalvis/davis/bike-packing/left.jpg}
        & \\
        \phantom{x} & \\
        Self-Mono-SF~\cite{Hur:2020:SSM} & Ours \\
        \includegraphics[width=.3\columnwidth,trim={0 0 0 30pt},clip]{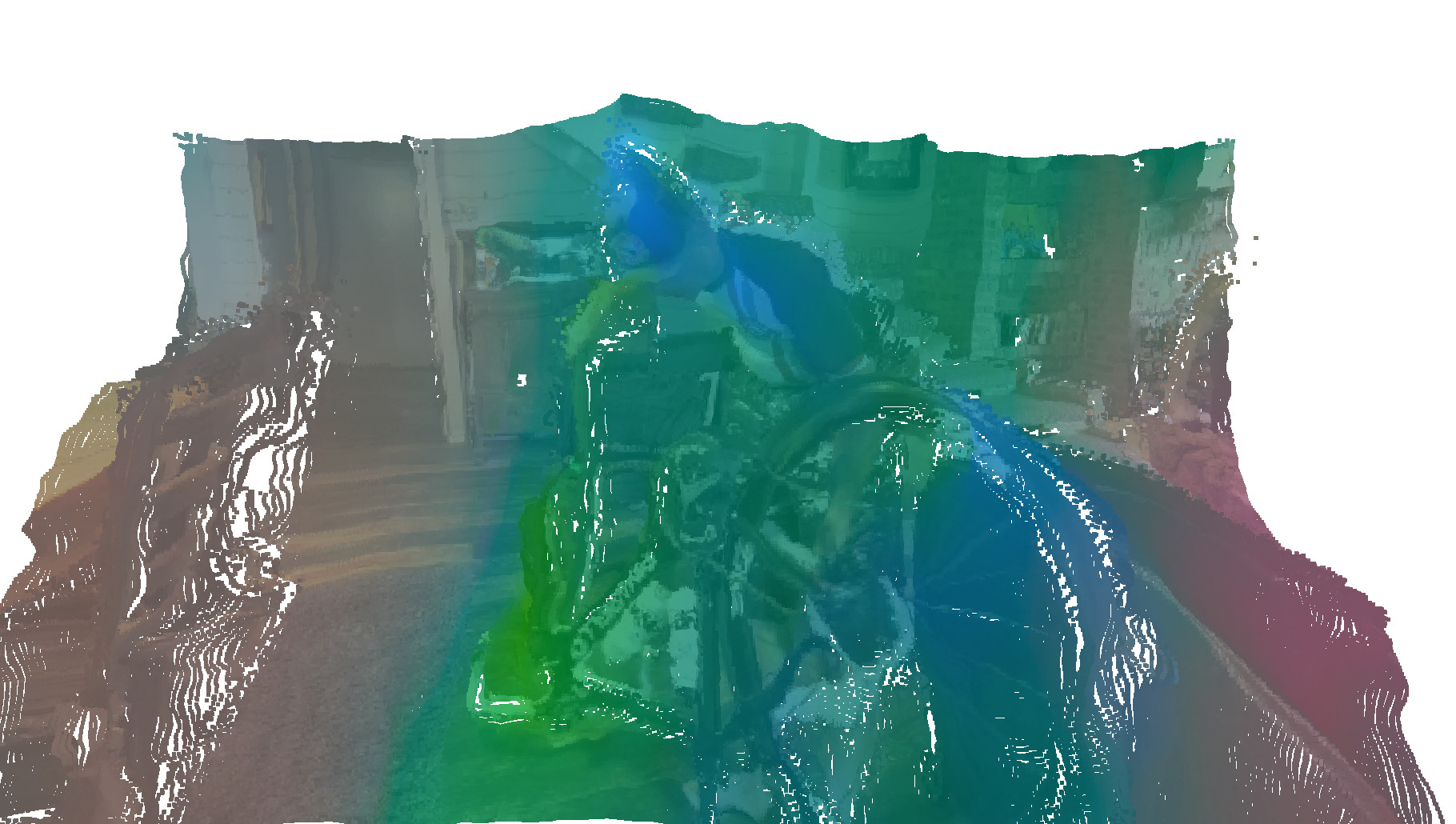}
        &
        \includegraphics[width=.3\columnwidth,trim={0 0 0 30pt},clip]{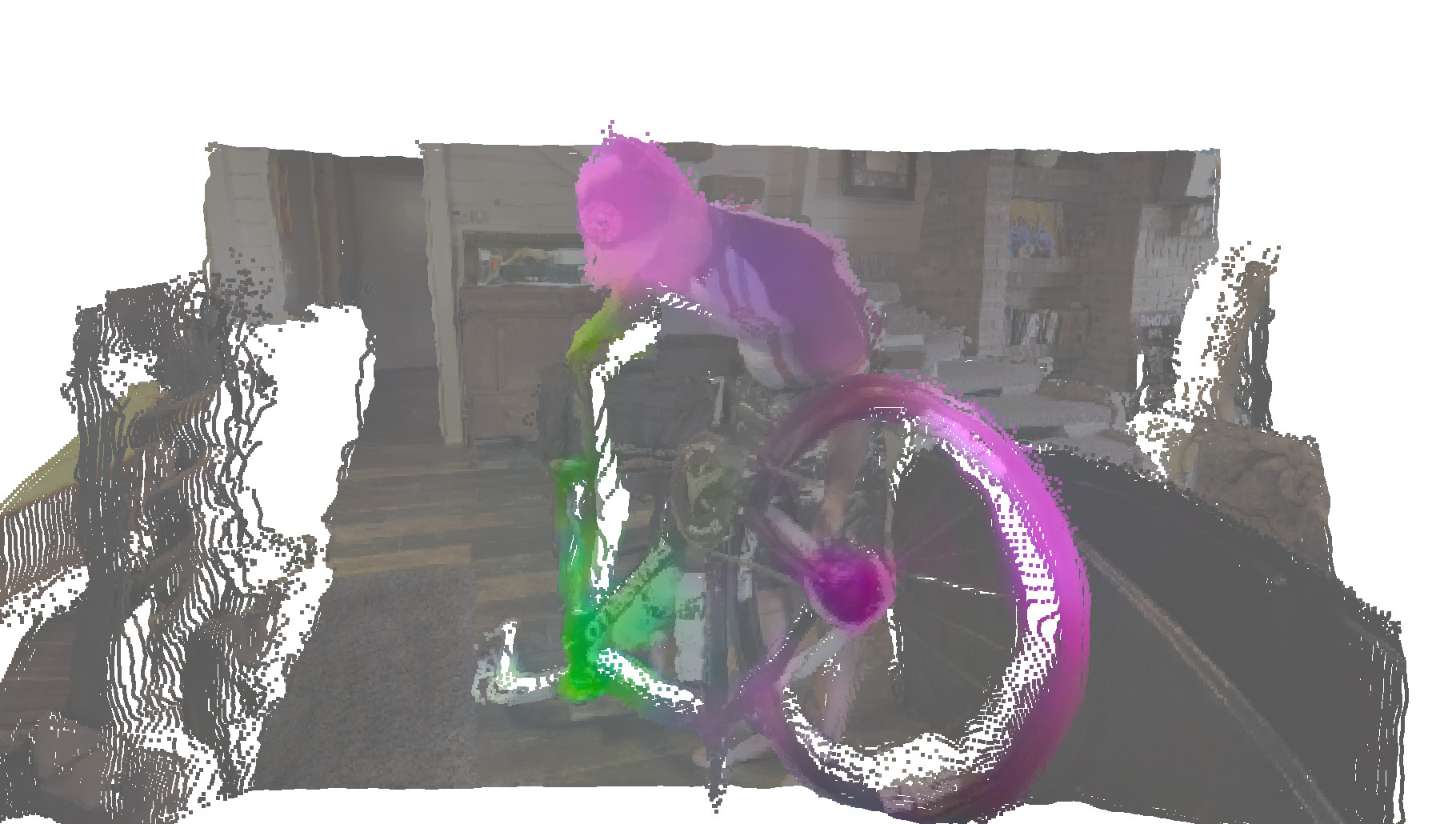}
        \\
        \phantom{x} & \\
        OpticalExpansion~\cite{yang2020upgrading} & MASt3R~\cite{mast3r} \\
        \includegraphics[width=.3\columnwidth,trim={0 0 0 30pt},clip]{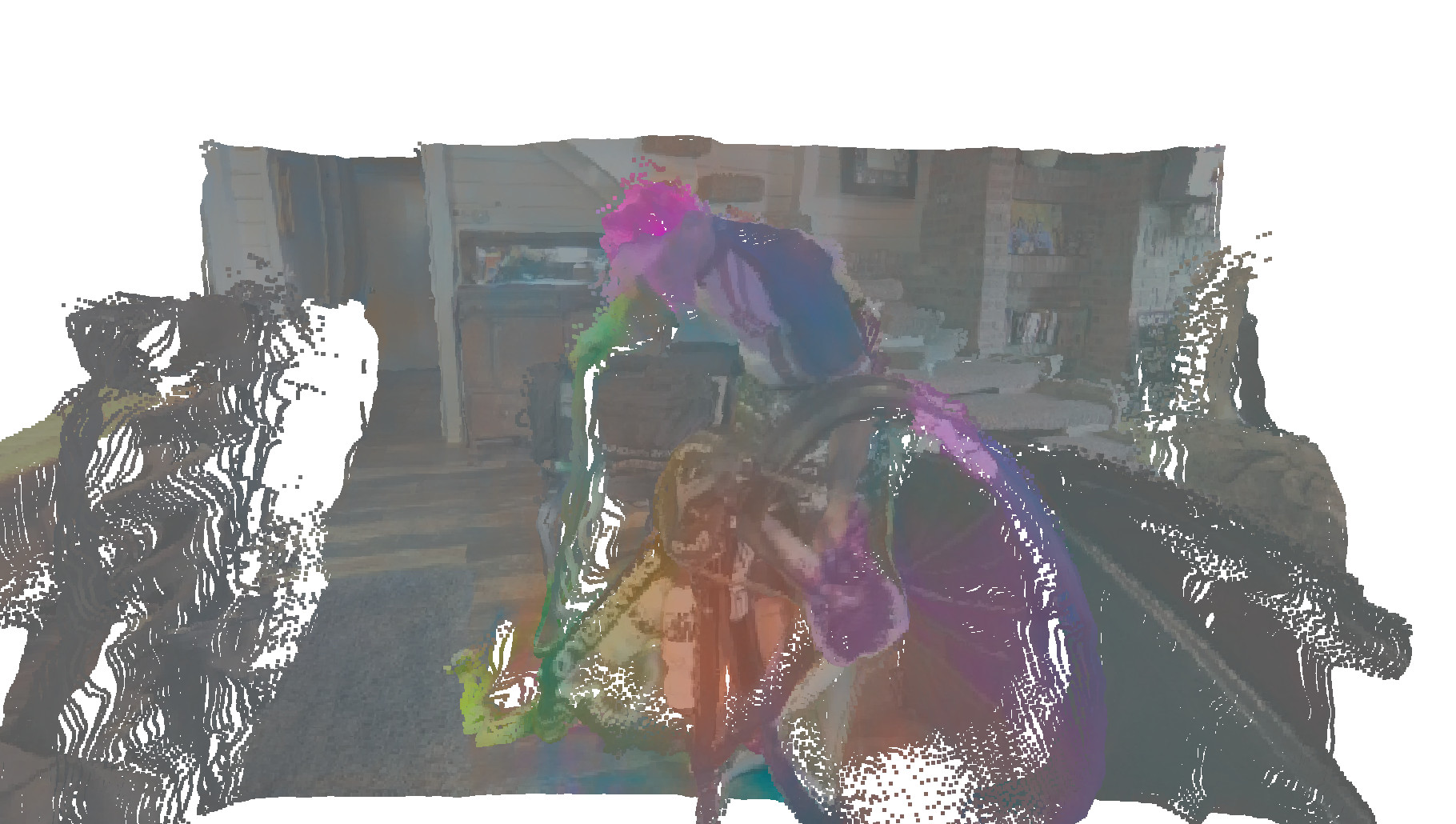}
        &
        \includegraphics[width=.3\columnwidth,trim={0 0 0 30pt},clip]{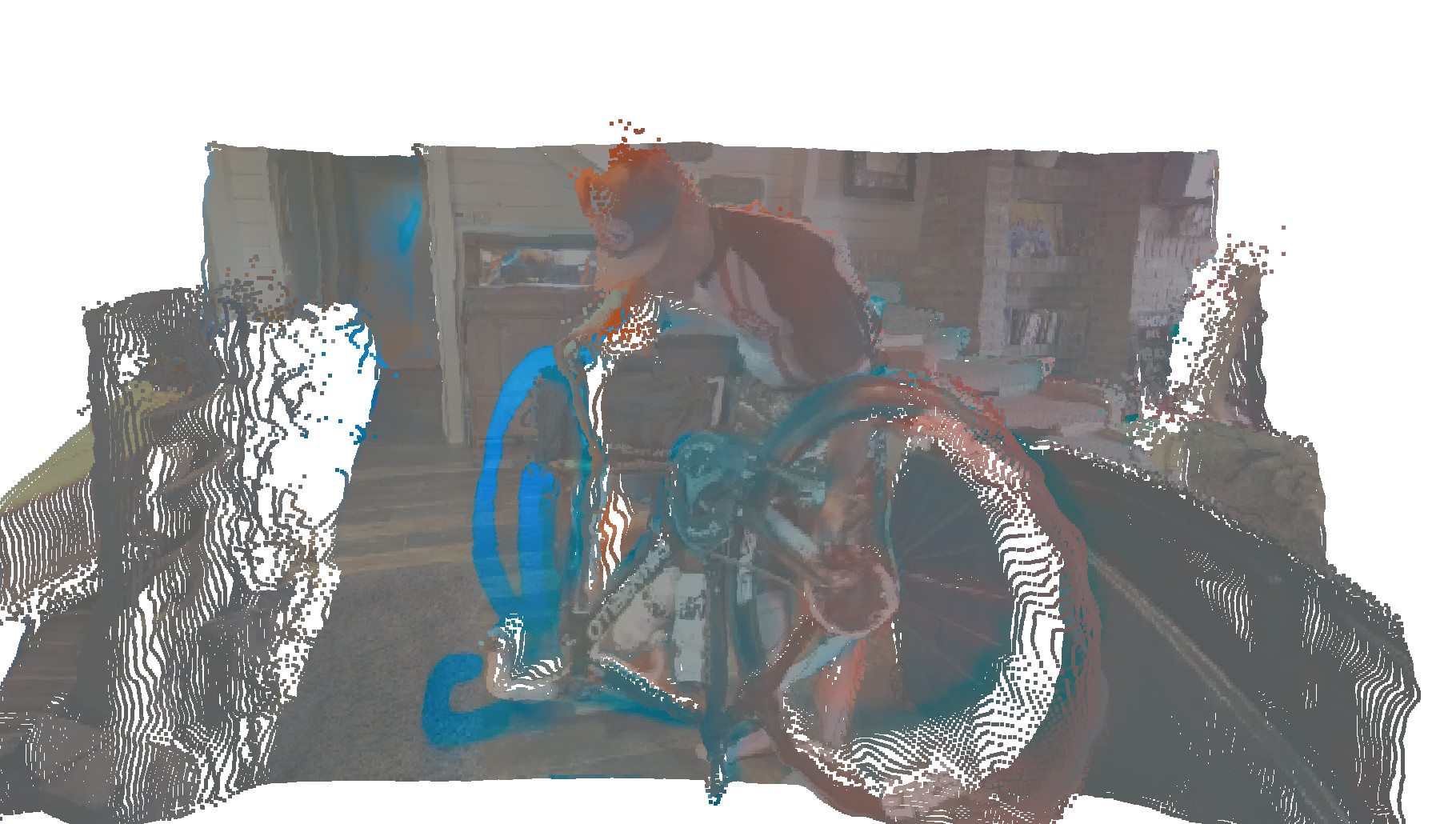}
        \\
    \end{tabular}
    }
    \vspace{-2mm}
    \caption{\textbf{Additional Qualitative Result on DAVIS~\cite{davis2016}}, 
    with scene flow represented by CIE-LAB~\cite{Hur:2021:SSM} colormap overlaid on corresponding 3D structure estimated for first input image.
    Our scene-flow result looks more consistent than other approaches with the bicycle turning motion in this example.}
    \label{tab:quali_davis-bike-packing}
\end{figure*}

\begin{figure*}[t]
    \setlength{\tabcolsep}{5pt}
    \renewcommand{\arraystretch}{0.6}  
    \vspace{-8mm}
    \resizebox{\textwidth}{!}{
        \tiny
    \begin{tabular}{rr}        
        Input & \\
        \includegraphics[width=.3\columnwidth,trim={0 0 0 0},clip]{figures/finalvis/davis/breakdance/left.jpg}
        &
        \\
        \phantom{x} & \\
        Self-Mono-SF~\cite{Hur:2020:SSM} & Ours \\
        \includegraphics[width=.3\columnwidth,trim={0 0 0 30pt},clip]{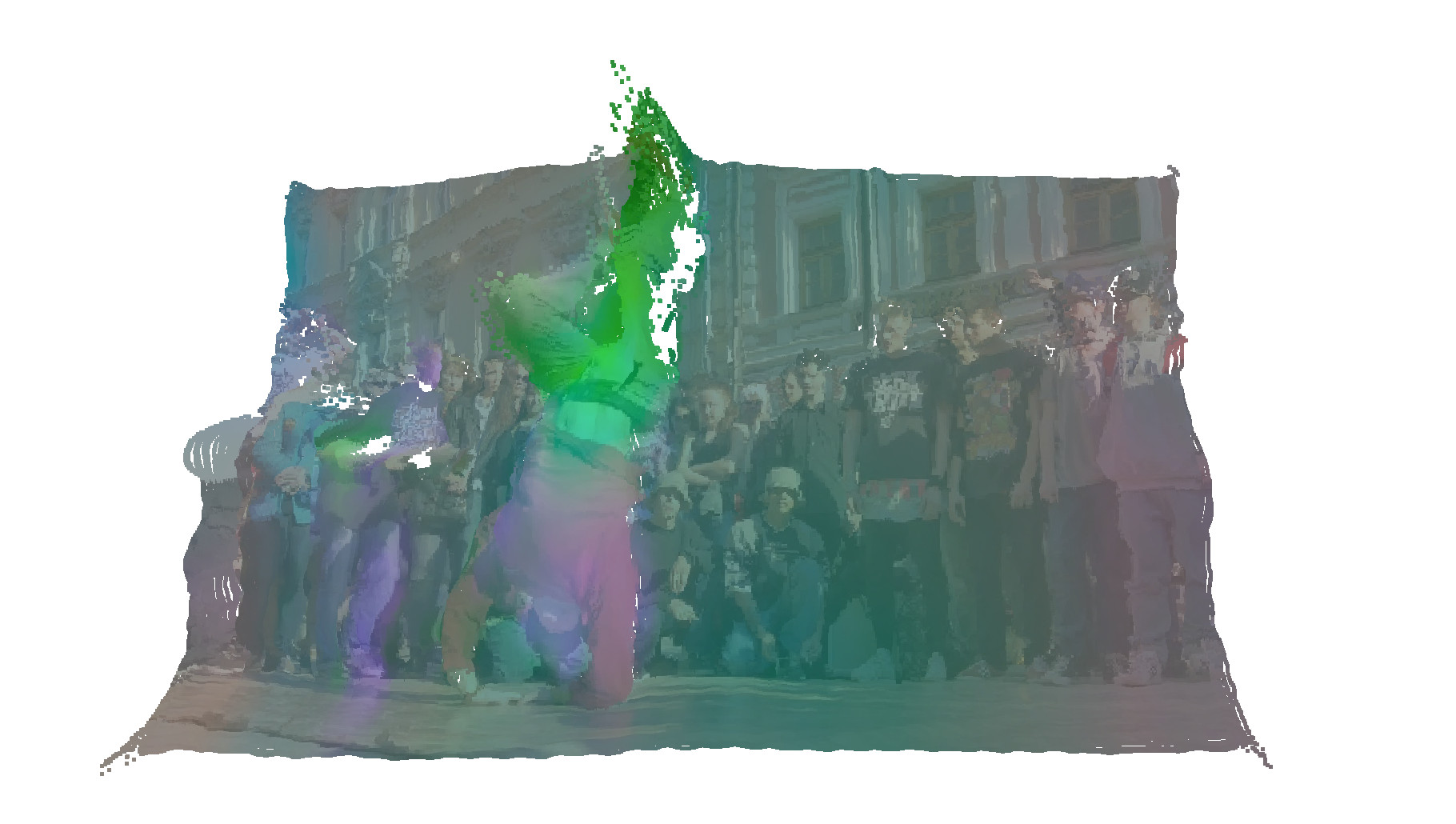}
        &
        \includegraphics[width=.3\columnwidth,trim={0 0 0 30pt},clip]{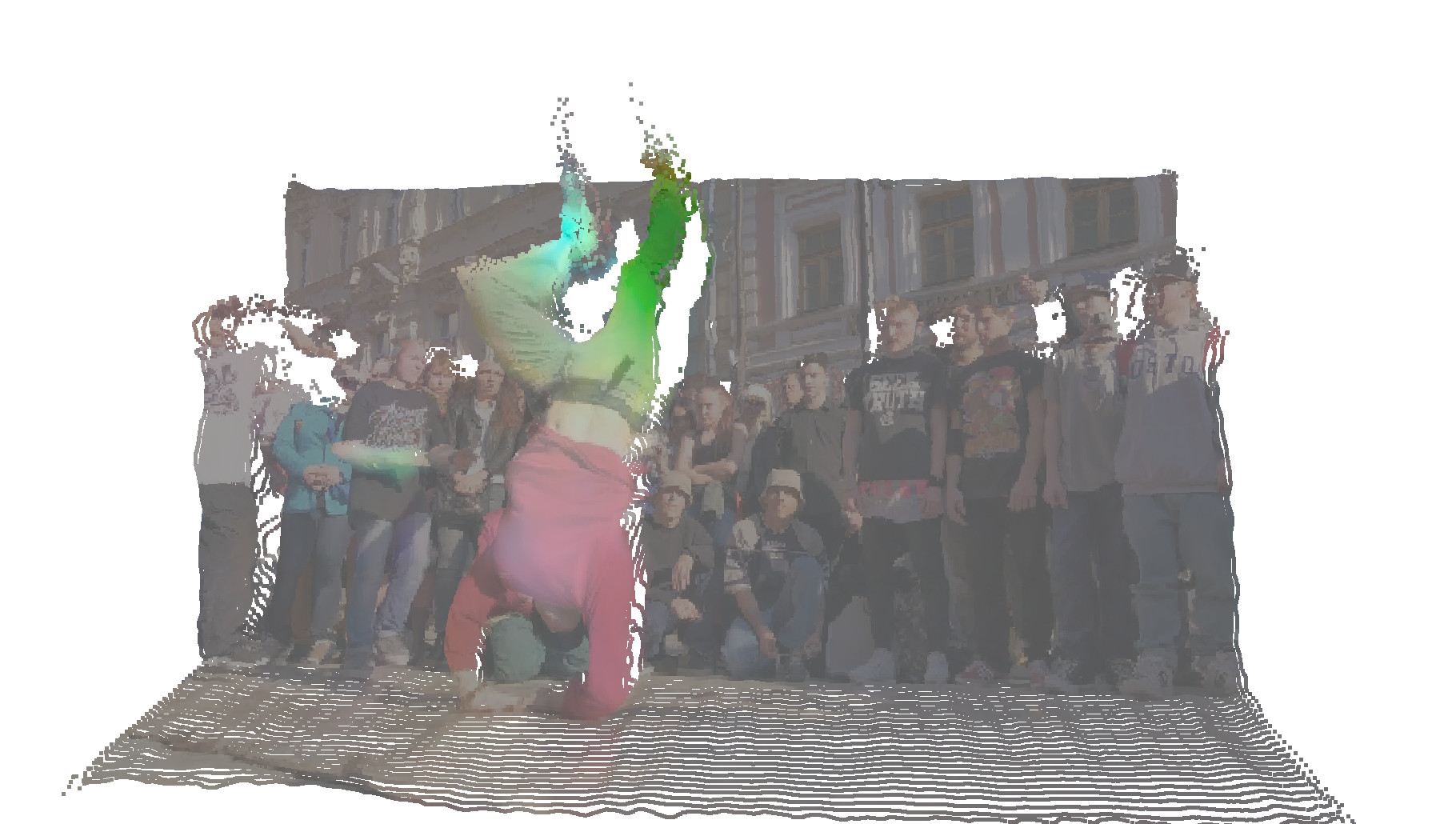}
        \\
        \phantom{x} & \\
        OpticalExpansion~\cite{yang2020upgrading} & MASt3R~\cite{mast3r}
       \\
       \includegraphics[width=.3\columnwidth,trim={0 0 0 30pt},clip]{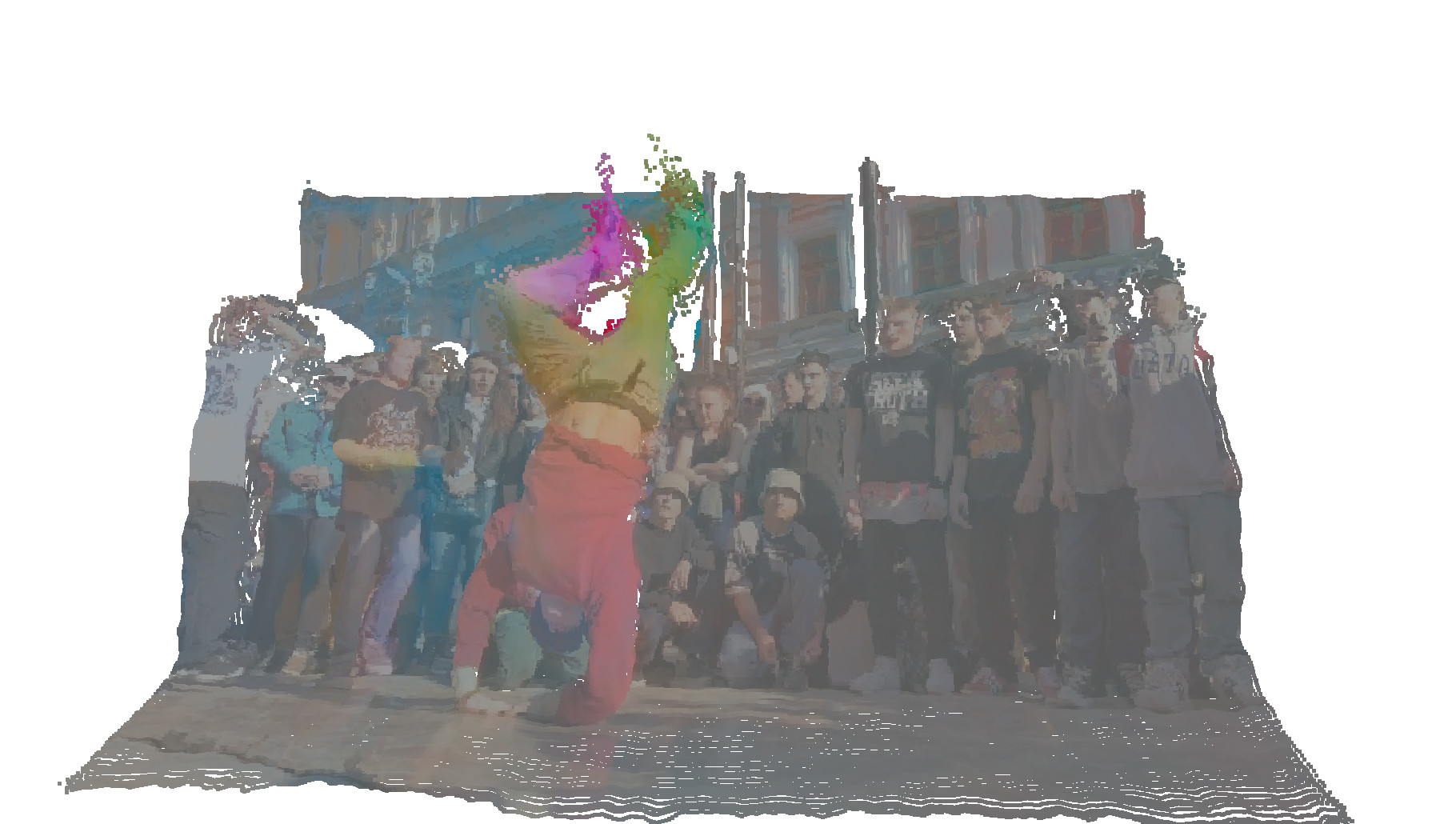}
        &
        \includegraphics[width=.3\columnwidth,trim={0 0 0 30pt},clip]{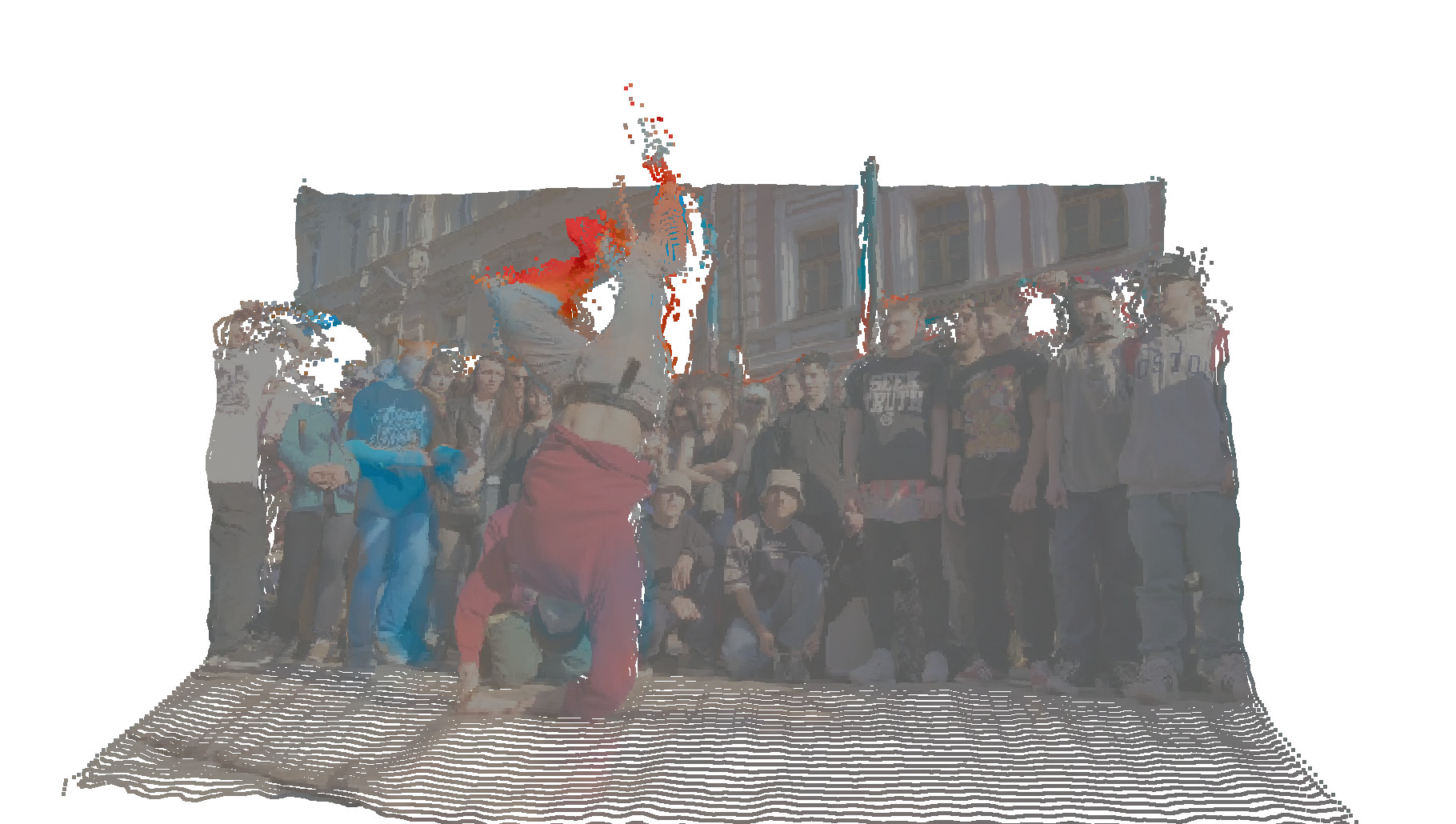}
        \\
    \end{tabular}
    }
    \vspace{-2mm}
    \caption{\textbf{Additional Qualitative Result on DAVIS~\cite{davis2016}}, 
    with scene flow represented by CIE-LAB~\cite{Hur:2021:SSM} colormap overlaid on corresponding 3D structure estimated for first input image.
    As compared to other approaches our approach correctly shows the motion on the legs of the dancer, while identifying most of the background as static.
    }
    \label{tab:quali_davis-breakdance}
\end{figure*}

\begin{figure*}[t]
    \setlength{\tabcolsep}{5pt}
    \renewcommand{\arraystretch}{0.6}  
    \vspace{-8mm}
    \resizebox{\textwidth}{!}{
        \tiny
    \begin{tabular}{rr}        
        Input \\
        \includegraphics[width=.3\columnwidth,trim={0 0 0 0},clip]{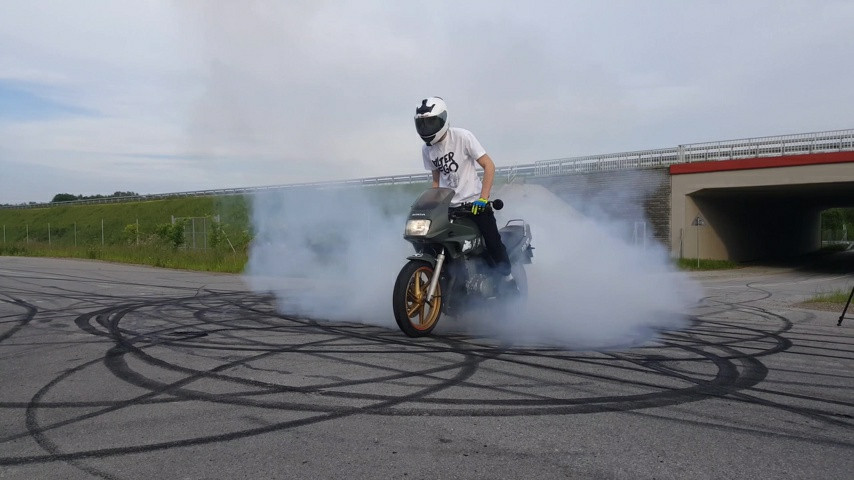}
        &
        \\
        \phantom{x} & \\
        Self-Mono-SF~\cite{Hur:2020:SSM} & Ours
        \\
        \includegraphics[width=.3\columnwidth,trim={0 0 0 10pt},clip]{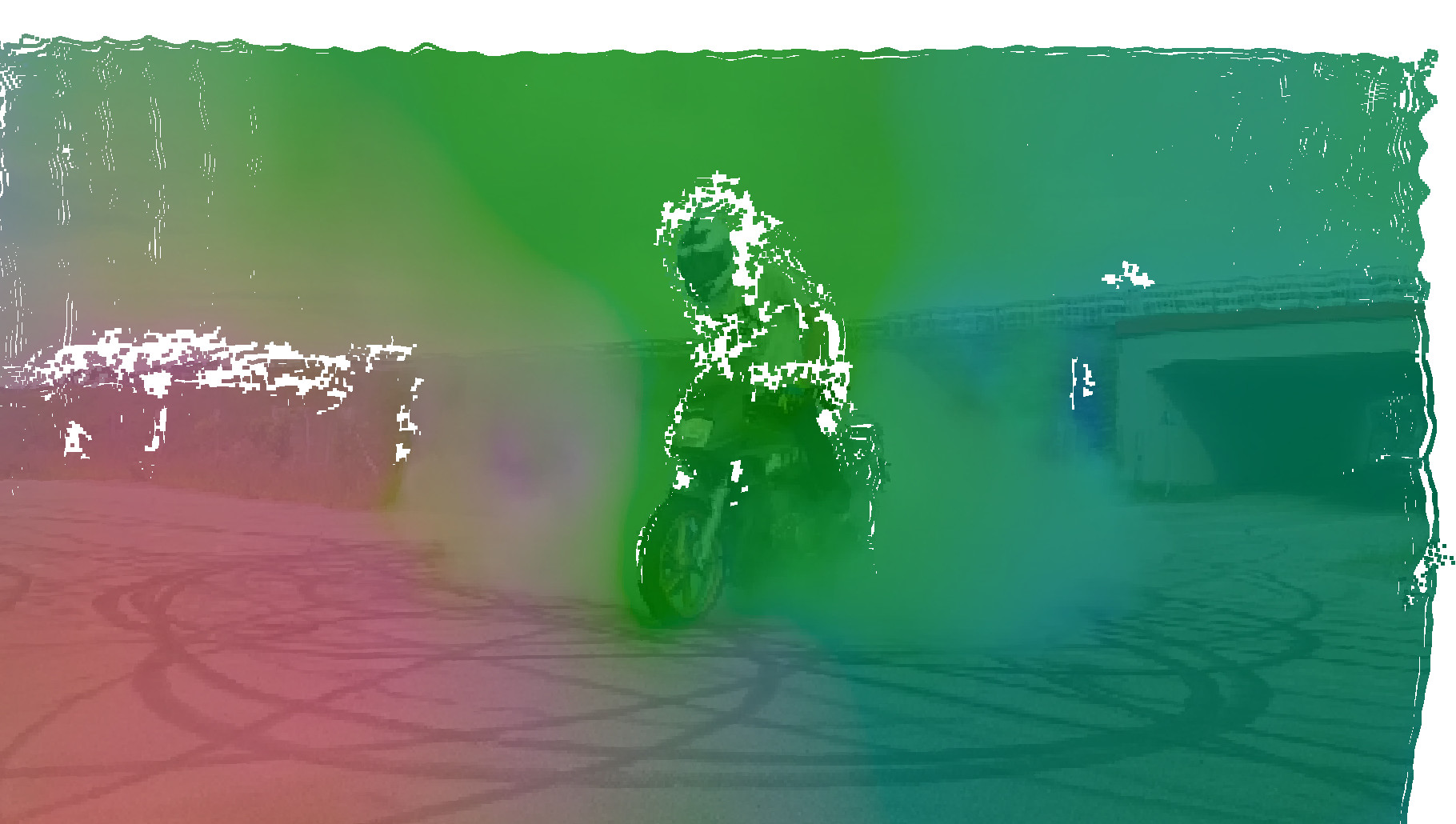}
        &
        \includegraphics[width=.3\columnwidth,trim={0 0 0 10pt},clip]{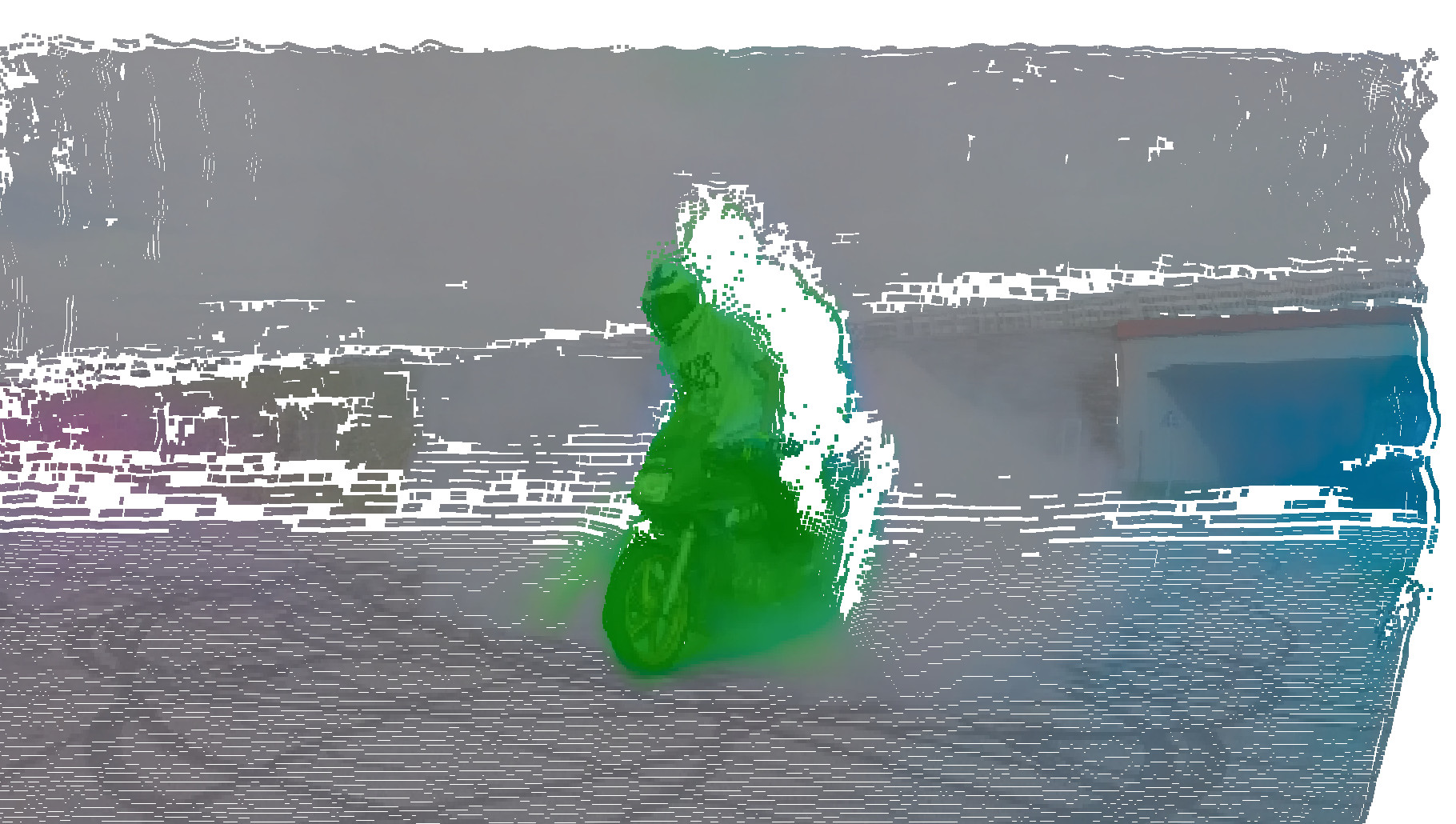}
        \\
        \phantom{x} & \\
        OpticalExpansion~\cite{yang2020upgrading} & MASt3R~\cite{mast3r}
        \\
        \includegraphics[width=.3\columnwidth,trim={0 0 0 10pt},clip]{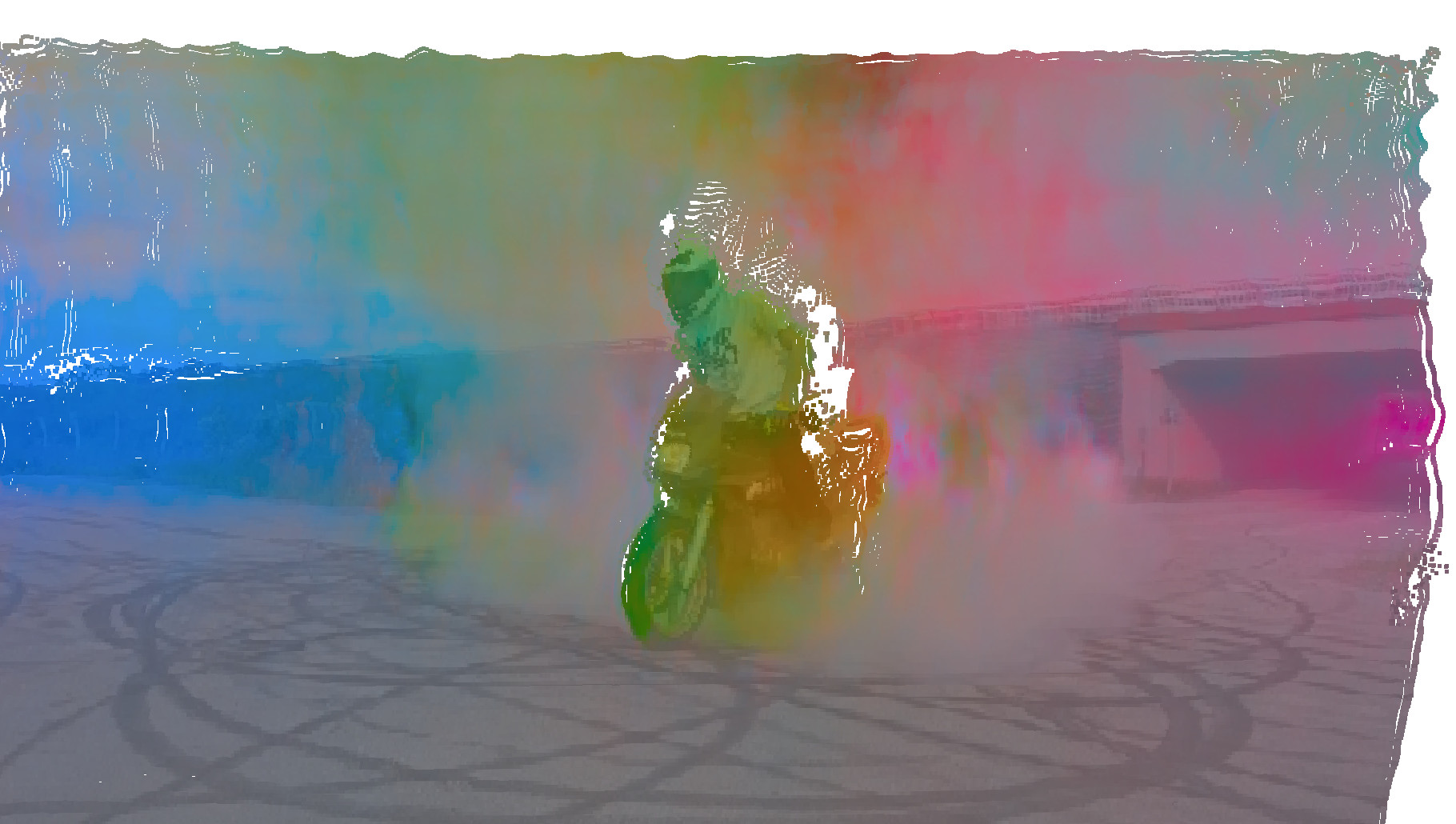}
        &
        \includegraphics[width=.3\columnwidth,trim={0 0 0 10pt},clip]{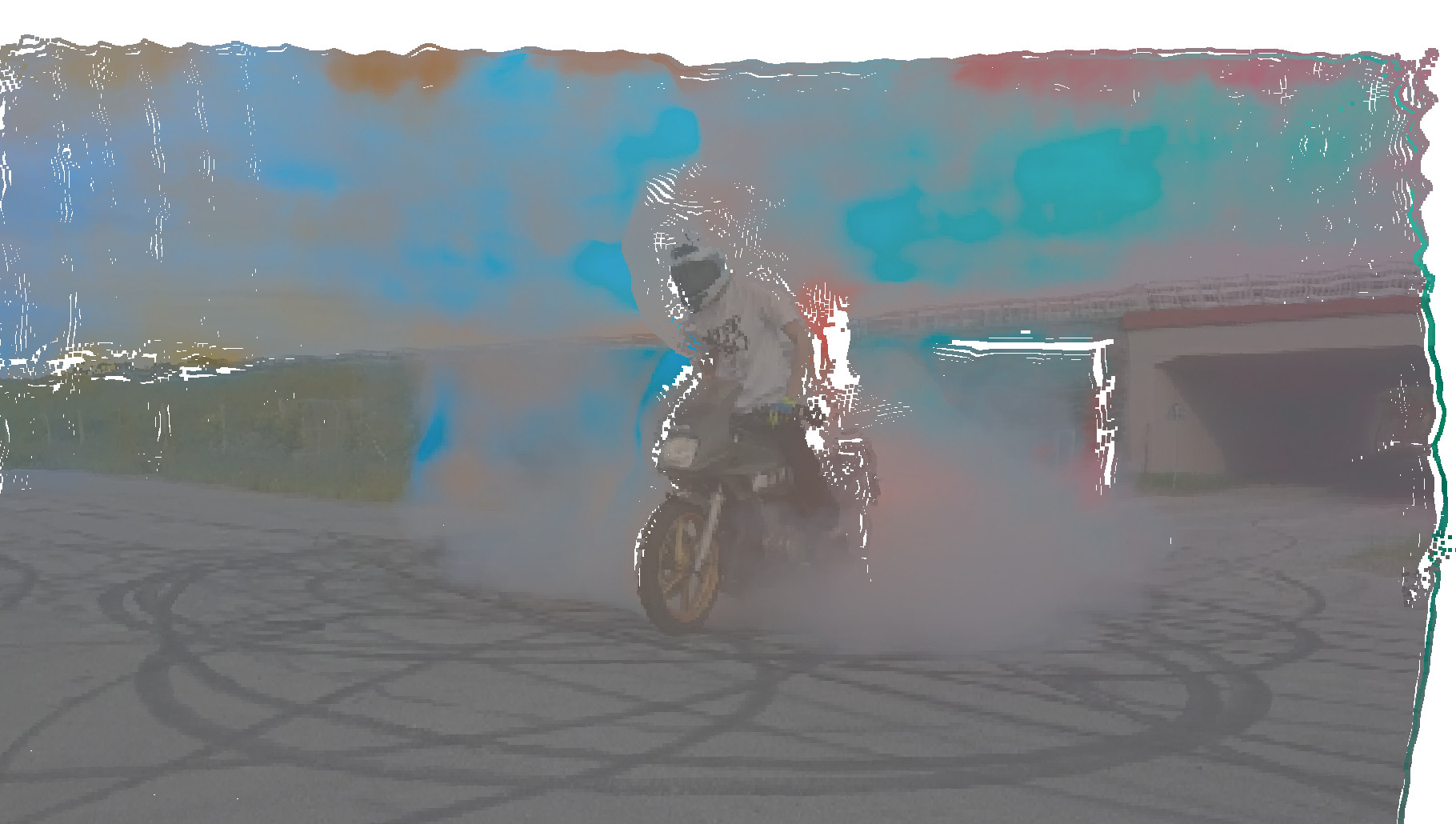}
        \\
    \end{tabular}
    }
    \vspace{-2mm}
    \caption{\textbf{Additional Qualitative Result on DAVIS~\cite{davis2016}}, 
    with scene flow represented by CIE-LAB~\cite{Hur:2021:SSM} colormap overlayed on corresponding 3D structure estimated for first input image.
    Our scene-flow result correctly identifies the foreground motion as compared to other approaches.}
    \label{tab:quali_davis-mbike-trick}
\end{figure*}

We show additional qualitative comparisons 
for KITTI (Fig.~\ref{tab:quali_kitti}) 
and DAVIS dataset (Fig.~\ref{tab:quali_davis-bike-packing},~\ref{tab:quali_davis-breakdance},~\ref{tab:quali_davis-mbike-trick}).
Overall, our approach is robust and performs consistently better across different datasets.

\section{Conclusion}

We present an algorithm for monocular scene flow that jointly estimate geometry and motion in a zero-shot feedforward fashion.
Our approach addresses three key challenges with new contributions: 
an architecture for joint geometry-motion estimation with a 3D prior, 
a data recipe to create 1\ M diverse training samples from general dynamic datasets and a scale-adaptive optimization to use them, 
and the careful selection of a scene flow parameterization that uses pointmaps with 3D motion offsets. 
The resulting model surpasses state-of-the-art methods and baselines built on large-scale models in terms of 3D EPE, and demonstrates strong zero-shot generalization to unseen scenarios from casually-captured videos in DAVIS to robotic manipulation videos in RoboTAP. 
This generalization ability makes scene flow estimation significantly more practical for real-world applications beyond autonomous driving, like augmented reality and robotics. 

Our method, benefiting from 3D prior knowledge learnt for geometry, 
currently cannot recover from any pitfalls of the pretrained model.
Future work should consider using more robust 3D priors to bootstrap our method.  
Also, camera ego-motion and scene motion are still entangled in our setting; 
further techniques could be introduced to decompose them and support more downstream applications.

\clearpage
{
    \small
    \bibliographystyle{ieeenat_fullname}
    \bibliography{main}
}

\appendix
\setcounter{page}{1}
\maketitlesupplementary

In this supplementary material, 
we present additional technical details and evaluations pertaining to our model architecture, dataset, and methodology. 
Please check our supplementary video for a high-level overview and abundant qualitative results with comparisons.

\section{Additional Experimental Setting}
\paragraph{Datasets.}

For evaluation on the Spring Dataset, we use a subset of the training set \{0005, 0009, 0013, 0017, 0023, 0037, 0044\}. 
For VKITTI2, we designate Scene18 as the test sequence due to its overlap with KITTI Scene Flow training data, preventing data contamination. 
VKITTI2 is created to reproduce several scenes of KITTI, thus we consider methods trained on VKITTI2 to be in-domain with KITTI experiments.

\paragraph{Implementation Details.}

All experiments are trained with $8$ NVIDIA A100 GPUs for $50$ epochs, which take $12$ hours.
We accumulate gradients every $2$ steps, and clip gradients by norm value $0.5$. 
We have $8$ DDP processes with a batch size of $4$, thus our effective batch size is $4\times8\times2=64$.

We select a learning rate of $1e^{-6}$ for the encoder-decoder, $1e^{-6}$ for the pointmap heads, and $1e^{-4}$ for the offset heads.
For the first $8$ epochs, all learning rates start from $0.1$ of above values, and gradually warm up.   
Note that, differently from DUSt3R~\cite{dust3r} or MASt3R~\cite{mast3r}, our heads do not predict confidence maps as we found empirically that they made our training unstable.

\noindent \underline{Training Datasets.} 
At each epoch, we re-sample $1e^4$ examples from the whole Data Recipe for training by balancing different datasets considering diversity and statistics~(\cref{tab:data-recipe}).
$10000$ samples for each epoch is made of: (SHIFT:$2024$, DynamicReplica:$1942$, VKITTI2:$958$, MOVi-F:$3560$, PointOdyssey:$1400$, Spring:$116$). 
We resize and crop all dataset samples into $288\times512$ resolution. 
For stereo cameras, we ignore right camera frames.

\noindent \underline{Resolution/Scale Alignment.} 
During evaluation, we resize inputs to be width-$512$ and resize outputs later. 
If scale alignment is required, we compute median scale between the predicted and ground-truth pointmap following DUSt3R~\cite{dust3r}, and scale both the pointmap and sceneflow.

\paragraph{Baselines.}
DUSt3R/MASt3R predict a pointmap for the second frame in the first frame's coordinate system.
However, for scene flow, we need geometry for the second frame in the second frame's coordinate system.
As such, for the Spring and VKITTI2 datasets, we use the ground truth camera pose to transform the pointmaps into the second frame's coordinate system such that we can create depth maps for the second frame.
For KITTI, we use DUSt3R's~\cite{dust3r} pose estimation method to infer camera pose.

\noindent \underline{Resolution/Scale Alignment.} 
For fariness, most methods resize inputs to be width-$512$ during inference.
Self-Mono-SF requires $256 \times 832$, while OpticalExpansion~\cite{yang2020upgrading} requires $384 \times 1280$.
For DUSt3R / MASt3R, we multiply pointmaps by estimating scaling factor.
For Self-Mono-SF and DepthAnythingV2-metric, the scale factor is computed for each depth estimation/groundtruth pair. 

\paragraph{Evaluation Metrics.}

\begin{description}[leftmargin=10pt,parsep=2pt,itemsep=2pt]
    \item\textbf{EPE}: $\|\widehat{sf} - sf\|_2$ averaged over each pixel, 
    where, $\widehat{sf}$ and $sf$ denote the estimated and ground truth scene flows respectively.
    \item\textbf{AccS}: Percentage of points where EPE $<0.05$ or relative error $<5\%$. 
    \item\textbf{AccR}: Percentage of points where EPE $<0.1$ or relative error $<10\%$. 
    \item\textbf{Out}: Percentage of points where EPE $>0.3$ or relative error $>10\%$.
    \item\textbf{AbsRel}: Absolute relative error $|d^* - d|/d$.
    \item$\boldsymbol{\delta_1}$: Percentage of $\max(d^*/d, d/d^*)< 1.25$.
\end{description}

\section{Additional experiments}

\subsection{Does Joint Estimation Help?}
\label{exp:prior-offset}

Key to our method's success is our design that forces the model to predict geometry and scene flow jointly.
To validate this statement we present an ablation study in \cref{tab:prior-offset}, which shows different pretraining strategies.

Our analysis allows us to make a few observations.
First, we compare the performance of our method when we train the offset head from scratch, and when we initialize it with either DUSt3R or MASt3R's pretrained model.
Initializing the offset head results in significantly lower (better) EPE and AbsR-r numbers in Tab.~\ref{tab:prior-offset}.
From them, the importance of high-quality 3D priors for the offset head is therefore clear, which also confirms the entanglement of depth and motion predictions.
Second, our approach outperforms the combination of MASt3R, a state-of-the-art depth estimation algorithm, and RAFT, a state-of-the-art optical flow algorithm---and the algorithm we use to generate the pseudo ground truth we train on.
Our approach combining MASt3R prior with the offset head achieves optimal results across both scene flow (EPE: $0.452$, AccR: $0.443$) and depth metrics (AbsR-r: $0.111$, $\delta_1-r$: $0.927$).

\begin{table}
    \centering
    \resizebox{\columnwidth}{!}{\begin{tabular}{l c@{\hspace{2mm}}c@{\hspace{2mm}}c@{\hspace{1mm}}c c@{\hspace{1mm}}c@{\hspace{2mm}}c@{\hspace{1mm}}c}
        \toprule
        \multirow{2}{*}{3D Prior} & \multicolumn{4}{c}{Scene Flow Estimation} & \multicolumn{4}{c}{Depth Estimation}\\ 
        \cmidrule(lr){2-5}  \cmidrule(lr){6-9} 
        & EPE$\downarrow$ & AccS$\uparrow$ & AccR$\uparrow$ & Out$\downarrow$ & AbsR-r$\downarrow$ & $\delta_1$-r$\uparrow$ & AbsR-m$\downarrow$ & $\delta_1$-m$\uparrow$\\
        \midrule
        Ours (scratch)  & 1.071 & \textbf{0.420} & \underline{0.442} & 0.957 & 0.353 & 0.433 & 0.547 & 0.049 \\
        Ours (DUSt3R~\cite{dust3r})  & \underline{0.588} & 0.378 & 0.408 & \underline{0.896} & 0.121 & 0.887 & \textbf{0.233} & \textbf{0.423} \\
        \multirow{1}{*}{MASt3R~\cite{mast3r}} 
         & 3.708 & 0.264 & 0.267 & 0.999 & \textbf{0.108} & \textbf{0.929} & 0.245 & 0.288 \\
        \rowcolor{gray!20}Ours (MASt3R~\cite{mast3r})  & \textbf{0.452} & \underline{0.398} & \textbf{0.443} & \textbf{0.873} & \underline{0.111} & \underline{0.927} & \underline{0.236} & \underline{0.345} \\
        \bottomrule
    \end{tabular}}
    \caption{\textbf{Ablation over Joint Estimation Pipelines.} Verify the effect of 3D Prior and Offset Heads on KITTI. In the table `Ours (w)' is our model initialized with the weights of method `w'.
    }
    \label{tab:prior-offset}
\end{table}

\subsection{Data Recipe is Key for Generalization}
\label{exp:data-recipe}

In \cref{tab:data-recipe}, we explore how the datasets used in training affect the performance on KITTI. 
We compare our method performance with three training strategies: training on all the datasets (all), training on all but VKITTI2 (exclude), and training only on VKITTI2 (specific).
Note that none of these strategies train on KITTI.
Unsurprisingly, training with all datasets yields the best overall performance for both scene flow and depth. 
However, when omitting VKITTI2, which is the only driving dataset use, we maintain robust performance (EPE: $0.641$, AbsR-m: $0.256$), which demonstrates the generalization capabilities gained with diverse data exposure.

\begin{table}
    \centering

    \resizebox{.85\columnwidth}{!}{%
    \begin{tabular}{c c@{\hspace{2mm}}c@{\hspace{2mm}}c@{\hspace{1mm}}c c@{\hspace{1mm}}c@{\hspace{2mm}}c@{\hspace{1mm}}c}
        \toprule
        \multirow{2}{*}{Data} & \multicolumn{4}{c}{Scene Flow Estimation} & \multicolumn{4}{c}{Depth Estimation}\\ 
        \cmidrule(lr){2-5}  \cmidrule(lr){6-9} 
        & EPE$\downarrow$ & AccS$\uparrow$ & AccR$\uparrow$ & Out$\downarrow$ & AbsR-r$\downarrow$ & $\delta_1$-r$\uparrow$ & AbsR-m$\downarrow$ & $\delta_1$-m$\uparrow$\\
        \midrule
        \multirow{1}{*}{\textit{exclude}} & 0.641 & 0.392 & 0.431 & 0.899 & 0.116 & 0.922 & 0.256 & 0.237 \\
        \multirow{1}{*}{\textit{specific}} & 0.569 & 0.317 & 0.393 & 0.911 & \textbf{0.090} & 0.925 & 0.580 & 0.095 \\
        \rowcolor{gray!20}\multirow{1}{*}{\textit{all}} & \textbf{0.452} & \textbf{0.398} & \textbf{0.443} & \textbf{0.873} & 0.111 & \textbf{0.927} & \textbf{0.236} & \textbf{0.345} \\
        \bottomrule
    \end{tabular}
    }
    \caption{\textbf{Ablation over Data Recipe} on KITTI.
    \textit{all}: using all datasets for training.
    \textit{exclude}: using all datasets, except for train set of VKITTI2.
    \textit{specific}: only using train set of VKITTI2.
    }
    \label{tab:data-recipe}
\end{table}

\subsection{Reference coordinate frame}

DUSt3R and MASt3R make predictions of pointmaps from pairs of images, where both sets of points are predicted in the coordinate frame of the first camera $C_1$. As scene flow estimation does not typically assume camera poses, it is defined as the (field of) vector from a 3D point in the first camera frame $C_1$ at time $t_1$ to a 3D point in the second camera frame $C_2$ at time $t_2$. This difference of output reference frames---either always predicting in $C_1$ or from $C_1$ into $C_2$---raises the question of whether the reference coordinate frame is important for prediction accuracy. This is especially true given that we fine tune the DUSt3R network that operates solely in $C_1$.

We ran an experiment to change the reference scene flow coordinate frame for our output scene flow: 
either 1) from $C_1$ into $C_2$, 
2) from $C_1$ to $C_1$ (sometimes called \emph{rectified scene flow}), or 
3) within a world coordinate frame. 
Scene flow training data labels are defined from $C_1$ to $C_2$ as in 1), so to produce the vector from $C_1$ to $C_1$ we must reproject the 3D end point of the vector into $C_1$'s reference frame using known ground truth camera poses. When computing metrics like end point error or accuracy against ground truth scene flow from $C_1$ to $C_2$, this transformation is also necessary. For synthetic datasets like Spring and VKITTI2, the transformation is simple as camera poses are known. For VKITTI2, the camera poses must be estimated and may be in error. 

Table ~\ref{tab:coordinateframes_quantitative} shows that the choice of coordinate frame is not significantly important when evaluated on Spring and VKITTI2, with any reference frame scoring approximately similarly. This suggests that the pretrained backbone can be successfully fine tuned with scene flow data and our training scheme for any coordinate frame. For VKITTI2, we observe large drops in performance as the reference frame changes. We attribute this to errors in the estimated camera poses, which affect training and metric evaluation.

\begin{table}[tb]
    \centering
    \caption{\textbf{Changing reference coordinate frame for the end point of scene flow.} For synthetic data with known camera poses from which to accurately translate coordinate frames for training data and evaluation, there is no clear gain from any of the reference frames (VKITTIv2, Spring). For real-world KITTI data, camera poses may be in error, and we see large increases in scene flow error when attempting to train and evaluate on scene flow data in a different reference frame.
    }
    \label{tab:coordinateframes_quantitative}
    \resizebox{\columnwidth}{!}{\begin{tabular}{l|cccc|cccc}
    \toprule
    Reference frame & \multicolumn{4}{c}{Scene Flow Estimation} & \multicolumn{4}{c}{Depth Estimation} \\
    \cmidrule(lr){1-2} \cmidrule(lr){2-5} \cmidrule(lr){6-9}
    VKITTI2 Dataset~\cite{cabon2020vkitti2} & EPE$\downarrow$ & AccS$\uparrow$ & AccR$\uparrow$ & Out$\downarrow$ & AbsR-r$\downarrow$ & $\delta_1$-r$\uparrow$ & AbsR-m$\downarrow$ & $\delta_1$-m$\uparrow$\\
    \emph{Known camera poses} & & & & & & & & \\
    World    & 0.154 & 0.744 & 0.830 & 0.511 & 0.140 & 0.856 & 0.222 & 0.682 \\
    Camera 1 & 0.202 & 0.808 & 0.886 & 0.449 & 0.139 & 0.858 & 0.210 & 0.711 \\
    Camera 2 & 0.190 & 0.780 & 0.876 & 0.484 & 0.137 & 0.859 & 0.236 & 0.634 \\
    \midrule
    Spring Dataset~\cite{Mehl2023_Spring} & EPE$\downarrow$ & AccS$\uparrow$ & AccR$\uparrow$ & Out$\downarrow$ & AbsR-r$\downarrow$ & $\delta_1$-r$\uparrow$ & AbsR-m$\downarrow$ & $\delta_1$-m$\uparrow$\\
    \emph{Known camera poses} & & & & & & & & \\
    World    & 0.010 & 0.986 & 0.998 & 0.812 & 0.265 & 0.600 & 1.132 & 0.043 \\
    Camera 1 & 0.012 & 0.990 & 1.000 & 0.811 & 0.272 & 0.594 & 0.716 & 0.103 \\
    Camera 2 & 0.013 & 0.989 & 0.999 & 0.813 & 0.280 & 0.597 & 0.679 & 0.119 \\
    \midrule
    KITTI~\cite{Menze2015ObjectSF} & EPE$\downarrow$ & AccS$\uparrow$ & AccR$\uparrow$ & Out$\downarrow$ & AbsR-r$\downarrow$ & $\delta_1$-r$\uparrow$ & AbsR-m$\downarrow$ & $\delta_1$-m$\uparrow$\\
    \emph{Estimated camera poses} & & & & & & & & \\
    World    & 2.637 & 0.268 & 0.271 & 0.995 & 0.110 & 0.928 & 0.212 & 0.546 \\
    Camera 1 & 4.202 & 0.251 & 0.253 & 0.992 & 0.108 & 0.928 & 0.214 & 0.519 \\
    Camera 2 & 0.452 & 0.398 & 0.443 & 0.873 & 0.111 & 0.927 & 0.236 & 0.345 \\
    \bottomrule
    \end{tabular}
    }
\end{table}

\subsection{Additional quantitative evaluation}

We show our method's superiority with comparison with different checkpoints of previous monocular scene flow methods in \cref{tab:add_msf}.

\begin{table*}
    \centering
    \resizebox{0.8\linewidth}{!}{%
    \begin{tabular}{@{\hspace{1mm}}cl@{\hspace{1mm}}c@{\hspace{1mm}}c@{\hspace{1mm}}c@{\hspace{1mm}}c c@{\hspace{1mm}}c@{\hspace{1mm}}c@{\hspace{1mm}}c}
        \toprule
        \multicolumn{10}{c}{KITTI~\cite{Menze2015ObjectSF} (Real)} \\
        \midrule
        \multirow{2}{*}{}  & \multirow{2}{*}{Method} & \multicolumn{4}{c}{Scene Flow Estimation} & \multicolumn{4}{c}{Depth Estimation}\\ 
        \cmidrule(lr){3-6}  \cmidrule(lr){7-10} 
        & & EPE$\downarrow$ & AccS$\uparrow$ & AccR$\uparrow$ & Out$\downarrow$ & AbsR-r$\downarrow$ & $\delta_1$-r$\uparrow$ & AbsR-m$\downarrow$ & $\delta_1$-m$\uparrow$\\
        \midrule
        \multirow{5}{*}{\parbox{0mm}{\vspace{-0mm}\rotatebox[origin=c]{90}{In}}} 
        & \textit{Ours} & 0.452 & 0.398 & 0.443 & 0.873 & 0.111 & 0.927 & 0.236 & 0.345 \\
        & Self-Mono-SF~\cite{Hur:2020:SSM}-kitti-train & 0.454 & 0.345 & 0.435 & 0.853 & 0.100 & 0.905 & 0.105 & 0.879 \\
        & Self-Mono-SF~\cite{Hur:2020:SSM}-kitti-eigen & 0.517 & 0.392 & 0.457 & 0.870 & 0.085 & 0.929 & 0.089 & 0.917 \\
        & OpticalExpansion~\cite{yang2020upgrading}-kitti-train & 2.703 & 0.357 & 0.366 & 0.970 & 0.132 & 0.871 & 0.529 & 0.172 \\
        & OpticalExpansion~\cite{yang2020upgrading}-kitti-trainval & 2.682 & 0.359 & 0.367 & 0.967 & 0.132 & 0.871 & 0.529 & 0.172 \\
        \midrule
        \multirow{3}{*}{\parbox{0mm}{\vspace{-0mm}\rotatebox[origin=c]{90}{Out}}}  
        & \textit{Ours}-\textit{exclude} & 0.641 & 0.392 & 0.431 & 0.899 & 0.116 & 0.922 & 0.256 & 0.237 \\
        & OpticalExpansion~\cite{yang2020upgrading}-driving & 2.537 & 0.348 & 0.363 & 0.968 & 0.132 & 0.871 & 0.529 & 0.172 \\
        & OpticalExpansion~\cite{yang2020upgrading}-robust & 2.702 & 0.359 & 0.366 & 0.969 & 0.132 & 0.871 & 0.529 & 0.172 \\
        \toprule
        \multicolumn{10}{c}{VKITTI2~\cite{cabon2020vkitti2} (Real)} \\
        \midrule
        \multirow{2}{*}{} & \multirow{2}{*}{Method} & \multicolumn{4}{c}{Scene Flow Estimation} & \multicolumn{4}{c}{Depth Estimation}\\ 
        \cmidrule(lr){3-6}  \cmidrule(lr){7-10} 
        & & EPE$\downarrow$ & AccS$\uparrow$ & AccR$\uparrow$ & Out$\downarrow$ & AbsR-r$\downarrow$ & $\delta_1$-r$\uparrow$ & AbsR-m$\downarrow$ & $\delta_1$-m$\uparrow$\\
        \midrule
        \multirow{5}{*}{\parbox{0mm}{\vspace{-0mm}\rotatebox[origin=c]{90}{In}}} 
        & \textit{Ours} & 0.190 & 0.780 & 0.876 & 0.484 & 0.137 & 0.859 & 0.236 & 0.634 \\
        & Self-Mono-SF~\cite{Hur:2020:SSM}-kitti-train & 0.294 & 0.694 & 0.751 & 0.589 & 0.244 & 0.568 & 0.224 & 0.589 \\
        & Self-Mono-SF~\cite{Hur:2020:SSM}-kitti-eigen & 0.373 & 0.688 & 0.738 & 0.666 & 0.200 & 0.668 & 0.189 & 0.669 \\
        & OpticalExpansion~\cite{yang2020upgrading}-kitti-train & 1.879 & 0.653 & 0.663 & 0.947 & 0.194 & 0.727 & 0.203 & 0.746 \\
        & OpticalExpansion~\cite{yang2020upgrading}-kitti-trainval & 1.845 & 0.655 & 0.664 & 0.946 & 0.194 & 0.727 & 0.203 & 0.746 \\
        \midrule
        \multirow{3}{*}{\parbox{0mm}{\vspace{-0mm}\rotatebox[origin=c]{90}{Out}}}
        & \textit{Ours}-\textit{exclude} & 0.409 & 0.727 & 0.772 & 0.696 & 0.144 & 0.836 & 0.168 & 0.778 \\
        & OpticalExpansion~\cite{yang2020upgrading}-driving & 1.809 & 0.654 & 0.662 & 0.946 & 0.194 & 0.727 & 0.203 & 0.746 \\
        & OpticalExpansion~\cite{yang2020upgrading}-robust & 1.915 & 0.654 & 0.662 & 0.958 & 0.194 & 0.727 & 0.203 & 0.746 \\
        \toprule
        \multicolumn{10}{c}{Spring~\cite{Mehl2023_Spring} (Synthetic)} \\
        \midrule
        \multirow{2}{*}{} & \multirow{2}{*}{Method} & \multicolumn{4}{c}{Scene Flow Estimation} & \multicolumn{4}{c}{Depth Estimation}\\ 
        \cmidrule(lr){3-6}  \cmidrule(lr){7-10} 
        & & EPE$\downarrow$ & AccS$\uparrow$ & AccR$\uparrow$ & Out$\downarrow$ & AbsR-r$\downarrow$ & $\delta_1$-r$\uparrow$ & AbsR-m$\downarrow$ & $\delta_1$-m$\uparrow$\\
        \midrule
        \multirow{1}{*}{\parbox{0mm}{\vspace{-0mm}\rotatebox[origin=c]{90}{In}}} 
        & \textit{Ours} & 0.013 & 0.989 & 0.999 & 0.813 & 0.280 & 0.597 & 0.679 & 0.119 \\
        \midrule
        \multirow{7}{*}{\parbox{0mm}{\vspace{-0mm}\rotatebox[origin=c]{90}{Out}}}
        & \textit{Ours}-\textit{exclude} & 0.014 & 0.992 & 1.000 & 0.787 & 0.294 & 0.599 & 0.682 & 0.130 \\
        & Self-Mono-SF~\cite{Hur:2020:SSM}-kitti-train & 1.005 & 0.251 & 0.328 & 0.880 & 0.501 & 0.353 & 0.784 & 0.044 \\
        & Self-Mono-SF~\cite{Hur:2020:SSM}-kitti-eigen & 0.717 & 0.272 & 0.365 & 0.895 & 0.532 & 0.344 & 0.772 & 0.050 \\
        & OpticalExpansion~\cite{yang2020upgrading}-driving & 0.029 & 0.873 & 0.971 & 0.805 & 0.430 & 0.468 & 0.804 & 0.004 \\
        & OpticalExpansion~\cite{yang2020upgrading}-kitti-train & 0.038 & 0.849 & 0.935 & 0.799 & 0.430 & 0.468 & 0.804 & 0.004 \\
        & OpticalExpansion~\cite{yang2020upgrading}-kitti-trainval & 0.027 & 0.916 & 0.967 & 0.800 & 0.430 & 0.468 & 0.804 & 0.004 \\
        & OpticalExpansion~\cite{yang2020upgrading}-robust & 0.016 & 0.969 & 0.995 & 0.801 & 0.430 & 0.468 & 0.804 & 0.004 \\
        \bottomrule
    \end{tabular}
    }
    \caption{\textbf{Comparison with Monocular Scene Flow Methods.}\\
    {
        \setlength{\tabcolsep}{0pt}
        \begin{tabular}{ll}
        \textit{Self-Mono-SF~\cite{Hur:2020:SSM}}-kitti-train:&\hspace{0.5cm}Trained on KITTI, train split;\\
        \textit{Self-Mono-SF~\cite{Hur:2020:SSM}}-kitti-eigen:&\hspace{0.5cm}Trained on KITTI, eigen split;\\
        \textit{OpticalExpansion~\cite{yang2020upgrading}}-driving:&\hspace{0.5cm}Trained on Driving~\cite{mayer2016};\\
        \textit{OpticalExpansion~\cite{yang2020upgrading}}-kitti-train:&\hspace{0.5cm}Trained on Driving~\cite{mayer2016}, and then finetune on KITTI, train split;\\
        \textit{OpticalExpansion~\cite{yang2020upgrading}}-kitti-trainval:&\hspace{0.5cm}Trained on Driving~\cite{mayer2016}, and then finetune on KITTI, trainval split;\\
        \textit{OpticalExpansion~\cite{yang2020upgrading}}-robust:&\hspace{0.5cm}Trained for the Robust Vision Challenge.\\
        \end{tabular}
    }
    }
    \label{tab:add_msf}
\end{table*}

\end{document}